\documentclass[10pt,twocolumn,letterpaper]{article}

\usepackage[pagenumbers]{cvpr} 

\renewcommand{\paragraph}[1]{\vspace{.5em}\noindent\textbf{#1.}}

\definecolor{cvprblue}{rgb}{0.21,0.49,0.74}
\usepackage[pagebackref,breaklinks,colorlinks,allcolors=cvprblue]{hyperref}

\usepackage{multirow}
\usepackage{multicol}
\usepackage{xcolor}         
\usepackage{colortbl}   
\usepackage{tabularx}
\usepackage{float}
\usepackage{tcolorbox}
\usepackage{makecell}
\newcommand{\highg}{\cellcolor{backgreen}}
\newcommand{\high}{\cellcolor{backblue}}
\usepackage{xspace}
\usepackage[shortlabels]{enumitem}
\usepackage{placeins}
\usepackage{listings}

\definecolor{backblue}{RGB}{210, 230, 250}
\definecolor{backgreen}{RGB}{226, 240, 217}
\definecolor{deepblue}{RGB}{59,113,170}
\definecolor{contentwhite}{RGB}{255,255,255}

\def\paperID{xxxxx} 
\def\confName{CVPR}
\def\confYear{2026}

\title{Chart-FR1: Visual Focus-Driven Fine-Grained Reasoning on Dense Charts}

\author{
    Hongkun Pan$^{1}$
    \quad Yuwei Wu$^{1}$
    \quad Wanyi Hong$^{1}$
    \quad Shenghui Hu$^{1}$
    \quad Qitong Yan$^{1}$ 
    \quad Yi Yang$^{2}$ \\
    \quad Rufei Han$^{3}$
    \quad Changju Zhou$^{3}$ 
    \quad Minfeng Zhu$^{1*}$
    \quad Dongming Han$^{1,3*}$
    \quad Wei Chen$^{2}$  \\
    $^1$ Zhejiang University \quad
    $^2$ State Key Lab of CAD\&CG, Zhejiang University \quad
    $^3$ HiThink Research
}

\begin{document}

\twocolumn[{%
\renewcommand\twocolumn[1][]{#1}%
\maketitle
\vspace{-1.0em}
\includegraphics[width=\textwidth]{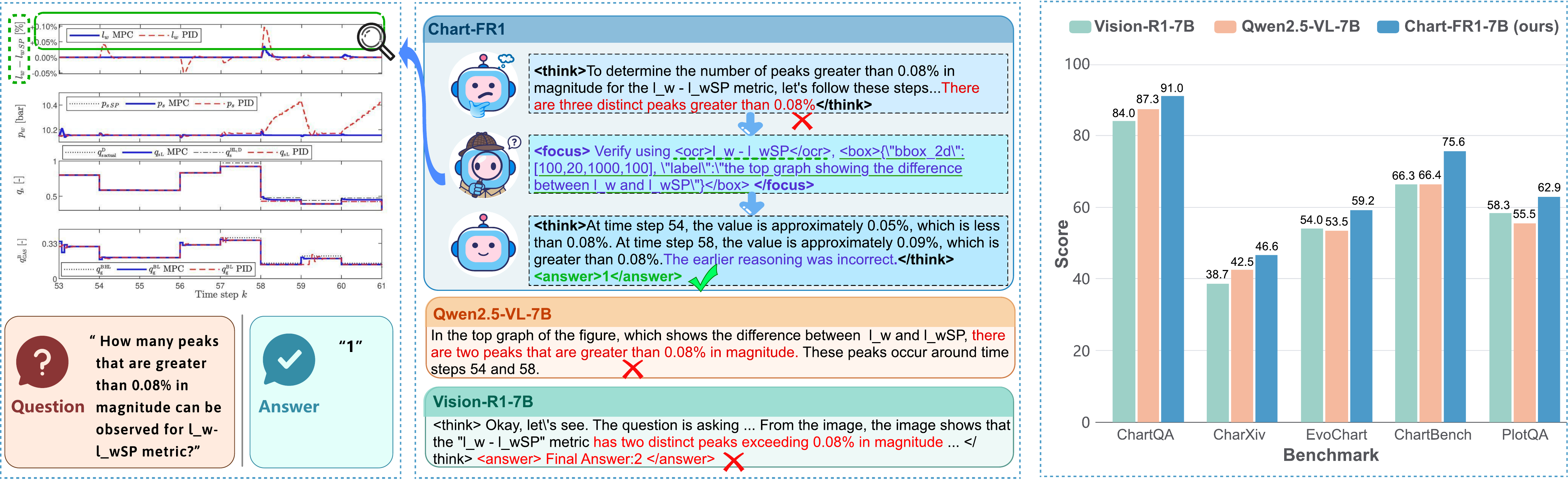}
\captionof{figure}{
Comparison between baseline models and Chart-FR1. This example shows that Qwen2.5-VL-7B fails to capture fine-grained key information for deep reasoning, while Vision-R1-7B struggles to associate reasoning steps with visual cues. In contrast, Chart-FR1-7B effectively focuses on key evidence for fine-grained reasoning. The superiority of our model is further demonstrated by the bar chart.
}
\vspace{1.0em}
\label{fig:teaser}
}]

\footnote{$^*$Corresponding author.}

\begin{abstract}
Multimodal large language models (MLLMs) have shown considerable potential in chart understanding and reasoning tasks. However, they still struggle with high information density (HID) charts characterized by multiple subplots, legends, and dense annotations due to three major challenges: (1) limited fine-grained perception results in the omission of critical visual cues; (2) redundant or noisy visual information undermines the performance of multimodal reasoning; (3) lack of adaptive deep reasoning relative to the amount of visual information. To tackle these challenges, we present a novel focus-driven fine-grained chart reasoning model, \textbf{Chart-FR1}, to improve perception, focusing efficiency, and adaptive deep reasoning on HID charts. Specifically, we propose \textbf{Focus-CoT}, a visual focusing chain-of-thought that enhances fine-grained perception by explicitly linking reasoning steps to key visual cues, such as local image regions and OCR signals. Building on this, we introduce \textbf{Focus-GRPO}, a focus-driven reinforcement learning algorithm with an information-efficiency reward that compresses redundant visual information for efficient focusing, and an adaptive KL penalty mechanism that enables flexible control over reasoning depth as more visual cues are discovered. Furthermore, to fill the gap in benchmarks for HID charts, we build \textbf{HID-Chart}, a challenging benchmark with an information-density metric designed to evaluate fine-grained chart reasoning capabilities. Extensive experiments on multiple chart benchmarks demonstrate that Chart-FR1 outperforms state-of-the-art MLLMs in chart understanding and reasoning. Code is available at \url{https://github.com/phkhub/Chart-FR1}.

\end{abstract}    
\begin{figure*}[t]
\centering
\includegraphics[width=\textwidth]{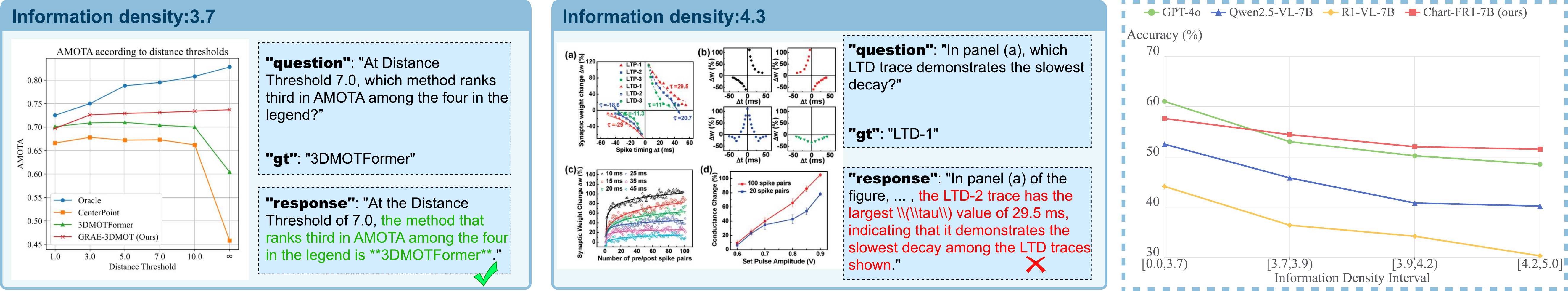}
\caption{An example of current MLLM (i.e., Qwen2.5-VL-7B) performance on two charts with different information density. The line chart reveals that existing MLLMs, such as GPT-4o, Qwen2.5-VL-7B, and R1-VL-7B, exhibit significant performance degradation as the information density of the chart increases. In contrast, Chart-FR1 demonstrates stable and robust performance.}
\label{img:challenge}
\end{figure*}

\section{Introduction}
Chart understanding and reasoning are crucial but challenging tasks. Charts encode quantitative information through the combination of text, geometric shapes, colors, and layouts. As information becomes densely packed, visual clutter severely hinders both perception and reasoning. Thus, processing high information density (HID) charts requires not only fine-grained perception to extract salient cues, but also deep compositional reasoning over these visual cues.

Recent advances in Multimodal Large Language Models (MLLMs), including general-purpose models such as GPT-4o~\cite{hurst2024gpt4o}, Qwen2.5-VL~\cite{bai2025qwen} and Gemma3~\cite{gemma2025gemma3}, as well as chart-specific MLLMs like ChartReasoner~\cite{jia2025chartreasoner}, EvoChart~\cite{huang2025evochart} and ChartGemma~\cite{masry2024chartgemma} have demonstrated remarkable progress in chart understanding. Furthermore, reasoning MLLMs such as R1-VL~\cite{zhang2025r1vl}, R1-Onevision~\cite{yang2025r1onevision}, Vision-R1~\cite{huang2025visionr1}, and OpenVLThinker~\cite{deng2025openvlthinker} leverage reinforcement learning (RL)~\cite{guo2025deepseek, Shulman2017PPO, kaelbling1996reinforcement, Rafailov2023DPO} to enhance multimodal deep reasoning capabilities.

Nevertheless, Fig.\ref{fig:teaser} highlights that MLLMs still face significant challenges in HID charts due to the following limitations:
(1) Limited fine-grained perception: most models rely on global visual embeddings, lacking fine-grained perception to isolate crucial visual cues from dense information.
(2) Visual redundancy and noise: MLLMs struggle to perform sufficiently deep reasoning with captured visual elements. The presence of redundant information further degrades the accuracy of reasoning.
(3) Lack of adaptive reasoning: existing RL algorithms employ a fixed KL penalty to constrain the divergence between the policy and reference models for training stability. This leads to over-penalization when longer reasoning chains with extensive visual cues are required during policy optimization.
These limitations make it difficult for current MLLMs to handle the challenges of HID charts.

Furthermore, existing chart benchmarks~\cite{masry2022chartqa, wang2024charxiv, methani2020plotqa, huang2025evochart, xu2023chartbench} exhibit limitations in chart diversity, domain coverage, and information density. Consequently, they are insufficient for rigorously evaluating MLLMs on HID charts, which hinders further progress in fine-grained chart reasoning.

In this paper, we develop a focus-driven fine-grained chart reasoning model termed \textbf{Chart-FR1} to enhance perception, maximize focusing efficiency, and enable adaptive deep reasoning in HID scenarios.
First, we propose \textbf{Focus-CoT}, a visual focusing Chain-of-Thought mechanism that explicitly augments reasoning steps with fine-grained visual evidence such as region-level image patches and OCR information. In addition, we design an automated data synthesis pipeline that generates high-quality Focus-CoT data to bootstrap the focusing behavior through supervised fine-tuning.
Then, we introduce \textbf{Focus-GRPO}, a focus-driven reinforcement learning algorithm that refines visual focusing and regulates reasoning complexity. Focus-GRPO includes an information-efficiency reward that penalizes redundant visual content to encourage efficient focusing, and an adaptive KL penalty that dynamically adjusts the constraint on reasoning depth during policy optimization according to the richness of visual cues. Also, the relaxed-accuracy reward provides a more stable learning environment in scenarios involving ambiguous numerical answers.

To systematically evaluate the performance of MLLMs on HID charts, we define an information-density metric and construct \textbf{HID-Chart}, which features diverse chart types, broad domains, and high information density, providing a comprehensive evaluation of fine-grained reasoning ability. A challenge highlighted by the HID-Chart is shown in Fig.~\ref{img:challenge}. When information density increases, visual clutter significantly degrades both perception and reasoning.

In summary, our main contributions are as follows:
\begin{enumerate}
    \item We propose \textbf{Focus-CoT} to explicitly link key visual cues with reasoning steps, enhancing the fine-grained reasoning capabilities of MLLMs in HID charts.
    \item We introduce \textbf{Focus-GRPO} to optimize visual focusing efficiency and facilitate adaptive deep reasoning through information-efficiency reward, relaxed-accuracy reward, and adaptive KL penalty mechanism.
    \item We construct a challenging HID chart reasoning benchmark \textbf{HID-Chart} and a chart information-density metric to fill the evaluation gap in this domain.
    \item We develop \textbf{Chart-FR1}, a model trained with a two-stage focused reasoning paradigm that achieves state-of-the-art performance among advanced MLLMs.
\end{enumerate}

\section{Related Work}

\noindent\textbf{Multimodal Large Language Models.} Multimodal large language models (MLLMs)~\cite{hurst2024gpt4o, bai2025qwen, liu2023llava, chen2024internvl, alayrac2022flamingo, li2025llavaonevision, kwai2025keye, wu2024deepseekvl2} build upon LLMs~\cite{brown2020language_model, achiam2023gpt4, touvron2023llama} by integrating novel architectures and multi-stage training paradigms, thereby achieving strong cross-modal understanding and reasoning capabilities. Furthermore, some research has attempted to apply MLLMs to charts. ChartLlama~\cite{han2023chartllama}, ChartGemma~\cite{masry2024chartgemma} and EvoChart~\cite{huang2025evochart} employ instruction tuning on synthesized instruction datasets, but their performance is constrained by the scale and quality of the instructions. TinyChart~\cite{zhang2024tinychart} introduces Program-of-Thought (PoT) reasoning through the generation of Python code. ChartReasoner~\cite{jia2025chartreasoner} preserves layout and semantics by converting charts into ECharts code. ChartPoint~\cite{xu2025chartpoint} associates local regions during reasoning and ChartSketcher~\cite{huang2025chartsketcher} adopts multi-round interactive reasoning with code-based annotations. However, they still struggle to fully perceive fine-grained cues and lack information-efficient supervision. In this paper, we address these challenges through Focus-CoT and Focus-GRPO.

\noindent\textbf{Reasoning MLLMs.} Reinforcement learning~\cite{kaelbling1996reinforcement, zhai2025vlm_rl, guo2025deepseek, Shulman2017PPO, Rafailov2023DPO} offers a new way to optimize reasoning in MLLMs. R1-Onevision~\cite{yang2025r1onevision} and OpenVLThinker~\cite{deng2025openvlthinker} bootstrap by converting images into formalized text to generate cross-modal Chain-of-Thought (CoT)~\cite{wei2022cot} datasets and employ the GRPO~\cite{guo2025deepseek, shao2024deepseekmath} algorithm to generalize the model's reasoning abilities. R1-VL~\cite{zhang2025r1vl} and VisualPRM~\cite{wang2025visualprm} mitigate the sparse reward problem through key-step matching and a process-reward model, respectively. NoisyRollout~\cite{liu2025noisyrollout} and R1-ShareVL~\cite{yao2025r1_share} expand the training distribution through image perturbations and question transformations, enhancing the model's reasoning capabilities and robustness. However, these models fail to efficiently associate visual cues during reasoning, and the fixed KL penalty coefficient limits adaptive deep reasoning. In this paper, we introduce Focus-GRPO to optimize focusing efficiency and perform adaptive reasoning according to the number of visual cues.

\noindent\textbf{Chart Benchmarks.} Existing chart benchmarks~\cite{masry2022chartqa, wang2024charxiv, kahou2017figureqa, kafle2018dvqa, huang2025evochart, xu2023chartbench, methani2020plotqa} have greatly promoted research in chart understanding and reasoning. ChartQA~\cite{masry2022chartqa} combines human and machine-generated data, but covers only three chart types. ChartBench~\cite{xu2023chartbench} aims to evaluate complex visual reasoning but still lacks openness and diversity. EvoChart~\cite{huang2025evochart} focuses on real-world adaptation, collecting 650 real charts from various websites and designing 1250 human-annotated questions. CharXiv~\cite{wang2024charxiv} gathers 2323 charts from arXiv papers across various scientific disciplines, introducing descriptive and reasoning questions. In this paper, we construct HID-Chart based on our proposed information-density metric and compare it with existing chart benchmarks from multiple perspectives.
\begin{figure*}[t]
\centering
\includegraphics[width=\textwidth]{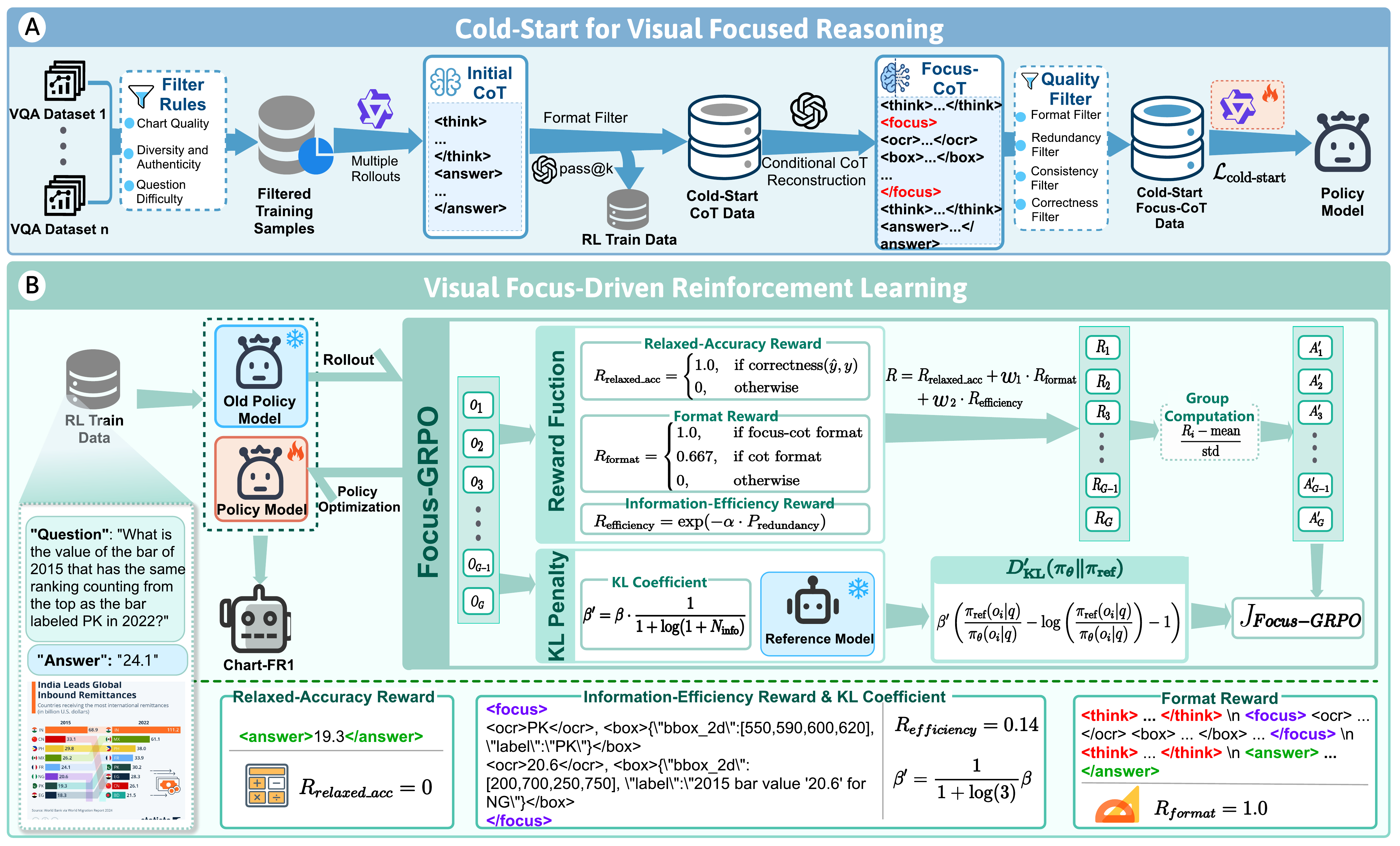}
\caption{
Overview of the two-stage focused reasoning training. Stage 1: High-quality Focus-CoT cold-start data is automatically constructed via a data generation pipeline, followed by supervised fine-tuning to activate the model’s visual focused reasoning ability. Stage 2: Focus-GRPO is introduced to further enhance fine-grained reasoning. The old policy $\pi_{\theta_{\mathrm{old}}}$ generates multiple reasoning paths, which are evaluated by three rewards—$R_{\text{relaxed\_acc}}$, $R_{\text{format}}$, and $R_{\text{efficiency}}$, encouraging accurate, well-structured reasoning and efficient focusing. The total reward guides advantage estimation via a group-relative method, and the KL penalty is adaptively adjusted according to the amount of visual cues, balancing reasoning depth and training stability. Finally, the policy $\pi_{\theta}$ is optimized with the objective $J_{\mathrm{Focus\text{-}GRPO}}$.
}
\label{img:main_method}
\end{figure*}

\section{Method}
We adopt a two-stage focused reasoning training framework to enhance the fine-grained reasoning capability of MLLMs in HID charts. We first introduce \textbf{Focus-CoT} in Sec.~\ref{sec:focuscot}, along with an automated data generation pipeline to construct a cold-start dataset. Building on this, we employ supervised fine-tuning to help the model learn to associate reasoning steps with key visual information. Subsequently, to alleviate the issues of reward sparsity, information redundancy, and excessive KL penalty observed in standard GRPO~\cite{guo2025deepseek} when applied to HID charts, we propose \textbf{Focus-GRPO} in Sec.~\ref{sec:focusgrpo} for second-stage training. More importantly, to facilitate future research on the fine-grained reasoning of HID charts, we establish a challenging benchmark, termed \textbf{HID-Chart}, as detailed in Sec.~\ref{sec:hidchart}.

\subsection{Cold-Start for Visual Focused Reasoning}
\label{sec:focuscot}
Chain-of-Thought (CoT)~\cite{wei2022cot} operates primarily at the linguistic level and lacks sufficient perception of visual cues. Also, Reinforcement Learning (RL)~\cite{Shulman2017PPO, guo2025deepseek} typically explores high-reward paths within the model’s existing knowledge. Therefore, we design an automated Focus-CoT generation pipeline and inject knowledge of visual focused reasoning into the model through supervised fine-tuning.

\noindent\textbf{Focus-CoT} introduces a \texttt{<focus>} tag to anchor reasoning steps to critical visual elements. Each focusing process involves two sub-actions: OCR text extraction and local image localization. Subsequent reasoning is conducted under the guidance of focused information, leading to a tight coupling between reasoning and perception. The automated Focus-CoT generation pipeline is shown in Fig.~\ref{img:main_method}(A).

\noindent\textbf{Data collection and CoT generation.}
We select training samples from existing chart datasets based on question difficulty, chart quality, diversity, and authenticity. For each sample, we apply CoT prompting to the base model Qwen2.5-VL~\cite{bai2025qwen} to generate eight reasoning paths.

\noindent\textbf{Format filtering and correctness evaluation.}
We filter out reasoning chains that do not conform to the initial CoT format and employ an LLM to evaluate the correctness of candidate paths. We calculate a pass@k metric for each sample. Based on pass@k, the samples are categorized as easy, medium, or hard, with ratios of 1:7:2 used to build the RL training set. The remaining easy and hard samples serve as the initial training set for the cold-start stage.

\noindent\textbf{Conditional CoT reconstruction.}
We leverage a powerful teacher model, GPT-5~\cite{openai2025gpt5}, to associate the initial CoT with crucial visual evidence. If the original reasoning is incorrect, the teacher model localizes the error, inserts a \texttt{<focus>} tag to gather the correct visual information, and revises the subsequent reasoning and answer accordingly. If the original reasoning is correct, the teacher model inserts a \texttt{<focus>} tag at key points and supplements the reasoning with a verification step supported by visual evidence, without altering the underlying logic. Through this process, we create valuable learning samples that provide key visual cues to guide correct fine-grained reasoning.

\noindent\textbf{Quality filtering.}
Finally, we employ both rule-based and LLM-based quality filters to conduct a final review of the generated Focus-CoT data, ensuring the correctness of answers and adherence to the format, while removing reasoning chains with redundant visual information.

In the cold-start stage, we follow the principle of preserving the base model’s general capabilities while introducing visual focused reasoning knowledge, thereby acquiring preliminary fine-grained reasoning abilities. We update the model parameters by minimizing the following loss:
\begin{equation}
\mathcal{L}_{\text{cold-start}} = - \mathbb{E}_{(x, q, r, a) \sim \mathcal{D}} \sum_{t=1}^{T} \log \pi_{\theta}(y_t \mid x, q, y_{<t}),
\label{eq:cold_start_loss}
\end{equation}
where $\mathcal{D}$ is the Focus-CoT dataset, $x$ is the image, $q$ is the question, $r$ represents the reasoning steps, $a$ is the final answer, and $y$ is the concatenated sequence of $r$ and $a$.

\subsection{Visual Focus-Driven Reinforcement Learning}
\label{sec:focusgrpo}
In the second stage of training, we propose the \textbf{Focus-GRPO} algorithm. Compared to standard GRPO, which mainly relies on the sparse reward of task accuracy as its supervision signal, Focus-GRPO introduces several key improvements: (1) \textbf{relaxed-accuracy reward} provides stable learning signals under ambiguous numerical answers; (2) \textbf{information-efficiency reward} penalizes redundant visual information, encouraging the model to focus on high-value cues; (3) \textbf{adaptive KL penalty} dynamically adjusts the exploration space to maintain a balance between reasoning depth and training stability during policy optimization.

An overview of the Focus-GRPO algorithm is shown in Fig.~\ref{img:main_method}(B). For each question $q$, the old policy model $\pi_{\theta_{\mathrm{old}}}$ samples a set of outputs $\{o_1, o_2, \dots, o_G\}$. We employ a multi-dimensional reward function to score each output $o_i$.

\textbf{Relaxed-Accuracy reward} is designed to evaluate the correctness of the model's prediction. To accommodate numerical fluctuation issues in chart QA tasks, we design a relaxed reward environment. Let $\hat{y}$ be the predicted answer and $y$ be the ground truth. The reward is defined as:
\begin{equation}
R_{\text{relaxed\_acc}} =
\begin{cases}
    1.0, & \text{if correctness}(\hat{y}, y), \\
    0, & \text{otherwise},
\end{cases}
\label{eq:relaxed_acc_reward}
\end{equation}
where the behavior of the $\text{correctness}(\hat{y}, y)$ function depends on the type of answer:
\begin{equation}
\text{correctness}(\hat{y}, y) =
\begin{cases}
    \frac{|\hat{y} - y|}{\max(|y|, \mu)} \leq 0.05, & \text{if numerical}, \\
    \hat{y} = y, & \text{otherwise},
\end{cases}
\label{eq:correctness_func}
\end{equation}
where $\mu$ is a small constant to prevent division by zero.

\textbf{Format reward} $R_{\text{format}}$ encourages the model to generate structured and interpretable Focus-CoT. This reward uses regular expressions to match the structure of the output, assigning different format scores: 1.0 for the Focus-CoT format, 0.667 for the CoT format, and 0 for others.

\textbf{Information-Efficiency reward} measures the efficiency of visual focusing action, which suppresses the redundancy of OCR information and local image regions in the \texttt{<focus>} tag, and enables the discovery of high-value visual cues. This reward is defined as an exponential decay function based on a redundancy penalty $P_{\text{redundancy}}$:
\begin{equation}
R_{\text{efficiency}} = \exp(-\alpha \cdot P_{\text{redundancy}}),
\label{eq:efficiency_reward}
\end{equation}
where $\alpha$ is a hyperparameter to control the decay rate. Let $\mathcal{T}=\{t_i\}$, $\mathcal{B}=\{b_j\}$, and $\mathcal{L}=\{l_k\}$ represent the OCR text set, bounding box set, and bounding box label set, respectively. These visual elements can be extracted from the \texttt{<ocr>} and \texttt{<box>} tags using regular expressions. The redundancy penalty is composed of three sub-components:

\begin{itemize}
    \item \textbf{OCR-OCR redundancy.} Calculates the text similarity between OCR texts $t_i, t_j \in \mathcal{T}$. The similarity $\text{sim}(\cdot, \cdot)$ is computed using the SequenceMatcher algorithm.
    \begin{equation}
    P_{tt} = \underset{i \neq j}{\text{avg}}\left\{\text{sim}(t_i, t_j) \mid \text{sim}(t_i, t_j) > \tau\right\}.
    \label{eq:p_tt}
    \end{equation}
    \item \textbf{Box-Box redundancy.} Calculates the Intersection over Union (IoU) between bounding boxes $b_i, b_j \in \mathcal{B}$.
    \begin{equation}
    P_{bb} = \underset{i \neq j}{\text{avg}}\left\{\text{IoU}(b_i, b_j)\right\}.
    \label{eq:p_bb}
    \end{equation}
    \item \textbf{OCR-Box redundancy.} Calculates the maximum text similarity between each OCR text $t_i \in \mathcal{T}$ and all bounding box labels $l_k \in \mathcal{L}$.
    \begin{equation}
    P_{tb} = \underset{t_i \in \mathcal{T}}{\text{avg}}\left\{\max_{l_k \in \mathcal{L}}\text{sim}(t_i, l_k) \mid \max_{l_k \in \mathcal{L}}\text{sim}(t_i, l_k) > \tau\right\},
    \label{eq:p_tb}
    \end{equation}
\end{itemize}
where $\tau$ is a similarity threshold to avoid unnecessary penalties. The final redundancy penalty $P_{\text{redundancy}}$ is the average of all detected redundancy terms.

In the end, the total reward is calculated as:
\begin{equation}
R = R_{\text{relaxed\_acc}} + w_{1} \cdot R_{\text{format}} + w_{2} \cdot R_{\text{efficiency}},
\label{eq:total_reward}
\end{equation}
where $w_1$ and $w_2$ are weight coefficients.

\textbf{Adaptive KL penalty.} RL commonly uses a fixed KL penalty coefficient $\beta$ to constrain the policy $\pi_\theta$ from deviating excessively from a reference policy $\pi_{\text{ref}}$. However, in HID charts, certain reasoning steps require deep exploration to discover the optimal path and a static $\beta$ overly restricts it. Therefore, we dynamically adjust $\beta$ based on the amount of focused information. Specifically, when the model attends to rich cues for deep reasoning, we relax the KL constraint to encourage exploration. When reasoning relies on limited cues, we tighten the constraint to ensure policy stability. Focused information is quantified as $N_{\text{info}} = (N_{\text{ocr}} + N_{\text{box}}) / 2$, and the adaptive penalty coefficient $\beta'$ is defined as:

\begin{equation}
\beta' = \beta \cdot \frac{1}{1 + \log(1 + N_{\text{info}})}.
\label{eq:adaptive_beta}
\end{equation}
The adaptive KL penalty term is expressed as:
\begin{equation}
D'_{\text{KL}}(\pi_{\theta} \| \pi_{\text{ref}}) = \beta' \left( \frac{\pi_{\text{ref}}(o_i \mid q)}{\pi_{\theta}(o_i \mid q)} - \log \left( \frac{\pi_{\text{ref}}(o_i \mid q)}{\pi_{\theta}(o_i \mid q)} \right) - 1 \right).
\label{eq:adaptive_kl}
\end{equation}
Finally, Focus-GRPO maximizes the following objective:
\begin{equation}
\small
\begin{aligned}
J_{\mathrm{Focus-GRPO}}(\theta) 
&= \mathbb{E}_{q \sim P(Q),\, \{o_i\}_{i=1}^G \sim \pi_{\theta_{\mathrm{old}}}(O \mid q)} \\
&\Biggl[
  \frac{1}{G} \sum_{i=1}^G
  \min\!\Bigl(
    \frac{\pi_{\theta}(o_i \mid q)}{\pi_{\theta_{\mathrm{old}}}(o_i \mid q)}\,A'_i,\, \\
    &\mathrm{clip}\!\Bigl(
      \frac{\pi_{\theta}(o_i \mid q)}{\pi_{\theta_{\mathrm{old}}}(o_i \mid q)},
      1-\varepsilon,\,
      1+\varepsilon
    \Bigr)
    A'_i
  \Bigr)  \\
  &-\,D'_{\mathrm{KL}}\bigl(\pi_{\theta}\,\big\|\,\pi_{\mathrm{ref}}\bigr)
\Biggr],
\end{aligned}
\label{eq1}
\end{equation}
where $\varepsilon$ is the PPO clipping hyperparameter and $A'_i$ is the computed advantage using the group rewards $\{R_j\}_{j=1}^G$:
\begin{equation}
A'_i = \frac{R_i - \text{mean}(\{R_1, R_2, \dots, R_G\})}{\text{std}(\{R_1, R_2, \dots, R_G\})}.
\label{eq:advantage}
\end{equation}

\subsection{HID-Chart Benchmark}
\label{sec:hidchart}

To systematically quantify the complexity of charts, we propose an information-density metric, \textbf{Chart-ID}. We employ an advanced MLLM, GPT-5~\cite{openai2025gpt5}, to score charts along four dimensions, each scored on a five-point scale: (1) \textbf{Information richness}: the diversity of information, such as the number of data series, dimensions, and annotations. (2) \textbf{Information efficiency}: the amount of effective information conveyed by a chart. (3) \textbf{Information clarity}: visual readability of the chart, such as the clarity of legends and labels. (4) \textbf{Information interactivity}: the ease of interaction between users and charts. The final score is defined as:
\begin{equation}
\text{Chart-ID} = \frac{S_{\text{rich}}}{2} + \frac{S_{\text{eff}}}{5} + \frac{S_{\text{clar}}}{5} + \frac{S_{\text{inter}}}{10}.
\label{eq:chart_id}
\end{equation}

The construction of HID-Chart follows a human-in-the-loop pipeline: (1) \textbf{Data curation}: We first collect approximately 2500 charts from scientific and social science publications, websites, public data visualization libraries, and industry reports from 2023 to 2025. This initial corpus ensures diversity in chart types and subject domains. (2) \textbf{Chart filtering}: We calculate the Chart-ID score for each chart, retaining only those with HID to form a high-quality candidate pool. (3) \textbf{Question generation}: For the selected charts, we employ GPT-5 to generate multiple candidate questions that emphasize diversity, complexity and realism, aiming to effectively evaluate the fine-grained reasoning ability of the model. (4) \textbf{Human annotation and rewriting}: A team of five graduate students completes the following tasks: (a) removing overly simple or ambiguous questions; (b) upgrading some single-step questions to complex ones requiring multi-step reasoning and information integration; (c) performing initial answer annotations for each question; (d) exchanging questions among annotators for the second annotation to ensure correctness. As shown in Table~\ref{tab:bench_parameter}, the final HID-Chart comprises 1561 high-quality QA pairs and 734 charts. We compare our benchmark with other chart benchmarks, as shown in Table~\ref{tab:benchmark_comparison}. HID-Chart features an average information density of 3.94, covering 10 chart types and 8 domains. This benchmark sets a new standard for evaluating the fine-grained reasoning capabilities of MLLMs in HID chart scenarios.

\begin{table}[t]
    \centering
    \renewcommand{\arraystretch}{0.9}
    \begin{tabular}{@{}l r@{}}
    \toprule
    \textbf{Statistics} & \textbf{Value} \\
    \midrule
    Total charts & 734 \\
    Average information density & 3.94 \\
    Domains / Chart types & 8 / 10\\
    Average size (px) & 1090 × 796 \\
    Maximum size (px) & 2487×1716 \\
    \midrule
    Number of unique questions & 1561 \\
    Number of unique question tokens & 7341 \\
    Average question length & 20.9 \\
    Maximum question length & 43 \\
    \midrule
    Number of unique answer tokens & 1795 \\
    Average answer length & 1.9 \\
    Maximum answer length & 27 \\
    \bottomrule
    \end{tabular}
    \caption{Key statistics of HID-Chart.}
    \vspace{-0.5em}
    \label{tab:bench_parameter}
\end{table}

\begin{table}[t]
\centering
\resizebox{\columnwidth}{!}{%
\begin{tabular}{@{}l|ccc@{}}
\toprule
\textbf{Benchmark} & \textbf{Information Density} & \textbf{Chart Type} & \textbf{Domain} \\ \midrule
ChartQA~\cite{masry2022chartqa} & 3.23 & 3 & 4 \\
CharXiv~\cite{wang2024charxiv} & 3.75 & - & \textbf{8} \\
EvoChart~\cite{huang2025evochart} & 3.56 & 4 & - \\
ChartBench~\cite{xu2023chartbench} & 3.19 & 9 & - \\
PlotQA~\cite{methani2020plotqa} & 3.10 & 3 & 3 \\ \midrule
\textbf{HID-Chart} & \textbf{3.94} & \textbf{10} & \textbf{8} \\ \bottomrule
\end{tabular}%
}
\caption{Comparison of HID-Chart with other chart benchmarks.}
\vspace{-1.0em}
\label{tab:benchmark_comparison}
\end{table}

\begin{table*}[t]
\footnotesize
\centering
\renewcommand{\arraystretch}{0.9}
\begin{tabularx}{\textwidth}{
    p{0.22\linewidth} 
    *{6}{|>{\centering\arraybackslash}X}
}
\toprule[1.pt]
\textbf{Model} & \textbf{ChartQA} & \textbf{CharXiv} & \textbf{EvoChart} & \textbf{ChartBench} & \textbf{PlotQA} & \textbf{Avg} \\
\midrule

\multicolumn{7}{c}{\emph{\textbf{Closed-Source MLLMs}}} \\ 
\midrule
Gemini-1.5-Pro~\cite{team2024gemini} & 87.2 & 43.3 & - & - & - & - \\
GPT-4o~\cite{hurst2024gpt4o} & 85.7 & \high{\textbf{47.1}} & \high{\textbf{63.9}} & 72.3 & 51.0 & 64.0 \\

\midrule

\multicolumn{7}{c}{\emph{\textbf{General MLLMs}}} \\ 
\midrule
Qwen2.5-VL-7B~\cite{bai2025qwen} (base) & 87.3 & 42.5 & 53.5 & 66.4 & 55.5 & 61.0 \\
LLaVA-Onevision-7B~\cite{li2025llavaonevision} & 80.6 & 26.7 & 43.6 & 60.8 & 51.4 & 52.6 \\
Keye-VL-8B~\cite{kwai2025keye} & 72.5 & 36.8 & 54.6 & 70.6 & 56.1 & 58.1 \\
DeepSeek-VL2~\cite{wu2024deepseekvl2} & 86.0 & 38.5 & 50.9 & 59.5 & 41.0 & 55.2 \\
Gemma-3-27B~\cite{gemma2025gemma3} & 78.0 & 25.7 & 43.6 & 69.0 & 48.0 & 52.9 \\
InternVL2.5-78B~\cite{chen2024internvl2_5} & 88.3 & 42.4 & 61.2 & 75.5 & 62.0 & 65.9 \\
\midrule

\multicolumn{7}{c}{\emph{\textbf{Chart MLLMs}}} \\ 
\midrule
ChartReasoner~\cite{jia2025chartreasoner} & 86.9 & - & 48.1 & 55.2 & - & - \\
EvoChart~\cite{huang2025evochart} & 81.5 & 31.4 & 54.2 & 67.6 & 46.2 & 56.2 \\
ChartGemma~\cite{masry2024chartgemma} & 80.2 & 9.0 & 30.6 & 47.5 & 20.0 & 37.5 \\
ChartPoint-7B~\cite{xu2025chartpoint} & 87.7 & - & - & 66.0 & - & - \\
ChartSketcher-72B~\cite{huang2025chartsketcher} & 88.9 & 36.6 & 63.3 & 68.3 & 57.1 & 62.8 \\
\midrule

\multicolumn{7}{c}{\emph{\textbf{General Reasoning MLLMs}}} \\ 
\midrule
R1-VL-7B~\cite{zhang2025r1vl} & 83.9 & 33.6 & 49.8 & 60.6 & 59.0 & 57.4 \\
R1-Onevision-7B~\cite{yang2025r1onevision} & 85.1 & 33.0 & 48.2 & 67.8 & 58.2 & 58.5 \\
OpenVLThinker-7B~\cite{deng2025openvlthinker} & 85.5 & 40.2 & 57.2 & 72.6 & 57.5 & 62.6 \\
Vision-R1-7B~\cite{huang2025visionr1} & 84.0 & 38.7 & 54.0 & 66.3 & 58.3 & 60.3 \\
\midrule

\multicolumn{7}{c}{\emph{\textbf{Our Model}}} \\ 
\midrule
Chart-FR1-7B & \high{\textbf{91.0}} & 46.6 & 59.2 & \high{\textbf{75.6}} & \high{\textbf{62.9}} & \high{\textbf{67.1}} \\
\bottomrule[1.pt]
\end{tabularx}

\caption{Results on five chart understanding and reasoning benchmarks. The table presents a comprehensive comparison with SOTA MLLMs (including closed-source, general, chart, and reasoning MLLMs) to verify the effectiveness of the two-stage focused reasoning training framework. The highest accuracy for each benchmark is marked in \colorbox{backblue}{blue}.}

\label{tab:results1}
\end{table*}

\begin{table*}[t]
\centering
\renewcommand{\arraystretch}{1.1} 

\resizebox{\textwidth}{!}{%
\begin{tabular}{l|c|cccc|*{8}{p{1.3cm}<{\centering}}}
\toprule
\multirow{2}{*}{\textbf{Model}} 
& \multirow{2}{*}{\textbf{Avg}} 
& \multicolumn{4}{|c|}{\textbf{Information Density Interval}} 
& \multicolumn{8}{|c}{\textbf{Category}} \\
\cmidrule(lr){3-6} \cmidrule(lr){7-14}
& & {[0.0,3.7)} & {[3.7,3.9)} & {[3.9,4.2)} & {[4.2,5.0]} 
& Physics & Chemistry & Biology & Math & CS & Engineer & Social & Finance \\
\midrule

\multicolumn{14}{l}{\textbf{\textit{Closed-Source MLLMs}}} \\ \midrule
Gemini-2.0-Flash~\cite{deepmind2024gemini} & 56.8 & 62.7 & 59.5 & 57.3 & 50.7 & 54.6 & 54.6 & 55.1 & 52.6 & 58.5 & 58.6 & 57.0 & 57.5 \\
GPT-4o~\cite{hurst2024gpt4o} & 51.2 & 61.0 & 53.0 & 50.2 & 48.5 & 48.7 & 54.6 & 45.6 & 48.5 & 55.1 & 54.3 & 51.3 & 49.2 \\
GPT-4.1~\cite{fachada2025gpt41} & \highg{\textbf{63.7}} & \highg{\textbf{69.5}} & \highg{\textbf{65.0}} & \highg{\textbf{63.6}} & \highg{\textbf{60.9}} & \highg{\textbf{59.7}} & \highg{\textbf{65.7}} & \highg{\textbf{61.0}} & \highg{\textbf{67.0}} & \highg{\textbf{73.7}} & \highg{\textbf{61.2}} & \highg{\textbf{59.4}} & \highg{\textbf{59.9}} \\
\midrule

\multicolumn{14}{l}{\textbf{\textit{Open-Source MLLMs}}} \\ \midrule
EvoChart~\cite{huang2025evochart} & 29.3 & 40.7 & 31.9 & 29.1 & 23.6 & 28.6 & 30.6 & 26.5 & 22.7 & 31.6 & 27.6 & 27.8 & 32.1 \\
ChartSketcher-72B~\cite{huang2025chartsketcher} & 40.0 & 46.4 & 42.9 & 40.1 & 34.6 & 38.7 & 41.7 & 42.6 & 37.1 & 44.6 & 33.6 & 37.3 & 40.4 \\
R1-VL-7B~\cite{zhang2025r1vl} & 34.5 & 44.1 & 36.4 & 34.2 & 30.3 & 30.3 & 50.0 & 30.1 & 29.9 & 29.1 & 31.0 & 34.9 & 40.1 \\
OpenVLThinker-7B~\cite{deng2025openvlthinker} & 38.5 & 42.4 & 38.8 & 38.3 & 37.8 & 38.7 & 38.9 & 36.0 & 39.2 & 41.2 & 35.3 & 37.9 & 38.2 \\
Vision-R1-7B~\cite{huang2025visionr1} & 40.6 & 52.5 & 41.9 & 39.2 & 38.9 & 39.5 & 47.2 & 33.8 & 43.3 & 42.1 & 40.5 & 40.0 & 40.1 \\
LLaVA-Onevision-7B~\cite{li2025llavaonevision} & 28.6 & 33.9 & 30.4 & 29.3 & 24.1 & 25.2 & 34.3 & 25.0 & 25.8 & 28.2 & 24.1 & 30.7 & 30.3 \\
Gemma-3-27B~\cite{gemma2025gemma3} & 35.6 & 49.2 & 37.0 & 35.5 & 31.4 & 35.3 & 38.9 & 34.6 & 36.1 & 36.2 & 44.0 & 34.9 & 32.1 \\
Keye-VL-8B~\cite{kwai2025keye} & 40.0 & 47.5 & 41.5 & 39.4 & 37.3 & 36.1 & 46.3 & 27.9 & 36.1 & 44.9 & 37.9 & 38.8 & 42.5 \\
InternVL3-8B~\cite{zhu2025internvl3} & 40.9 & 44.1 & 41.7 & 40.5 & 39.9 & 37.8 & 46.3 & 37.5 & 37.1 & 44.0 & 44.0 & 41.5 & 38.2 \\
InternVL2.5-78B~\cite{chen2024internvl2_5} & 42.7 & 50.8 & 43.4 & 41.6 & 42.1 & 42.0 & 49.1 & 36.0 & 42.3 & 47.1 & 45.7 & 37.6 & 43.7 \\
Qwen2.5-VL-7B~\cite{bai2025qwen} (base) & 43.0 & 52.5 & 45.8 & 40.8 & 40.2 & 43.7 & 49.1 & 37.5 & 41.2 & 43.7 & 44.8 & 41.5 & 43.7 \\
Qwen2.5-VL-72B~\cite{bai2025qwen} & 51.5 & 55.9 & 53.5 & 50.9 & 48.5 & \high{\textbf{53.8}} & \high{\textbf{53.7}} & 42.6 & 47.4 & 57.3 & 47.4 & \high{\textbf{52.2}} & 49.8 \\
\midrule

Chart-FR1-7B (ours) & \high{\textbf{53.0}} & \high{\textbf{57.6}} & \high{\textbf{54.4}} & \high{\textbf{52.0}} & \high{\textbf{51.5}} & 52.9 & 50.0 & \high{\textbf{51.5}} & \high{\textbf{50.5}} & \high{\textbf{60.4}} & \high{\textbf{49.1}} & 51.9 & \high{\textbf{50.5}} \\
\bottomrule
\end{tabular}
}
\centering
\caption{Evaluation Results on the HID-Chart Benchmark. The table presents the performance of Chart-FR1 against closed-source and open-source SOTA MLLMs. The results are categorized into four information density intervals and eight subject categories. The highest accuracy of the closed-source model is highlighted in \colorbox{backgreen}{green}, and the best result among the remaining models is highlighted in \colorbox{backblue}{blue}.}

\vspace{-0.3cm}
\label{tab:results2}
\end{table*}

\section{Experiment}
\label{sec:experiment}

\subsection{Experiment Settings}

\paragraph{Dataset}
To construct the dataset for the cold-start and Focus-GRPO stages, we first collate samples from several established chart datasets, including NovaChart~\cite{hu2024novachart}, EvoChart~\cite{huang2025evochart}, and ChartQA~\cite{masry2022chartqa}. We then apply the data generation pipeline introduced in Sec.~\ref{sec:focuscot}. The final cold-start dataset contains 6.4k samples, and the Focus-GRPO stage comprises 30k training samples.

\paragraph{Baselines and Benchmarks}
To comprehensively evaluate the effectiveness of our method, we compare it against advanced MLLMs, categorized into four groups: closed-source MLLMs, general MLLMs, chart MLLMs, and general reasoning MLLMs. We evaluate these MLLMs on five popular chart benchmarks: ChartQA~\cite{masry2022chartqa}, CharXiv~\cite{wang2024charxiv}, EvoChart~\cite{huang2025evochart}, ChartBench~\cite{xu2023chartbench}, and PlotQA~\cite{methani2020plotqa}. For CharXiv, we use the CharXiv-RQ subset. For ChartBench, we use the test mini split, consisting of 1890 samples. For PlotQA, we randomly sample 2000 test data using seed 42.

\paragraph{Implementation details}
We use Qwen2.5-VL-7B as the base model. All experiments are conducted on 8 NVIDIA H100 GPUs. In the first stage, to avoid overfitting, we train for 1 epoch on the Focus-CoT dataset with a learning rate of $2 \times 10^{-6}$ and a global batch size of 256. In the second stage, we use Focus-GRPO to train on the RL dataset for 3 epochs with a learning rate of $1 \times 10^{-6}$, a global batch size of 512, and 8 rollouts. We set $\alpha$ to 2, $\tau$ to 0.9, and $\beta$ to $1 \times 10^{-2}$. $w_1$ and $w_2$ are set to 0.1 to ensure training stability. For further details, see the supplementary material.

\subsection{Main Results}

\paragraph{Result 1: two-stage focused reasoning training significantly enhances MLLM's performance on chart understanding and reasoning} 
As shown in Table~\ref{tab:results1}, Chart-FR1 achieves outstanding performance in five chart benchmarks, outperforming Qwen2.5-VL-7B by an average margin of 6.1\%. Chart-FR1 also demonstrates the best average performance among all categories of MLLMs, even surpassing the closed-source GPT-4o by 3.1\%. This outcome demonstrates the superiority of Focus-CoT and Focus-GRPO.

\paragraph{Result 2: two-stage focused reasoning training significantly improves MLLM’s performance on HID charts}
As shown in Table~\ref{tab:results2}, Chart-FR1 is the best among open-source MLLMs, outperforming Qwen2.5-VL-7B by 10.0\%, surpassing the larger Qwen2.5-VL-72B by 1.5\%, and exceeding the closed-source GPT-4o by 1.8\%. The results also show that model performance continues to degrade as information density increases, revealing the challenge of MLLMs on HID charts and the value of our method.

\paragraph{Result 3: Focus-GRPO effectively addresses the limitations of GRPO when handling HID charts}
As shown in Table~\ref{tab:focus_grpo_ablation}, Focus-GRPO improves the average performance by 3\% over GRPO when the two algorithms are applied in the second training stage. This demonstrates that Focus-GRPO effectively mitigates issues of sparse rewards, redundant information, and over-penalization in HID chart scenarios.

\begin{table}[t]
\centering
{\normalsize
\resizebox{1.0\linewidth}{!}{
\begin{tabular}{l|cccccc}
\toprule
\textbf{Method} & \textbf{ChartQA} & \textbf{CharXiv} & \textbf{EvoChart} & \textbf{ChartBench} & \textbf{PlotQA} & \textbf{Avg} \\
\midrule
GRPO
& 89.6 & 40.9 & 58.5 & 70.6 & 61.1 & 64.1 \\
\midrule
Focus-GRPO
& \high{\textbf{91.0}} & \high{\textbf{46.6}} & 59.2 & \high{\textbf{75.6}} & \high{\textbf{62.9}} & \high{\textbf{67.1}} \\
w/o Adaptive KL Penalty
& 90.2 & 44.5 & \high{\textbf{59.3}} & 72.9 & 62.3 & 65.8 \\
w/o $R_{\text{efficiency}}$
& 90.3 & 43.0 & 59.0 & 74.0 & 62.6 & 65.8 \\
w/o Adaptive KL Penalty \& $R_{\text{efficiency}}$
& 90.5 & 43.0 & 58.9 & 73.4 & 61.8 & 65.5 \\
\bottomrule
\end{tabular}
}
}
\caption{Performance comparison between GRPO and Focus-GRPO, along with an ablation study of Focus-GRPO components.}
\label{tab:focus_grpo_ablation}
\end{table}

\begin{table}[t]
\centering
{\normalsize
\resizebox{1.0\linewidth}{!}{
\begin{tabular}{l|cccccc}
\toprule
\textbf{Model} & \textbf{ChartQA} & \textbf{CharXiv} & \textbf{EvoChart} & \textbf{ChartBench} & \textbf{PlotQA} & \textbf{Avg} \\
\midrule
Qwen2.5-VL-7B
& 87.3 & 42.5 & 53.5 & 66.4 & 55.5 & 61.0 \\
\midrule
Chart-FR1-7B
& \high{\textbf{91.0}} & \high{\textbf{46.6}} & 59.2 & \high{\textbf{75.6}} & \high{\textbf{62.9}} & \high{\textbf{67.1}} \\
w/o Focus-GRPO
& 87.6 & 40.8 & 54.8 & 71.6 & 58.5 & 62.7 \\
w/o Cold-Start
& 90.0 & 42.0 & 57.7 & 72.3 & 61.5 & 64.7 \\
w/o OCR
& 89.6 & 42.5 & 59.0 & 71.5 & 60.1 & 64.5 \\
w/o box
& 89.9 & 43.2 & \high{\textbf{59.7}} & 72.4 & 60.6 & 65.2 \\
\bottomrule
\end{tabular}
}
}
\caption{Ablation study of the two-stage training framework and the cues focused by Focus-CoT.}
\label{tab:framework_ablation}
\end{table}

\begin{table}[t]
\centering
{\normalsize
\resizebox{1.0\linewidth}{!}{
\begin{tabular}{l|ccccccc}
\toprule
\textbf{Model} & \textbf{ChartQA} & \textbf{CharXiv} & \textbf{EvoChart} & \textbf{ChartBench} & \textbf{PlotQA} & \textbf{HID-Chart} & \textbf{Avg} \\
\midrule
Qwen2.5-VL-3B 
& 84.0 & 31.3 & 39.3 & 53.1 & 45.5 & 33.2 & 47.7 \\
Chart-FR1-3B 
& \high{\textbf{88.5}} & \high{\textbf{35.7}} & \high{\textbf{52.1}} & \high{\textbf{69.2}} & \high{\textbf{56.6}} & \high{\textbf{38.2}} & \high{\textbf{56.7}} \\
\midrule
Qwen3-VL-8B 
& 89.6 & 46.4 & 66.9 & 65.0 & 63.4 & 52.3 & 63.9 \\
Chart-FR1-8B 
& \high{\textbf{92.5}} & \high{\textbf{56.2}} & \high{\textbf{72.0}} & \high{\textbf{76.7}} & \high{\textbf{65.1}} & \high{\textbf{55.5}} & \high{\textbf{69.7}} \\
\bottomrule
\end{tabular}
}
}
\caption{The generalization of our method across different models.}
\label{tab:method_generalization}
\end{table}

\begin{table}[t]
\centering
{\normalsize
\resizebox{1.0\linewidth}{!}{
\begin{tabular}{l|cccccc}
\toprule
\textbf{Teacher Model} & \textbf{ChartQA} & \textbf{CharXiv} & \textbf{EvoChart} & \textbf{ChartBench} & \textbf{PlotQA} & \textbf{Avg} \\
\midrule
Qwen2.5-VL-72B
& 90.2 & 44.7 & 57.6 & 73.4 & 61.4 & 65.5 \\
\midrule
Qwen3-VL-32B
& 89.9 & \high{\textbf{47.9}} & \high{\textbf{59.8}} & 74.8 & 61.0 & 66.7 \\
\midrule
GPT-5
& \high{\textbf{91.0}} & 46.6 & 59.2 & \high{\textbf{75.6}} & \high{\textbf{62.9}} & \high{\textbf{67.1}} \\
\bottomrule
\end{tabular}
}
}
\caption{The impact of teacher models on model performance.}
\label{tab:teacher_model_comparison}
\end{table}

\paragraph{Result 4: Focus-CoT outperforms CoT on HID charts}
As shown in Table~\ref{tab:zero_shot_result}, Qwen2.5-VL-7B zero-shot with Focus-CoT prompt outperforms CoT prompt by 0.4\% on CharXiv and 2.3\% on HID-Chart, showing the potential of Focus-CoT before training. Comparisons in Table~\ref{tab:results1} and ~\ref{tab:results2} against reasoning MLLMs confirm that two-stage focused reasoning training amplifies the advantages of Focus-CoT.

\begin{table}[t]
\setlength{\tabcolsep}{12pt}
\centering
\scriptsize
\begin{tabular}{lcc}
\toprule
\textbf{Qwen2.5-VL-7B} & \textbf{CharXiv} & \textbf{HID-Chart} \\
\midrule
CoT prompt & 42.6 & 42.5 \\
Focus-CoT prompt & \high{\textbf{43.0}} & \high{\textbf{44.8}} \\
\bottomrule
\end{tabular}
\caption{Zero-Shot comparison of Focus-CoT and CoT.}
\label{tab:zero_shot_result}
\end{table}

\begin{figure}
    \centering
    \includegraphics[width=\linewidth]{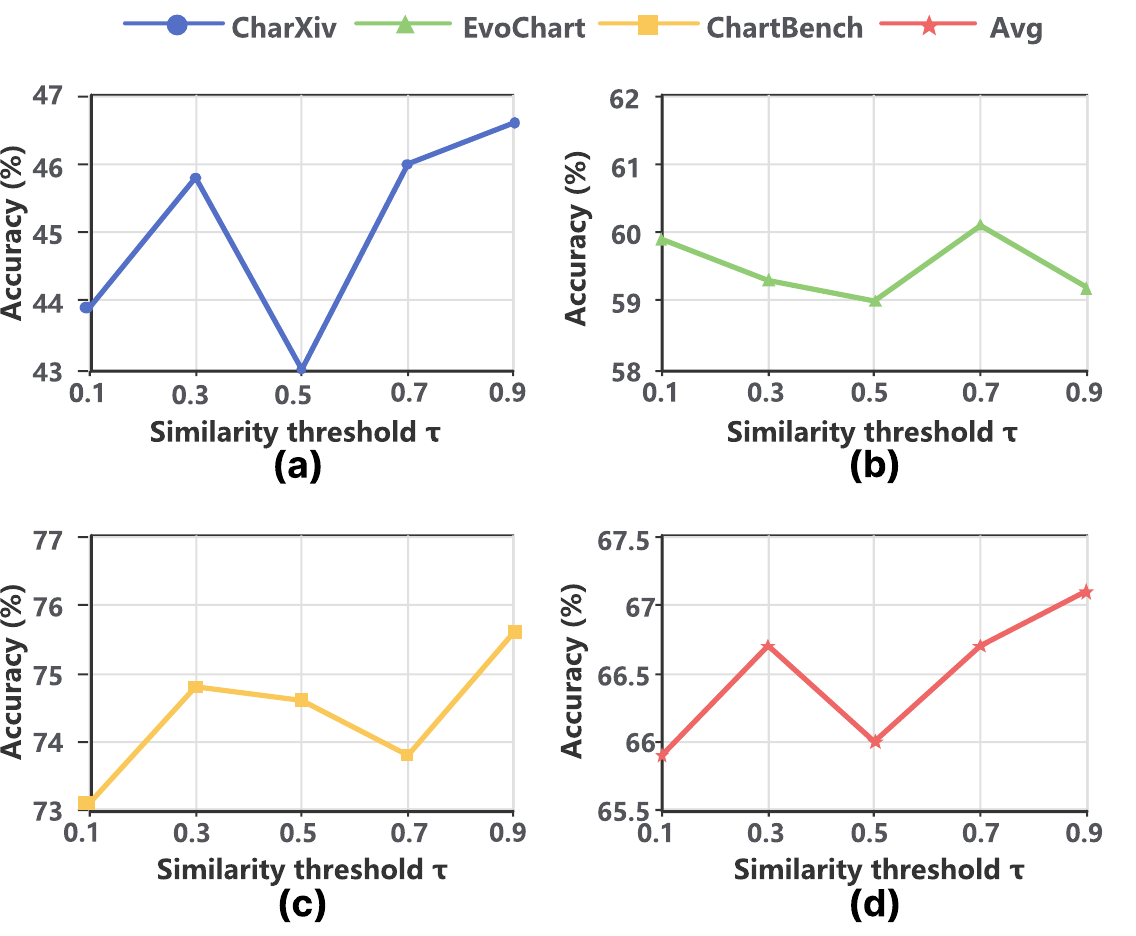}
    \caption{The impact of similarity threshold $\tau$}
    \label{img:line}
\end{figure}

\subsection{Ablation Study}

\paragraph{Effectiveness of Focus-GRPO components}
As shown in Table~\ref{tab:focus_grpo_ablation}, excluding the adaptive KL penalty causes a 1.3\% drop, indicating that the model is excessively penalized during policy optimization when deep reasoning relies on many focused cues. Removing the information-efficiency reward leads to a 1.3\% decline, suggesting that redundant information degrades reasoning accuracy. Removing both still outperforms GRPO by 1.4\%, showing that relaxed-accuracy reward provides a more stable learning environment.

\paragraph{Effectiveness of the two-stage focused reasoning training}
As shown in Table~\ref{tab:framework_ablation}, removing the cold-start stage leads to a 2.4\% drop, indicating that this stage initially activates the model's visual focused reasoning capabilities. Removing Focus-GRPO results in a 4.4\% decrease, showing that Focus-GRPO improves visual focus efficiency and enables adaptive deep reasoning.

\paragraph{Impact of the cues focused by Focus-CoT}
As shown in Table~\ref{tab:framework_ablation}, removing OCR texts during focusing causes a 2.6\% drop, while removing boxes leads to a 1.9\% decrease, showing that both OCR text and local regions are visual cues that the model should attend to during focused reasoning.

\paragraph{Generalization of our method}
As shown in Table~\ref{tab:method_generalization}, our method exhibits excellent performance when applied to Qwen2.5-VL-3B and Qwen3-VL-8B. This demonstrates the strong generalization and robustness of our method.

\paragraph{Impact of the teacher model}
Table \ref{tab:teacher_model_comparison} shows stable performance when replacing GPT-5 with open-source MLLMs. This indicates that our method relies on the focus-driven CoT mechanism rather than on a specific teacher model.

\paragraph{Impact of different similarity thresholds $\tau$}
As shown in Fig.~\ref{img:line}, the similarity threshold $\tau$ affects the model's performance and $\tau=0.9$ yields the highest average accuracy in five benchmarks, suggesting that penalizing highly similar visual information effectively prevents redundancy.

\section{Conclusion}
\label{conclusion}

In this paper, we propose Focus-CoT, a novel chain-of-thought paradigm that explicitly aligns reasoning with visual cues, effectively enhancing the fine-grained reasoning capabilities of MLLMs on HID charts. Furthermore, we introduce Focus-GRPO, which suppresses redundant information through an information-efficiency reward and employs an adaptive KL penalty to dynamically adjust the constraint on reasoning depth according to the number of focused cues, enabling efficient focusing and adaptive fine-grained reasoning. To address the lack of benchmarks for HID charts, we construct HID-Chart, which reveals the challenges of MLLMs and promotes research in this domain. We develop Chart-FR1 through two-stage focused reasoning training. Extensive experiments demonstrate that Chart-FR1 significantly outperforms advanced MLLMs.

\section*{Acknowledgment}

This work was supported by the National Key R\&D Program of China under Grant No. 2024YFB4505500 and No. 2024YFB4505503, the National Natural Science Foundation of China (No. 62302435 and No. 62132017), and the Zhejiang Provincial Natural Science Foundation of China under Grant No. LQ24F020006.

{
    \small
    \bibliographystyle{ieeenat_fullname}
    \bibliography{main}

@String(AAAI = {AAAI})

@article{bai2025qwen,
  title={Qwen2.5-VL Technical Report},
  author={Bai, Shuai and Chen, Keqin and Liu, Xuejing and Wang, Jialin and Ge, Wenbin and Song, Sibo and Dang, Kai and Wang, Peng and Wang, Shijie and Tang, Jun and Zhong, Humen and Zhu, Yuanzhi and Yang, Mingkun and Li, Zhaohai and Wan, Jianqiang and Wang, Pengfei and Ding, Wei and Fu, Zheren and Xu, Yiheng and Ye, Jiabo and Zhang, Xi and Xie, Tianbao and Cheng, Zesen and Zhang, Hang and Yang, Zhibo and Xu, Haiyang and Lin, Junyang},
  journal={arXiv preprint arXiv:2502.13923},
  year={2025}
}

@article{guo2025deepseek,
  title={DeepSeek-R1 incentivizes reasoning in LLMs through reinforcement learning},
  author={Guo, Daya and Yang, Dejian and Zhang, Haowei and Song, Junxiao and Wang, Peiyi and Zhu, Qihao and Xu, Runxin and Zhang, Ruoyu and Ma, Shirong and Bi, Xiao and others},
  journal={Nature},
  volume={645},
  number={8081},
  pages={633--638},
  year={2025},
  publisher={Nature Publishing Group UK London}
}

@misc{deepmind2024gemini,
  author={Google DeepMind},
  title={Introducing Gemini 2.0: our new AI model for the agentic era},
  year={2024},
  url={https://blog.google/technology/google-deepmind/google-gemini-ai-update-december-2024/}
}

@article{huang2025visionr1,
  title={Vision-r1: Incentivizing reasoning capability in multimodal large language models},
  author={Huang, Wenxuan and Jia, Bohan and Zhai, Zijie and Cao, Shaosheng and Ye, Zheyu and Zhao, Fei and Xu, Zhe and Hu, Yao and Lin, Shaohui},
  journal={arXiv preprint arXiv:2503.06749},
  year={2025}
}

@inproceedings{kafle2018dvqa,
  title={DVQA: Understanding data visualizations via question answering},
  author={Kafle, Kushal and Price, Brian and Cohen, Scott and Kanan, Christopher},
  booktitle={Proceedings of the IEEE/CVF Conference on Computer Vision and Pattern Recognition},
  pages={5648--5656},
  year={2018}
}

@article{hurst2024gpt4o,
  title={Gpt-4o system card},
  author={Hurst, Aaron and Lerer, Adam and Goucher, Adam P and Perelman, Adam and Ramesh, Aditya and Clark, Aidan and Ostrow, AJ and Welihinda, Akila and Hayes, Alan and Radford, Alec and others},
  journal={arXiv preprint arXiv:2410.21276},
  year={2024}
}

@article{deng2025openvlthinker,
  title={Openvlthinker: Complex vision-language reasoning via iterative sft-rl cycles},
  author={Deng, Yihe and Bansal, Hritik and Yin, Fan and Peng, Nanyun and Wang, Wei and Chang, Kai-Wei},
  journal={arXiv preprint arXiv:2503.17352},
  year={2025}
}

@article{wang2024charxiv,
  title={Charxiv: Charting gaps in realistic chart understanding in multimodal llms},
  author={Wang, Zirui and Xia, Mengzhou and He, Luxi and Chen, Howard and Liu, Yitao and Zhu, Richard and Liang, Kaiqu and Wu, Xindi and Liu, Haotian and Malladi, Sadhika and others},
  journal={Advances in Neural Information Processing Systems},
  volume={37},
  pages={113569--113697},
  year={2024}
}

@inproceedings{masry2022chartqa,
  title={Chartqa: A benchmark for question answering about charts with visual and logical reasoning},
  author={Masry, Ahmed and Long, Do Xuan and Tan, Jia Qing and Joty, Shafiq and Hoque, Enamul},
  booktitle={Findings of the association for computational linguistics: ACL 2022},
  pages={2263--2279},
  year={2022}
}

@inproceedings{huang2025evochart,
  title={Evochart: A benchmark and a self-training approach towards real-world chart understanding},
  author={Huang, Muye and Lai, Han and Zhang, Xinyu and Wu, Wenjun and Ma, Jie and Zhang, Lingling and Liu, Jun},
  booktitle={Proceedings of the AAAI Conference on Artificial Intelligence},
  volume={39},
  number={4},
  pages={3680--3688},
  year={2025}
}

@inproceedings{methani2020plotqa,
  title={Plotqa: Reasoning over scientific plots},
  author={Methani, Nitesh and Ganguly, Pritha and Khapra, Mitesh M and Kumar, Pratyush},
  booktitle={Proceedings of the ieee/cvf winter conference on applications of computer vision},
  pages={1527--1536},
  year={2020}
}

@article{xu2023chartbench,
  title={Chartbench: A benchmark for complex visual reasoning in charts},
  author={Xu, Zhengzhuo and Du, Sinan and Qi, Yiyan and Xu, Chengjin and Yuan, Chun and Guo, Jian},
  journal={arXiv preprint arXiv:2312.15915},
  year={2023}
}

@article{gemma2025gemma3,
  title={Gemma 3 technical report},
  author={Gemma Team and Kamath, Aishwarya and Ferret, Johan and Pathak, Shreya and Vieillard, Nino and Merhej, Ramona and Perrin, Sarah and Matejovicova, Tatiana and Rame, Alexandre and Riviere, Morgane and others},
  journal={arXiv preprint arXiv:2503.19786},
  year={2025}
}

@article{kwai2025keye,
  title={Kwai keye-vl technical report},
  author={Team, Kwai Keye and Yang, Biao and Wen, Bin and Liu, Changyi and Chu, Chenglong and Song, Chengru and Rao, Chongling and Yi, Chuan and Li, Da and Zang, Dunju and others},
  journal={arXiv preprint arXiv:2507.01949},
  year={2025}
}

@article{chen2024internvl2_5,
  title={Expanding performance boundaries of open-source multimodal models with model, data, and test-time scaling},
  author={Chen, Zhe and Wang, Weiyun and Cao, Yue and Liu, Yangzhou and Gao, Zhangwei and Cui, Erfei and Zhu, Jinguo and Ye, Shenglong and Tian, Hao and Liu, Zhaoyang and others},
  journal={arXiv preprint arXiv:2412.05271},
  year={2024}
}

@inproceedings{chen2024internvl,
  title={Internvl: Scaling up vision foundation models and aligning for generic visual-linguistic tasks},
  author={Chen, Zhe and Wu, Jiannan and Wang, Wenhai and Su, Weijie and Chen, Guo and Xing, Sen and Zhong, Muyan and Zhang, Qinglong and Zhu, Xizhou and Lu, Lewei and others},
  booktitle={Proceedings of the IEEE/CVF conference on computer vision and pattern recognition},
  pages={24185--24198},
  year={2024}
}

@article{fachada2025gpt41,
  title={GPT-4.1 sets the standard in automated experiment design using novel python libraries},
  author={Fachada, Nuno and Fernandes, Daniel and others},
  journal={arXiv preprint arXiv:2508.00033},
  year={2025}
}

@article{team2024gemini,
  title={Gemini 1.5: Unlocking multimodal understanding across millions of tokens of context},
  author={Team, Gemini and Georgiev, Petko and Lei, Ving Ian and Burnell, Ryan and Bai, Libin and Gulati, Anmol and Tanzer, Garrett and Vincent, Damien and Pan, Zhufeng and Wang, Shibo and others},
  journal={arXiv preprint arXiv:2403.05530},
  year={2024}
}

@article{jia2025chartreasoner,
  title={Chartreasoner: Code-driven modality bridging for long-chain reasoning in chart question answering},
  author={Jia, Caijun and Xu, Nan and Wei, Jingxuan and Wang, Qingli and Wang, Lei and Yu, Bihui and Zhu, Junnan},
  journal={arXiv preprint arXiv:2506.10116},
  year={2025}
}

@article{han2023chartllama,
  title={Chartllama: A multimodal llm for chart understanding and generation},
  author={Han, Yucheng and Zhang, Chi and Chen, Xin and Yang, Xu and Wang, Zhibin and Yu, Gang and Fu, Bin and Zhang, Hanwang},
  journal={arXiv preprint arXiv:2311.16483},
  year={2023}
}

@article{masry2024chartgemma,
  title={ChartGemma: Visual instruction-tuning for chart reasoning in the wild},
  author={Masry, Ahmed and Thakkar, Megh and Bajaj, Aayush and Kartha, Aaryaman and Hoque, Enamul and Joty, Shafiq},
  journal={arXiv preprint arXiv:2407.04172},
  year={2024}
}

@article{liu2023llava,
  title={Visual instruction tuning},
  author={Liu, Haotian and Li, Chunyuan and Wu, Qingyang and Lee, Yong Jae},
  journal={Advances in Neural Information Processing Systems},
  volume={36},
  pages={34892--34916},
  year={2023}
}

@article{alayrac2022flamingo,
  title={Flamingo: a visual language model for few-shot learning},
  author={Alayrac, Jean-Baptiste and Donahue, Jeff and Luc, Pauline and Miech, Antoine and Barr, Iain and Hasson, Yana and Lenc, Karel and Mensch, Arthur and Millican, Katherine and Reynolds, Malcolm and others},
  journal={Advances in neural information processing systems},
  volume={35},
  pages={23716--23736},
  year={2022}
}

@article{zhang2024tinychart,
  title={Tinychart: Efficient chart understanding with visual token merging and program-of-thoughts learning},
  author={Zhang, Liang and Hu, Anwen and Xu, Haiyang and Yan, Ming and Xu, Yichen and Jin, Qin and Zhang, Ji and Huang, Fei},
  journal={arXiv preprint arXiv:2404.16635},
  year={2024}
}

@article{kaelbling1996reinforcement,
  title={Reinforcement learning: A survey},
  author={Kaelbling, Leslie Pack and Littman, Michael L and Moore, Andrew W},
  journal={Journal of artificial intelligence research},
  volume={4},
  pages={237--285},
  year={1996}
}

@article{kahou2017figureqa,
  title={Figureqa: An annotated figure dataset for visual reasoning},
  author={Kahou, Samira Ebrahimi and Michalski, Vincent and Atkinson, Adam and Kadar, Akos and Trischler, Adam and Bengio, Yoshua},
  journal={arXiv preprint arXiv:1710.07300},
  year={2017}
}

@article{wei2022cot,
  title={Chain-of-thought prompting elicits reasoning in large language models},
  author={Wei, Jason and Wang, Xuezhi and Schuurmans, Dale and Bosma, Maarten and Xia, Fei and Chi, Ed and Le, Quoc V and Zhou, Denny and others},
  journal={Advances in neural information processing systems},
  volume={35},
  pages={24824--24837},
  year={2022}
}

@article{li2025llavaonevision,
  title={Llava-onevision: Easy visual task transfer},
  author={Li, Bo and Zhang, Yuanhan and Guo, Dong and Zhang, Renrui and Li, Feng and Zhang, Hao and Zhang, Kaichen and Zhang, Peiyuan and Li, Yanwei and Liu, Ziwei and others},
  journal={arXiv preprint arXiv:2408.03326},
  year={2024}
}

@inproceedings{zhai2025vlm_rl,
 author = {Zhai, Yuexiang and Bai, Hao and Lin, Zipeng and Pan, Jiayi and Tong, Shengbang and Zhou, Yifei and Suhr, Alane and Xie, Saining and LeCun, Yann and Ma, Yi and Levine, Sergey},
 booktitle = {Advances in Neural Information Processing Systems},
 pages = {110935--110971},
 title = {Fine-Tuning Large Vision-Language Models as Decision-Making Agents via Reinforcement Learning},
 year = {2024}
}

@article{openai2025gpt5,
  title={Openai gpt-5 system card},
  author={Singh, Aaditya and Fry, Adam and Perelman, Adam and Tart, Adam and Ganesh, Adi and El-Kishky, Ahmed and McLaughlin, Aidan and Low, Aiden and Ostrow, AJ and Ananthram, Akhila and others},
  journal={arXiv preprint arXiv:2601.03267},
  year={2025}
}

@article{brown2020language_model,
  title={Language models are few-shot learners},
  author={B. Brown, Tom and Mann, Benjamin and Ryder, Nick and Subbiah, Melanie and Kaplan, Jared and Dhariwal, Prafulla and Neelakantan, Arvind and Shyam, Pranav and Sastry, Girish and Askell, Amanda and others},
  journal={Advances in neural information processing systems},
  volume={33},
  pages={1877--1901},
  year={2020}
}

@article{achiam2023gpt4,
  title={Gpt-4 technical report},
  author={Achiam, Josh and Adler, Steven and Agarwal, Sandhini and Ahmad, Lama and Akkaya, Ilge and Aleman, Florencia Leoni and Almeida, Diogo and Altenschmidt, Janko and Altman, Sam and Anadkat, Shyamal and others},
  journal={arXiv preprint arXiv:2303.08774},
  year={2023}
}

@article{touvron2023llama,
  title={Llama: Open and efficient foundation language models},
  author={Touvron, Hugo and Lavril, Thibaut and Izacard, Gautier and Martinet, Xavier and Lachaux, Marie-Anne and Lacroix, Timothee and Roziere, Baptiste and Goyal, Naman and Hambro, Eric and Azhar, Faisal and others},
  journal={arXiv preprint arXiv:2302.13971},
  year={2023}
}

@article{shao2024deepseekmath,
  title={Deepseekmath: Pushing the limits of mathematical reasoning in open language models},
  author={Shao, Zhihong and Wang, Peiyi and Zhu, Qihao and Xu, Runxin and Song, Junxiao and Bi, Xiao and Zhang, Haowei and Zhang, Mingchuan and others},
  journal={arXiv preprint arXiv:2402.03300},
  year={2024}
}

@article{wang2025visualprm,
  title={Visualprm: An effective process reward model for multimodal reasoning},
  author={Wang, Weiyun and Gao, Zhangwei and Chen, Lianjie and Chen, Zhe and Zhu, Jinguo and Zhao, Xiangyu and Liu, Yangzhou and Cao, Yue and Ye, Shenglong and Zhu, Xizhou and others},
  journal={arXiv preprint arXiv:2503.10291},
  year={2025}
}

@article{yao2025r1_share,
  title={R1-sharevl: Incentivizing reasoning capability of multimodal large language models via share-grpo},
  author={Yao, Huanjin and Yin, Qixiang and Zhang, Jingyi and Yang, Min and Wang, Yibo and Wu, Wenhao and Su, Fei and Shen, Li and Qiu, Minghui and Tao, Dacheng and others},
  journal={arXiv preprint arXiv:2505.16673},
  year={2025}
}

@article{liu2025noisyrollout,
  title={Noisyrollout: Reinforcing visual reasoning with data augmentation},
  author={Liu, Xiangyan and Ni, Jinjie and Wu, Zijian and Du, Chao and Dou, Longxu and Wang, Haonan and Pang, Tianyu and Shieh, Michael Qizhe},
  journal={arXiv preprint arXiv:2504.13055},
  year={2025}
}

@inproceedings{hu2024novachart,
  title={Novachart: A large-scale dataset towards chart understanding and generation of multimodal large language models},
  author={Hu, Linmei and Wang, Duokang and Pan, Yiming and Yu, Jifan and Shao, Yingxia and Feng, Chong and Nie, Liqiang},
  booktitle={Proceedings of the 32nd ACM International Conference on Multimedia},
  pages={3917--3925},
  year={2024}
}

@article{zhu2025internvl3,
  title={Internvl3: Exploring advanced training and test-time recipes for open-source multimodal models},
  author={Zhu, Jinguo and Wang, Weiyun and Chen, Zhe and Liu, Zhaoyang and Ye, Shenglong and Gu, Lixin and Tian, Hao and Duan, Yuchen and Su, Weijie and Shao, Jie and others},
  journal={arXiv preprint arXiv:2504.10479},
  year={2025}
}

@article{Shulman2017PPO,
  title={Proximal policy optimization algorithms},
  author={Schulman, John and Wolski, Filip and Dhariwal, Prafulla and Radford, Alec and Klimov, Oleg},
  journal={arXiv preprint arXiv:1707.06347},
  year={2017}
}

@article{Rafailov2023DPO,
  title={Direct preference optimization: Your language model is secretly a reward model},
  author={Rafailov, Rafael and Sharma, Archit and Mitchell, Eric and Manning, Christopher D and Ermon, Stefano and Finn, Chelsea},
  journal={Advances in neural information processing systems},
  volume={36},
  pages={53728--53741},
  year={2023}
}

@inproceedings{yang2025r1onevision,
  title={R1-onevision: Advancing generalized multimodal reasoning through cross-modal formalization},
  author={Yang, Yi and He, Xiaoxuan and Pan, Hongkun and Jiang, Xiyan and Deng, Yan and Yang, Xingtao and Lu, Haoyu and Yin, Dacheng and Rao, Fengyun and Zhu, Minfeng and others},
  booktitle={Proceedings of the IEEE/CVF International Conference on Computer Vision},
  pages={2376--2385},
  year={2025}
}

@inproceedings{zhang2025r1vl,
  title={R1-VL: Learning to reason with multimodal large language models via step-wise group relative policy optimization},
  author={Zhang, Jingyi and Huang, Jiaxing and Yao, Huanjin and Liu, Shunyu and Zhang, Xikun and Lu, Shijian and Tao, Dacheng},
  booktitle={Proceedings of the IEEE/CVF International Conference on Computer Vision},
  pages={1859--1869},
  year={2025}
}

@inproceedings{zheng2024llamafactory,
  title={Llamafactory: Unified efficient fine-tuning of 100+ language models},
  author={Zheng, Yaowei and Zhang, Richong and Zhang, Junhao and Ye, Yanhan and Luo, Zheyan},
  booktitle={Proceedings of the 62nd annual meeting of the association for computational linguistics (volume 3: system demonstrations)},
  pages={400--410},
  year={2024}
}

@inproceedings{sheng2025verl,
  title={Hybridflow: A flexible and efficient rlhf framework},
  author={Sheng, Guangming and Zhang, Chi and Ye, Zilingfeng and Wu, Xibin and Zhang, Wang and Zhang, Ru and Peng, Yanghua and Lin, Haibin and Wu, Chuan},
  booktitle={Proceedings of the Twentieth European Conference on Computer Systems},
  pages={1279--1297},
  year={2025}
}

@inproceedings{duan2024vlmevalkit,
  title={Vlmevalkit: An open-source toolkit for evaluating large multi-modality models},
  author={Duan, Haodong and Yang, Junming and Qiao, Yuxuan and Fang, Xinyu and Chen, Lin and Liu, Yuan and Dong, Xiaoyi and Zang, Yuhang and Zhang, Pan and Wang, Jiaqi and others},
  booktitle={Proceedings of the 32nd ACM international conference on multimedia},
  pages={11198--11201},
  year={2024}
}

@inproceedings{kwon2023vllm,
  title={Efficient memory management for large language model serving with pagedattention},
  author={Kwon, Woosuk and Li, Zhuohan and Zhuang, Siyuan and Sheng, Ying and Zheng, Lianmin and Yu, Cody Hao and Gonzalez, Joseph and Zhang, Hao and Stoica, Ion},
  booktitle={Proceedings of the 29th symposium on operating systems principles},
  pages={611--626},
  year={2023}
}

@inproceedings{xu2025chartpoint,
  title={Chartpoint: Guiding mllms with grounding reflection for chart reasoning},
  author={Xu, Zhengzhuo and Du, SiNan and Qi, Yiyan and Lu, Siwen and Xu, Chengjin and Yuan, Chun and Guo, Jian},
  booktitle={Proceedings of the IEEE/CVF International Conference on Computer Vision},
  pages={426--436},
  year={2025}
}

@article{huang2025chartsketcher,
  title={Chartsketcher: Reasoning with multimodal feedback and reflection for chart understanding},
  author={Huang, Muye and Zhang, Lingling and Ma, Jie and Lai, Han and Xu, Fangzhi and Li, Yifei and Wu, Wenjun and Wu, Yaqiang and Liu, Jun},
  journal={arXiv preprint arXiv:2505.19076},
  year={2025}
}

@article{wu2024deepseekvl2,
  title={Deepseek-vl2: Mixture-of-experts vision-language models for advanced multimodal understanding},
  author={Wu, Zhiyu and Chen, Xiaokang and Pan, Zizheng and Liu, Xingchao and Liu, Wen and Dai, Damai and Gao, Huazuo and Ma, Yiyang and Wu, Chengyue and Wang, Bingxuan and others},
  journal={arXiv preprint arXiv:2412.10302},
  year={2024}
}
}

\clearpage

\appendix
\section*{Appendix}

\begin{figure*}[t]
\section{Cold-Start Data Examples} \label{sec:cold_start}
\centering
\includegraphics[width=\textwidth]{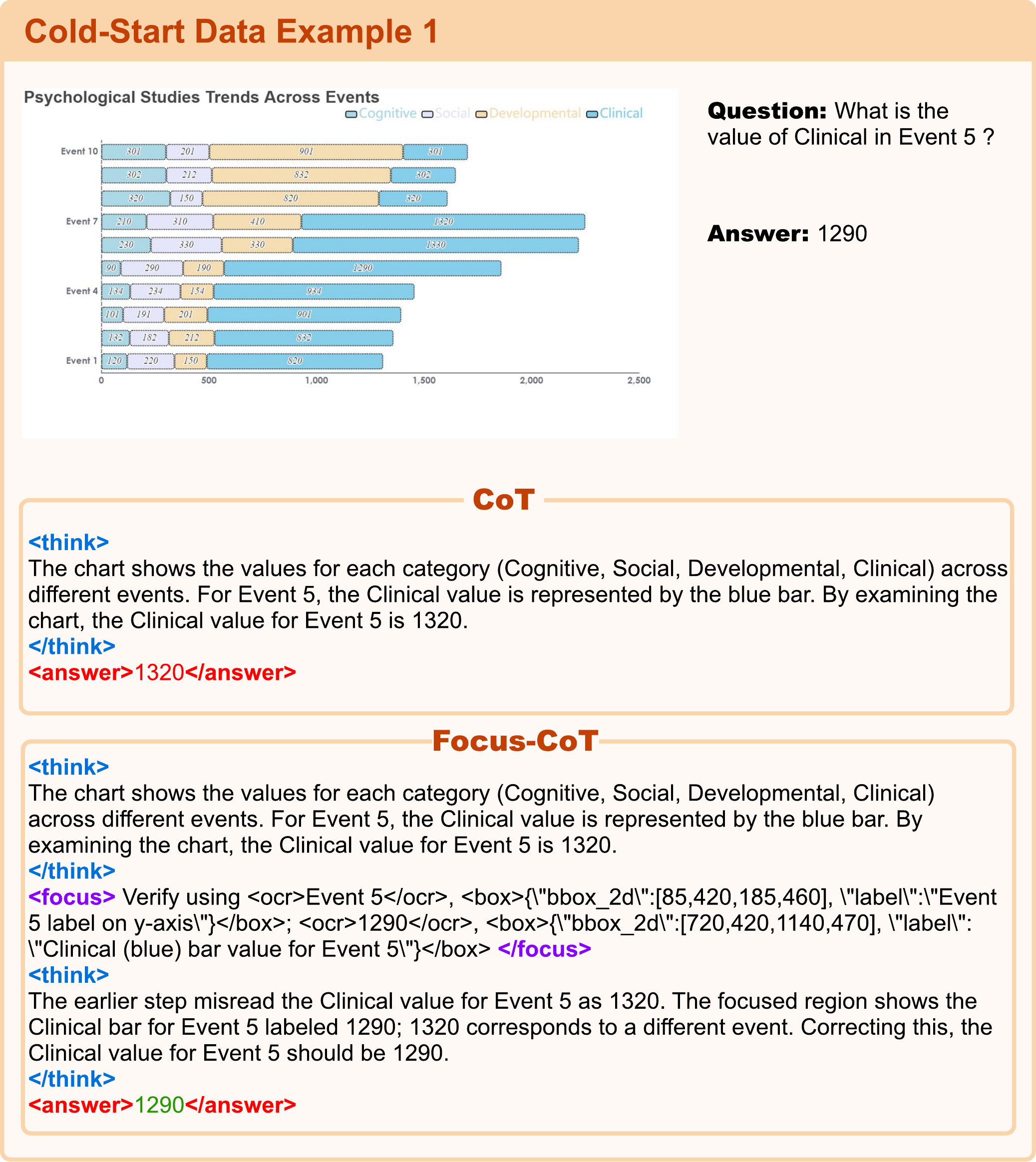}
\label{img:sft_case1}
\end{figure*}

\begin{figure*}[t]
\section{Chart-FR1 Inference Examples} \label{sec:chart_fr1}
\centering
\includegraphics[width=\textwidth]{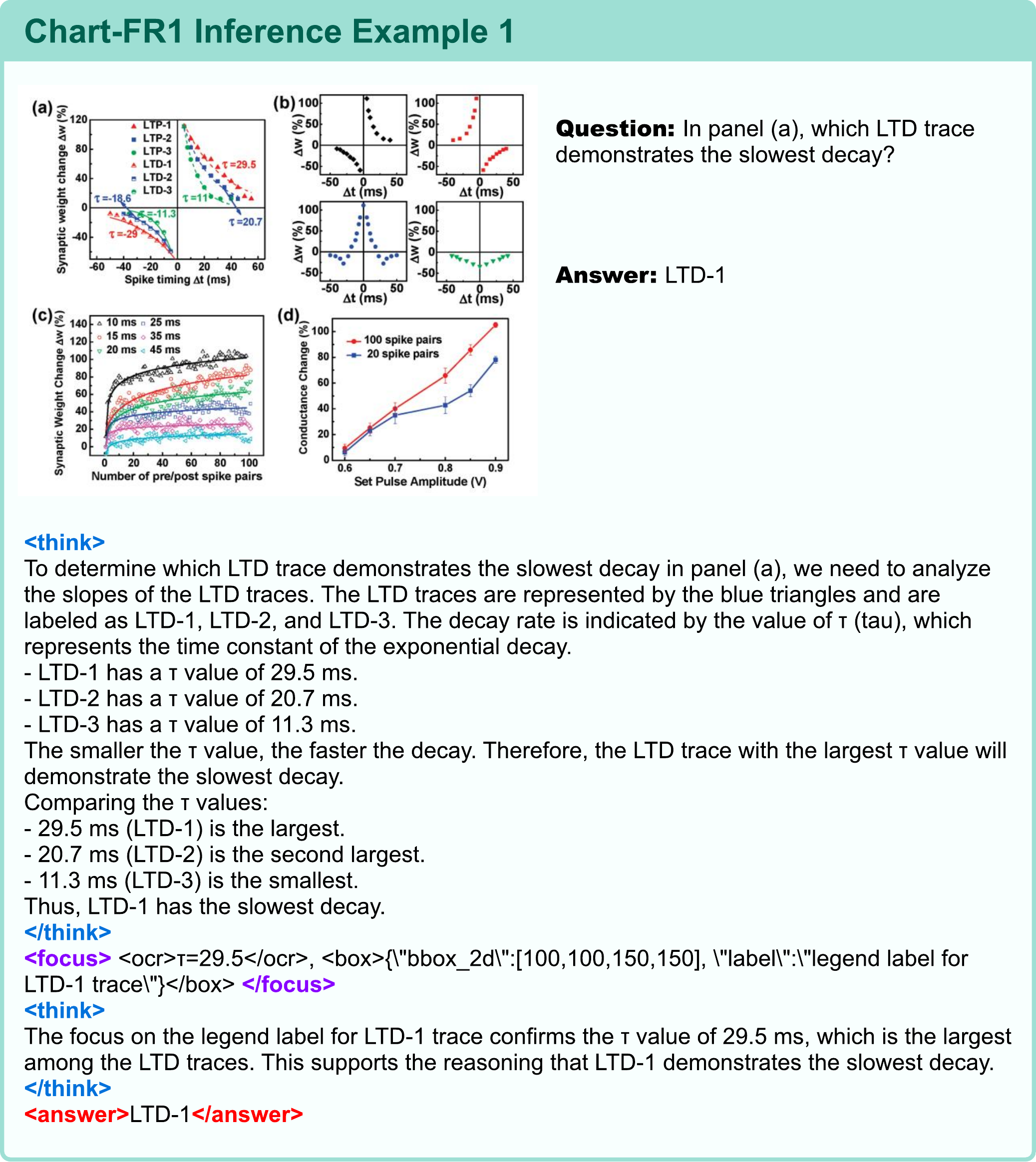}
\label{img:inference_case1}
\end{figure*}

\begin{figure*}[!t]
\section{HID-Chart Examples} \label{sec:hid_chart}
\centering
\includegraphics[width=\textwidth]{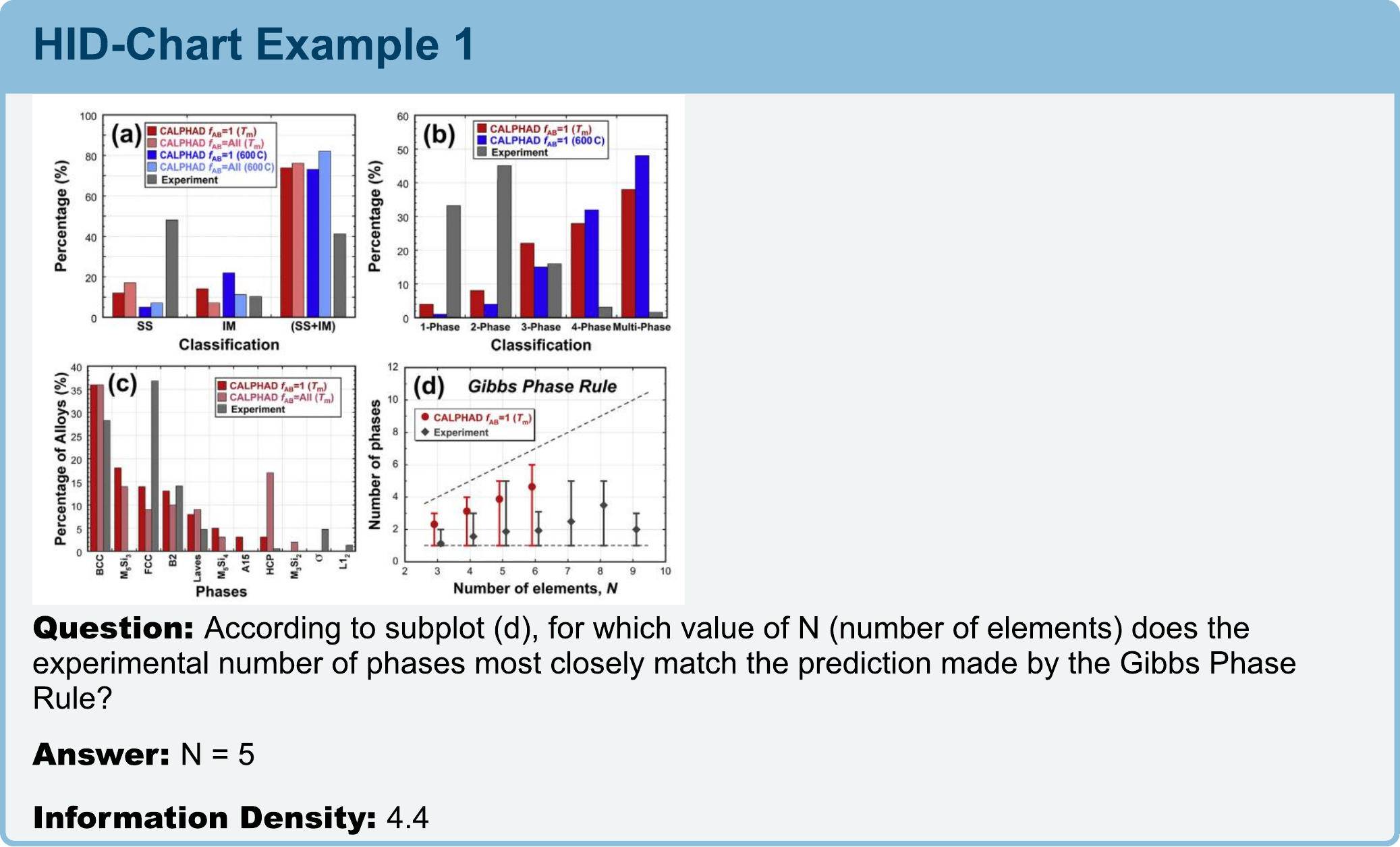}

\vspace{1em}

\includegraphics[width=\textwidth]{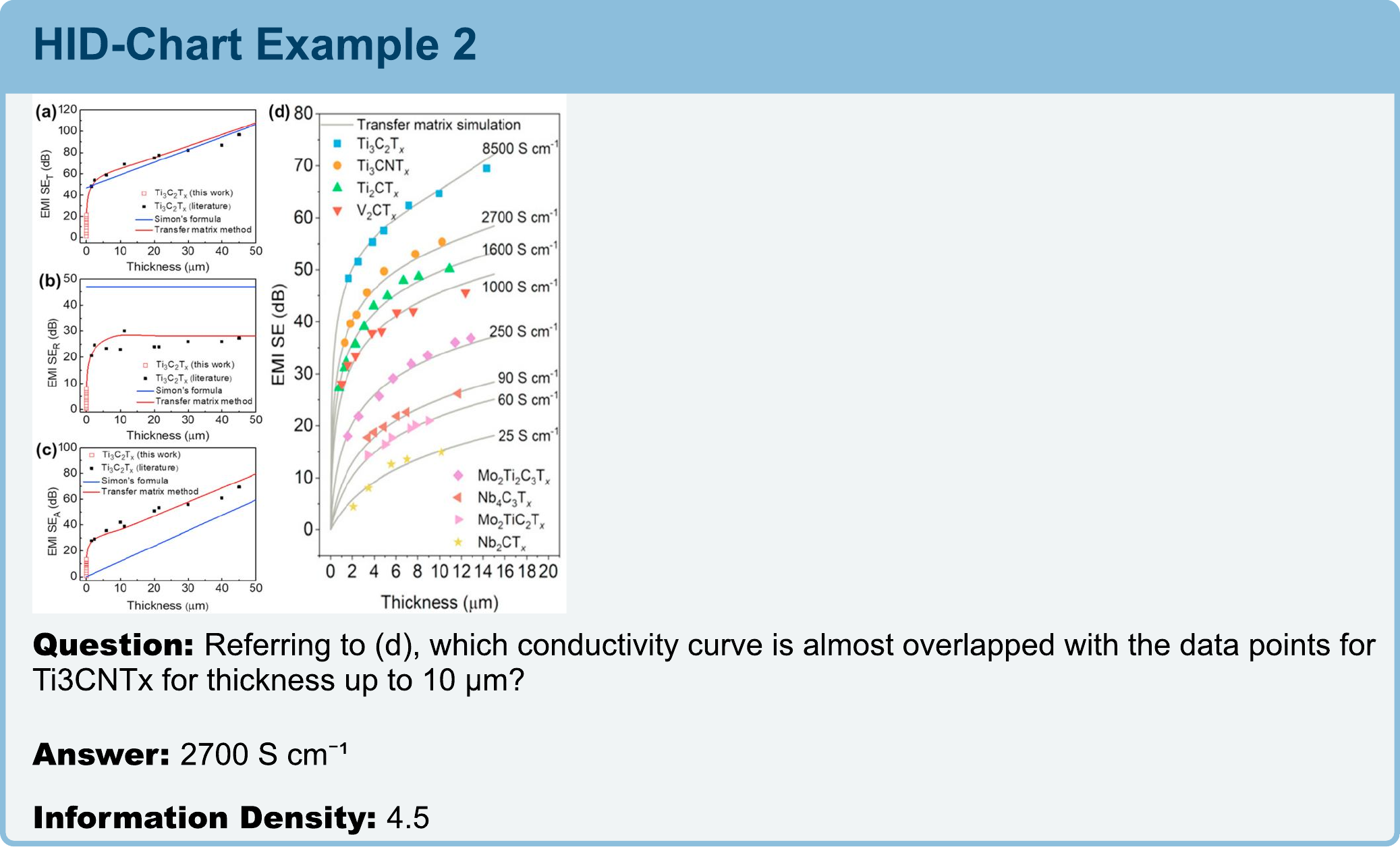}
\label{fig:benchmark_cases}
\end{figure*}

In this appendix, we provide additional details and results to complement the main paper. Specifically, Sec.~\ref{sec:model_training} outlines model and training details, Sec.~\ref{sec:data} describes data details, Sec.~\ref{sec:evaluation} provides evaluation details, Sec.~\ref{sec:cold_start} presents cold-start data examples, Sec.~\ref{sec:chart_fr1} shows Chart-FR1 inference examples, and Sec.~\ref{sec:hid_chart} presents HID-Chart examples.

\section{Model and Training Details} \label{sec:model_training}
\begin{figure}[t]
    \centering
    \includegraphics[width=\columnwidth]{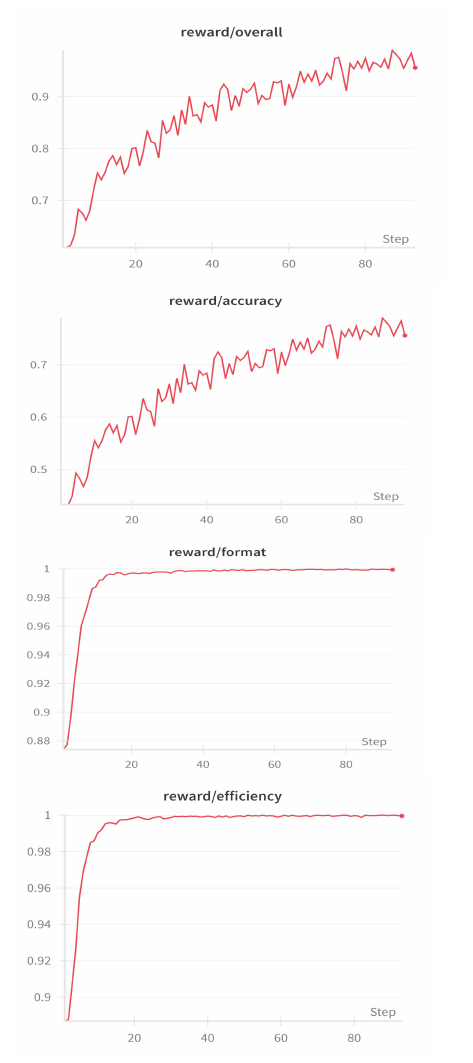}
    \caption{Reward curves during Focus-GRPO training.}
    \label{img:reward_curve}
\end{figure}

\begin{figure*}[t]
\centering
\includegraphics[width=\textwidth]{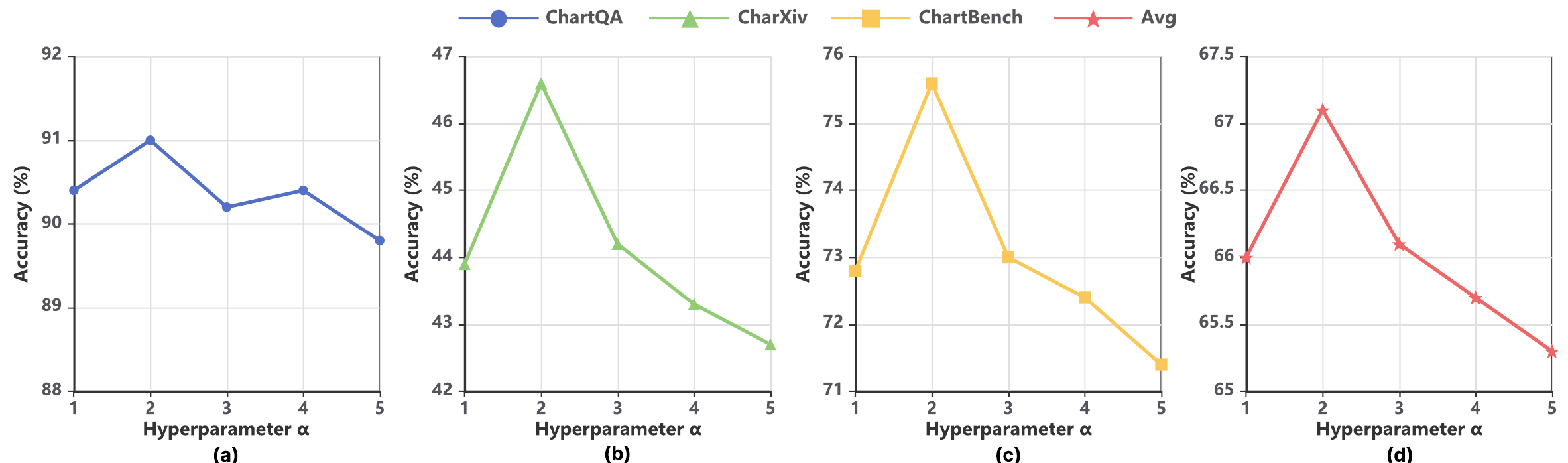}
\caption{Effect of the hyperparameter $\alpha$ on model performance. Setting $\alpha = 2$ achieves the best performance in five chart benchmarks.}
\label{img:line_alpha}
\end{figure*}

\begin{figure*}[t]
\centering
\includegraphics[width=\textwidth]{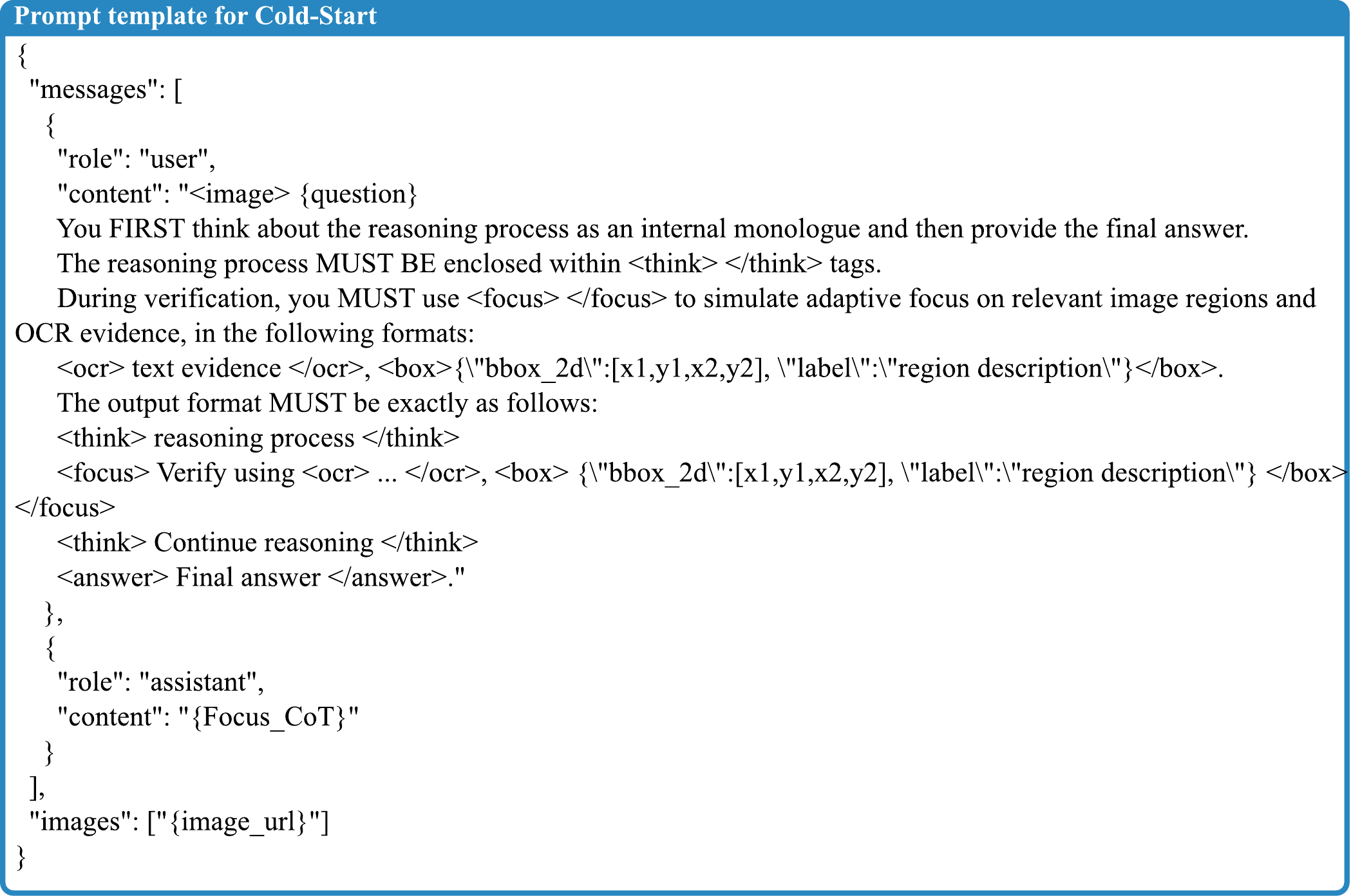}
\caption{The prompt template for Cold-Start.}
\label{img:sft_template}
\end{figure*}

\begin{figure}[!htbp]
  \centering

  \includegraphics[width=\columnwidth]{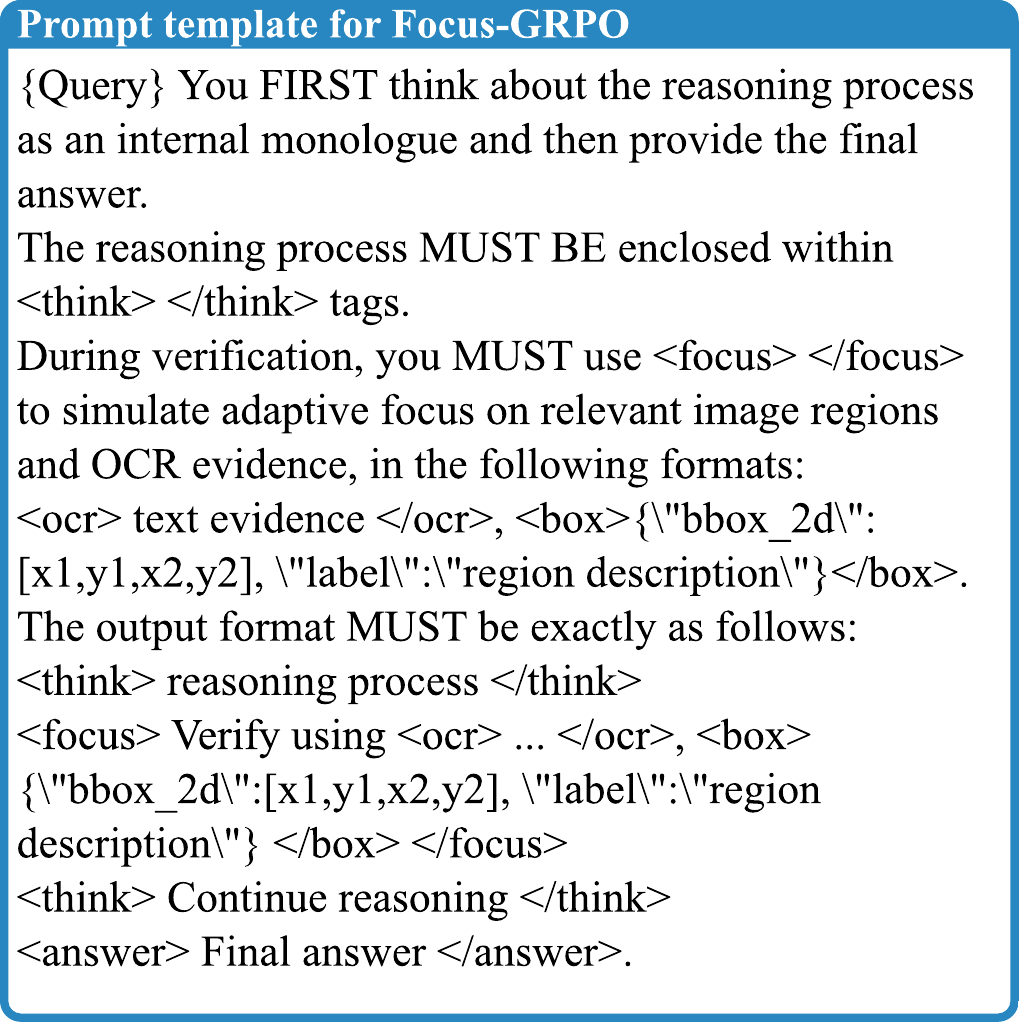}
  \caption{The prompt template for Focus-GRPO.}
  \label{img:rl_template}



\end{figure}

We adopt Qwen2.5-VL-7B~\cite{bai2025qwen} as our base model. Training follows a two-stage focused reasoning paradigm, and code implementation is based on the Llama-Factory~\cite{zheng2024llamafactory} and VeRL~\cite{sheng2025verl} frameworks. The training parameters are summarized in Table~\ref{tab:training_params_SFT} and \ref{tab:training_params_RL}. The reward curves during training are shown in Fig.~\ref{img:reward_curve}. The effect of the hyperparameter $\alpha$ on model performance is shown in Fig.~\ref{img:line_alpha}. The impact of the adaptive KL penalty on reasoning length is reported in Table~\ref{tab:token_stats}. The prompt templates are shown in Fig.~\ref{img:sft_template} and \ref{img:rl_template}.

\begin{table}[h]
\small 
    \centering
    \begin{tabular}{lc}
        \toprule
        Parameter & Value \\
        \midrule
        Learning Rate & $2 \times 10^{-6}$ \\
        Optimizer & BAdam \\
        Global Batch Size & 256 \\
        Warmup Ratio & 0.1 \\
        Finetuning Type & Full \\
        Epochs & 1 \\
        \bottomrule
    \end{tabular}
    \caption{Key training parameters used for Cold-Start.}
    \label{tab:training_params_SFT}
\end{table}

\begin{table}[h]
\small 
    \centering
    \begin{tabular}{lc}
        \toprule
        Parameter & Value \\
        \midrule
        Learning Rate & $1 \times 10^{-6}$ \\
        Weight Decay & $1 \times 10^{-2}$ \\
        Optimizer & AdamW \\
        Global Batch Size & 512 \\
        Max Prompt Length & 2048 \\
        Max Response Length & 2048 \\
        Rollout & 8 \\
        Temperature & 1.0 \\
        Top\_p & 1.0 \\
        $\beta$ & $1 \times 10^{-2}$ \\
        $\alpha$ & 2 \\
        $\tau$ & 0.9 \\
        $w_1$ & 0.1 \\
        $w_2$ & 0.1 \\
        Epochs & 3 \\
        \bottomrule
    \end{tabular}
    \caption{Key training parameters used for Focus-GRPO.}
    \label{tab:training_params_RL}
\end{table}

\vspace{10mm}
 
\begin{table*}[t]
\small
\centering
\newcolumntype{C}{>{\centering\arraybackslash}X}
\begin{tabularx}{\textwidth}{l|CCCCC|C}
\toprule
\textbf{$N_{\text{ocr}} + N_{\text{box}}$} & \textbf{ChartQA} & \textbf{CharXiv} & \textbf{EvoChart} & \textbf{ChartBench} & \textbf{PlotQA} & \textbf{Avg} \\
\midrule
$[0, 2]$ & 218 & 330 & 228 & 221 & 247 & 249 \\
$[3, 4]$ & 265 & 341 & 282 & 242 & 318 & 290 \\
$[5, +\infty]$ & 385 & 459 & 416 & 314 & 419 & 399 \\
\bottomrule
\end{tabularx}
\caption{The effect of adaptive KL penalty on reasoning length. We incorporate an adaptive KL penalty into the GRPO algorithm, partition the number of focused visual cues into three intervals, and report the corresponding reasoning length. The results show that the adaptive KL penalty leads to longer reasoning as the number of cues increases.}
\label{tab:token_stats}
\end{table*}

\vspace{10mm}

\clearpage
\FloatBarrier

\begin{figure}[!htbp]
\centering
\includegraphics[width=\columnwidth]{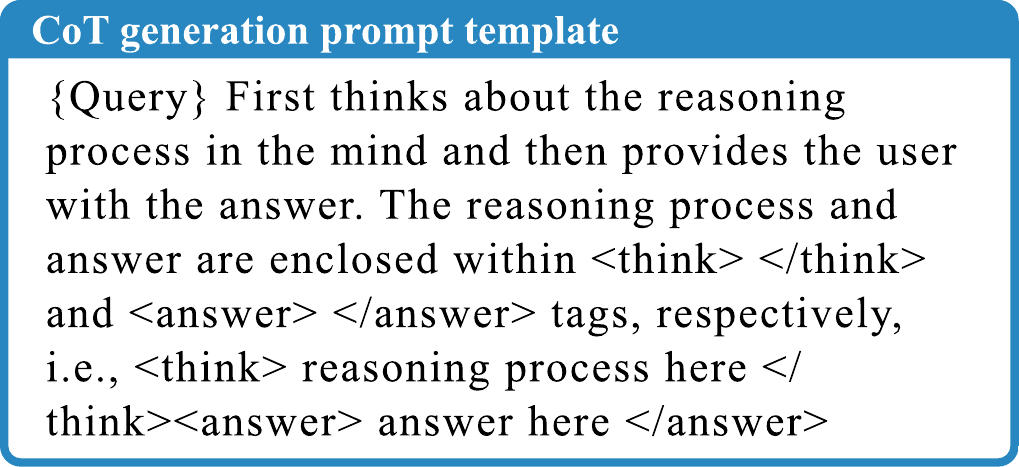}
\caption{The prompt template for CoT generation}
\label{img:cot_generation_template}
\end{figure}

\begin{figure}[!htbp]
\centering
\includegraphics[width=\columnwidth]{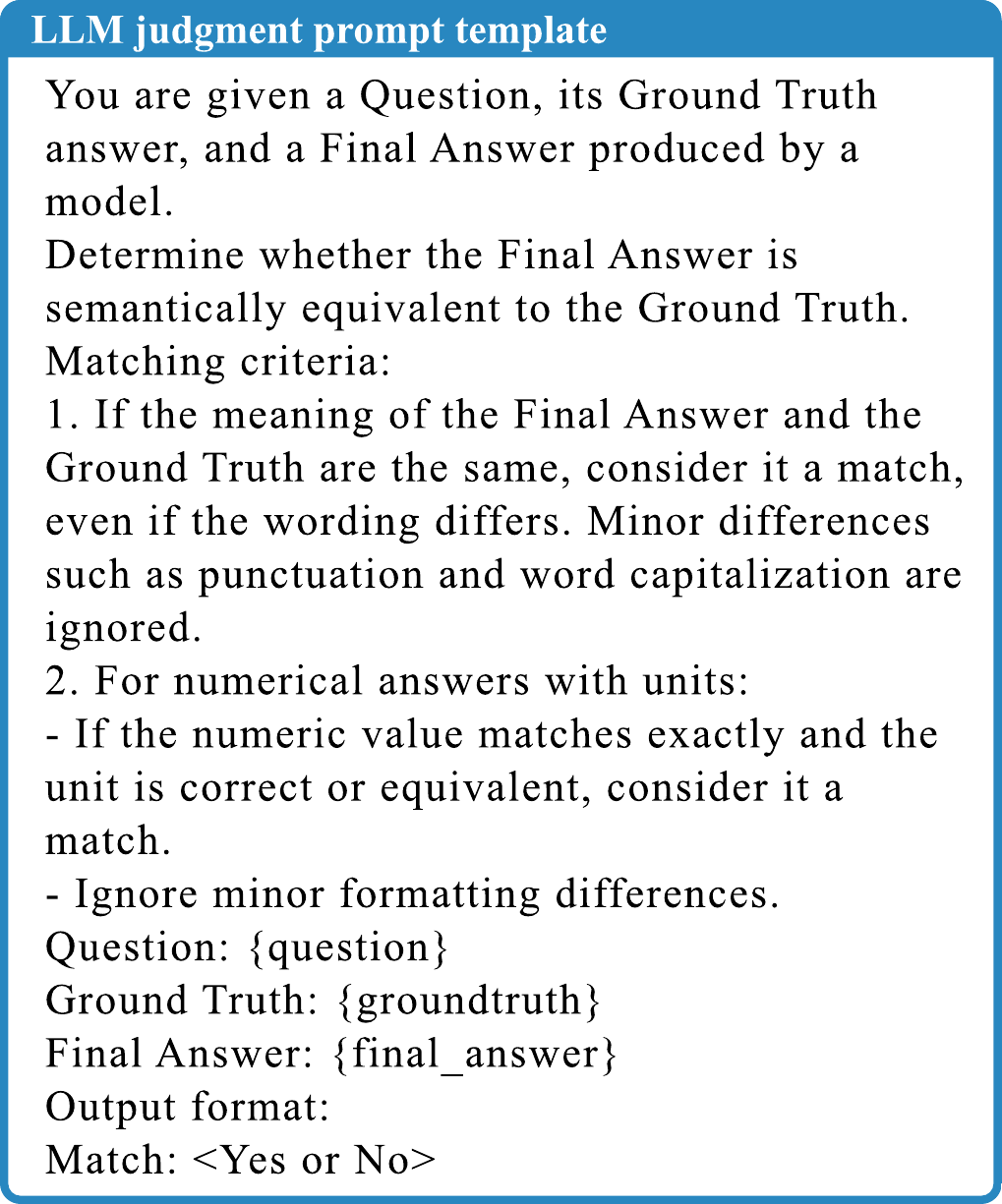}
\caption{The prompt template for LLM judgment}
\label{img:LLM_judge_template}
\end{figure}


\section{Data Details} \label{sec:data}
\begin{figure*}[t]
\centering
\includegraphics[width=\textwidth]{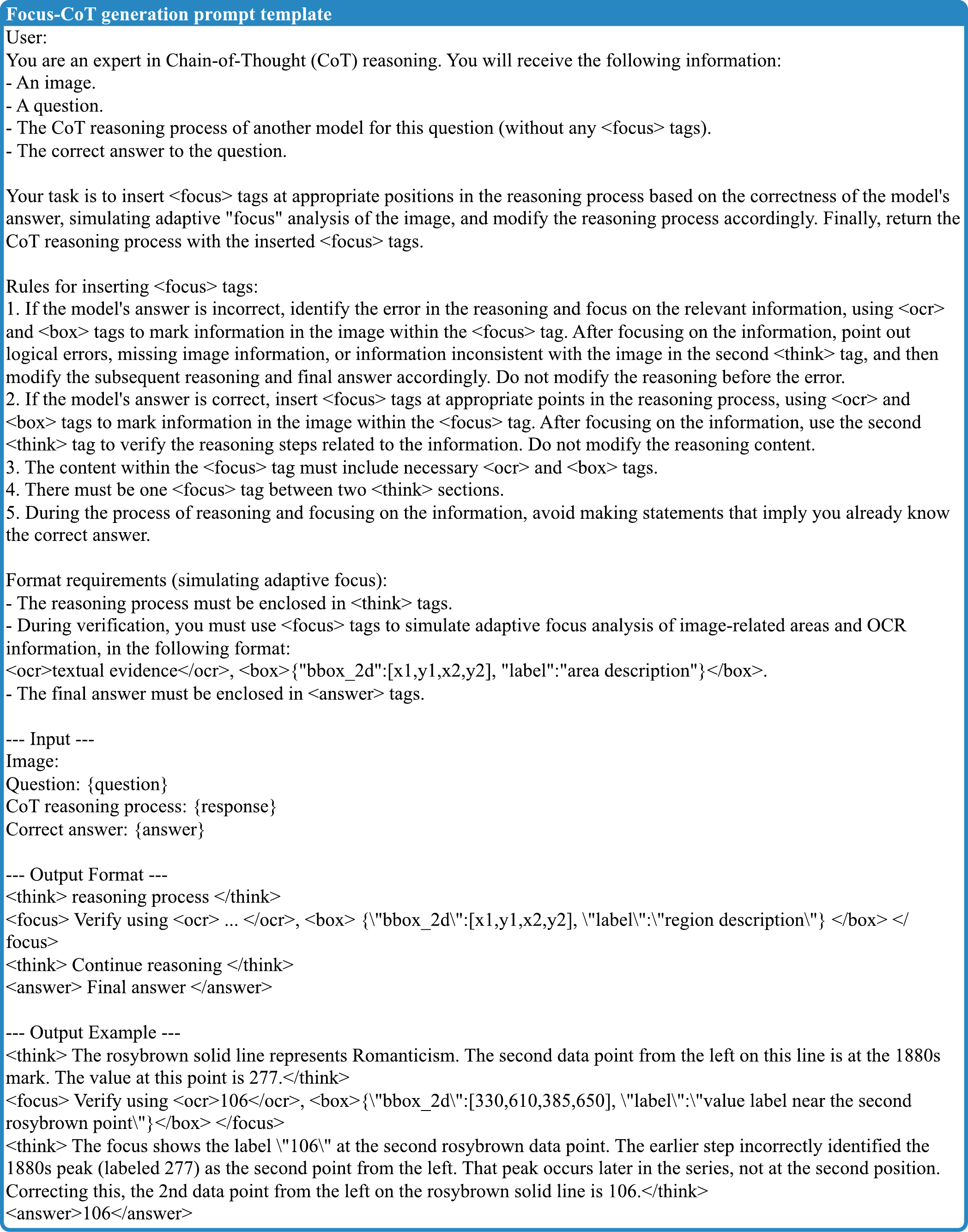}
\caption{The prompt template for Focus-CoT generation}
\label{img:generation_template}
\end{figure*}

We design an automated Focus-CoT generation pipeline and present the prompt templates used in this section. We employ the baseline model Qwen2.5-VL-7B~\cite{bai2025qwen} to produce the initial CoT, with the corresponding prompt template shown in Fig.~\ref{img:cot_generation_template}. The initial CoT is then conditionally reconstructed by GPT-5~\cite{openai2025gpt5} to obtain the Focus-CoT, and the prompt template for this step is illustrated in Fig.~\ref{img:generation_template}. During correctness filtering, the prompt template used for LLM-based judgment is provided in Fig.~\ref{img:LLM_judge_template}.

\section{Evaluation Details} \label{sec:evaluation}
\begin{figure*}[t]
\centering
\includegraphics[width=\textwidth]{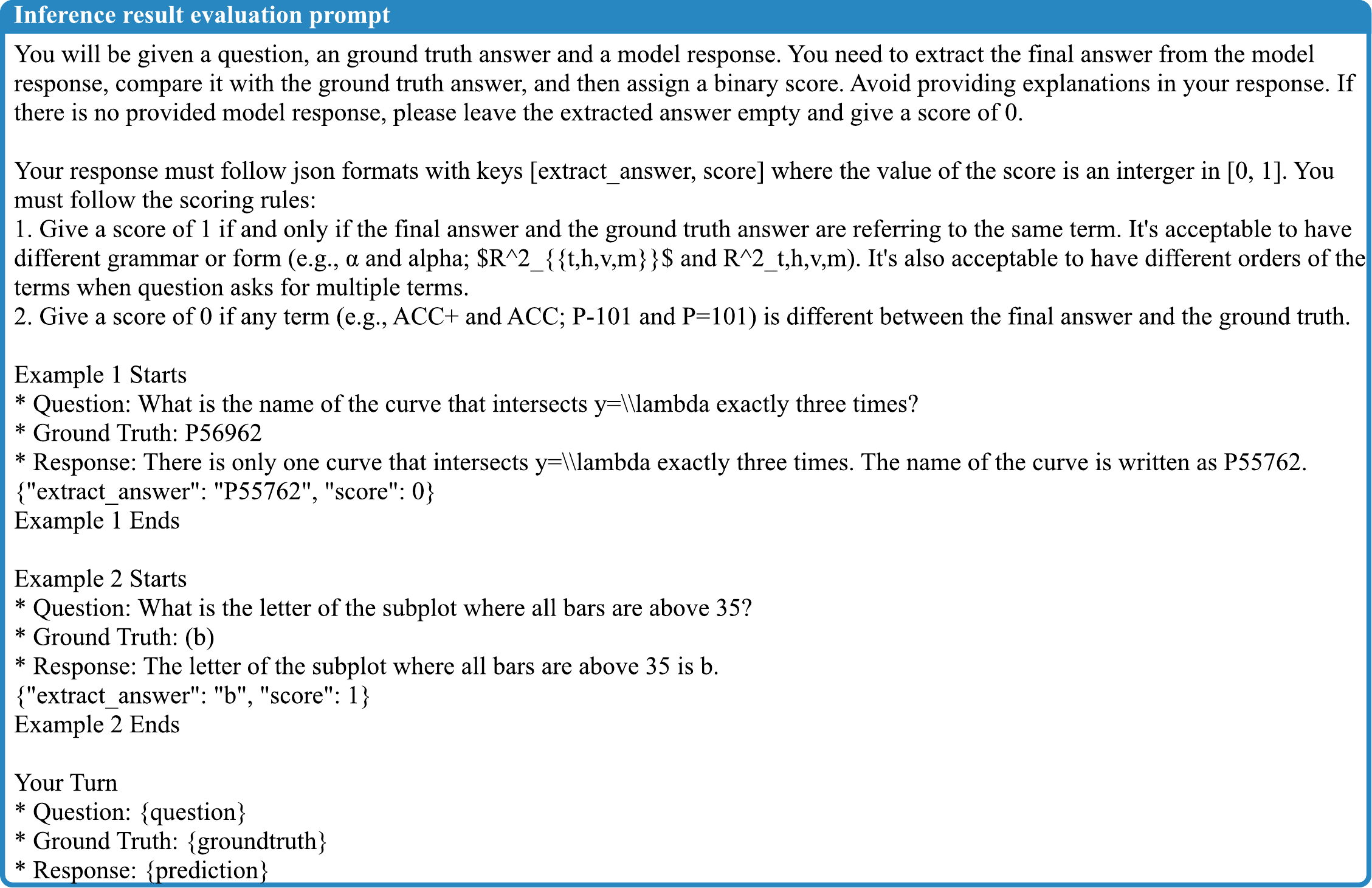}
\caption{The prompt template for evaluation.}
\label{img:evaluation}
\end{figure*}

\begin{figure*}[t]
\centering
\includegraphics[width=\textwidth]{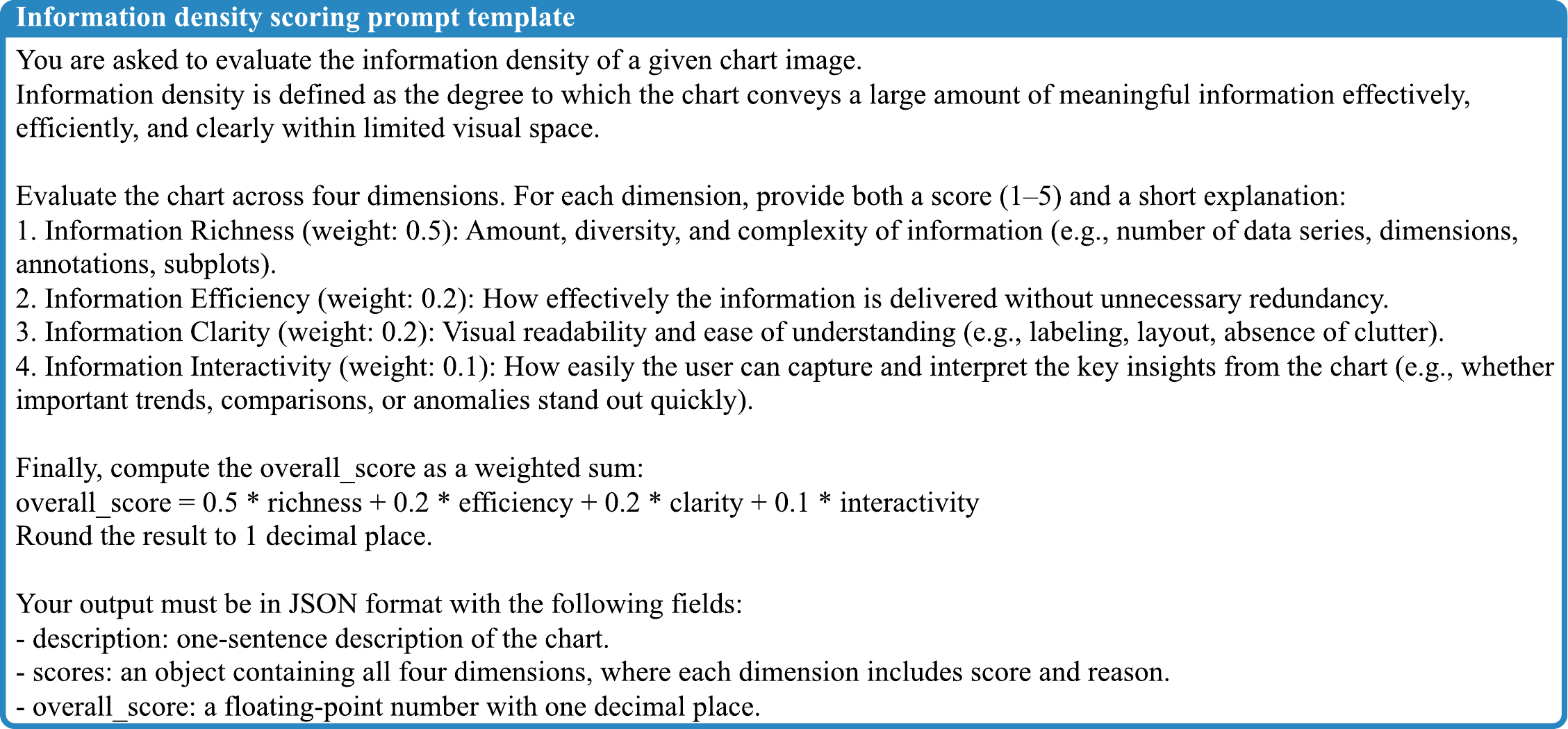}
\caption{The prompt template for calculating the Chart-ID.}
\label{img:chart_id}
\end{figure*}

We conduct comprehensive evaluations on five common chart benchmarks and our HID-Chart. For ChartQA~\cite{masry2022chartqa} and CharXiv~\cite{wang2024charxiv}, we adopt the VLMEvalKit~\cite{duan2024vlmevalkit} framework. For EvoChart~\cite{huang2025evochart}, ChartBench~\cite{xu2023chartbench}, PlotQA~\cite{methani2020plotqa} and HID-Chart, we perform inference using vLLM~\cite{kwon2023vllm} and evaluate the results with GPT-5 mini using a scoring prompt, shown in Fig.~\ref{img:evaluation}. In addition, our proposed Chart-ID information density metric is scored by GPT-5 along four dimensions, and the corresponding prompt is provided in Fig.~\ref{img:chart_id}.

\FloatBarrier

\FloatBarrier

\begin{figure*}[t]
\centering
\includegraphics[width=\textwidth]{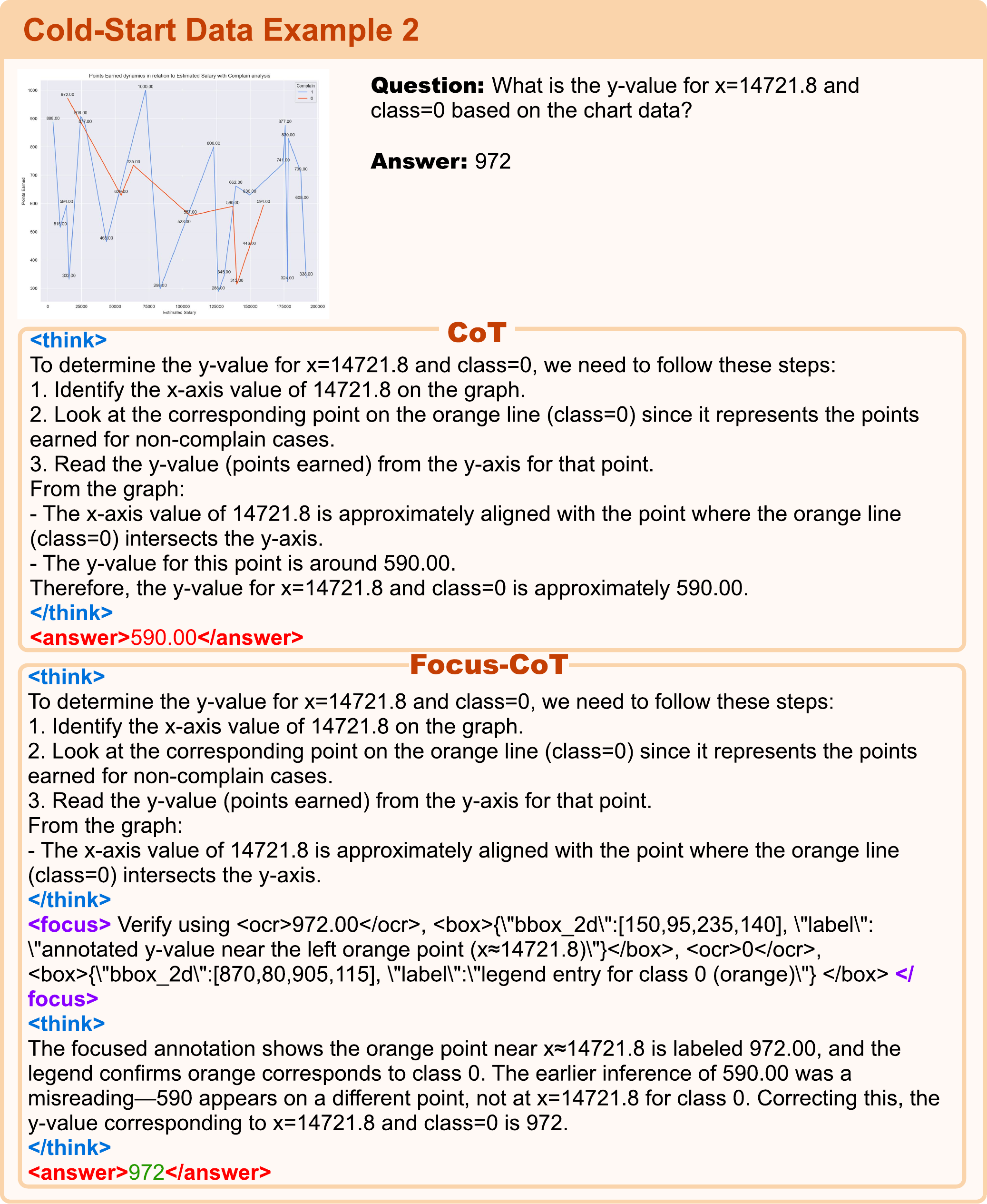}
\label{img:sft_case2}
\end{figure*}

\begin{figure*}[t]
\centering
\includegraphics[width=\textwidth]{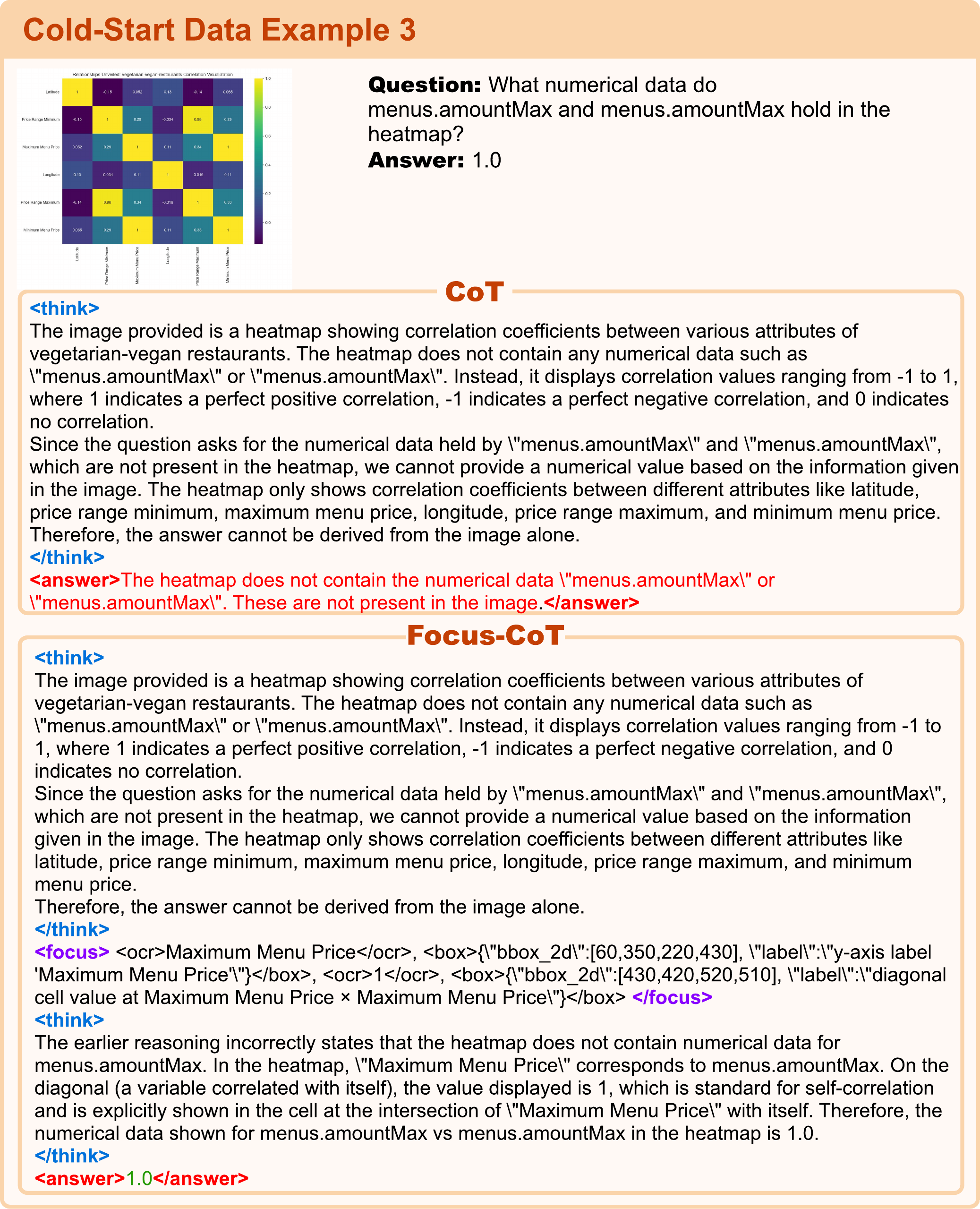}
\label{img:sft_case3}
\end{figure*}

\begin{figure*}[t]
\centering
\includegraphics[width=\textwidth]{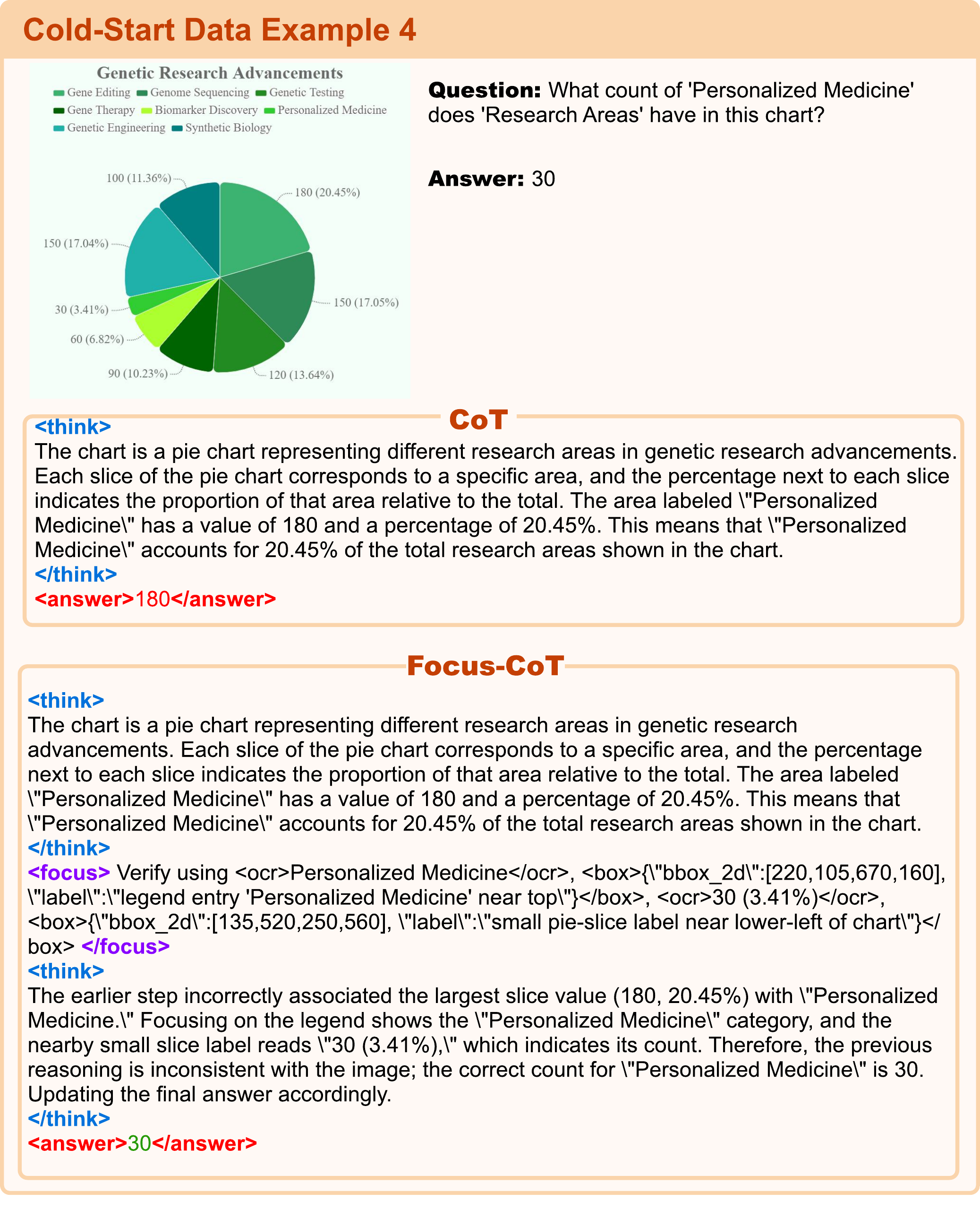}
\label{img:sft_case4}
\end{figure*}

\begin{figure*}[t]
\centering
\includegraphics[width=\textwidth]{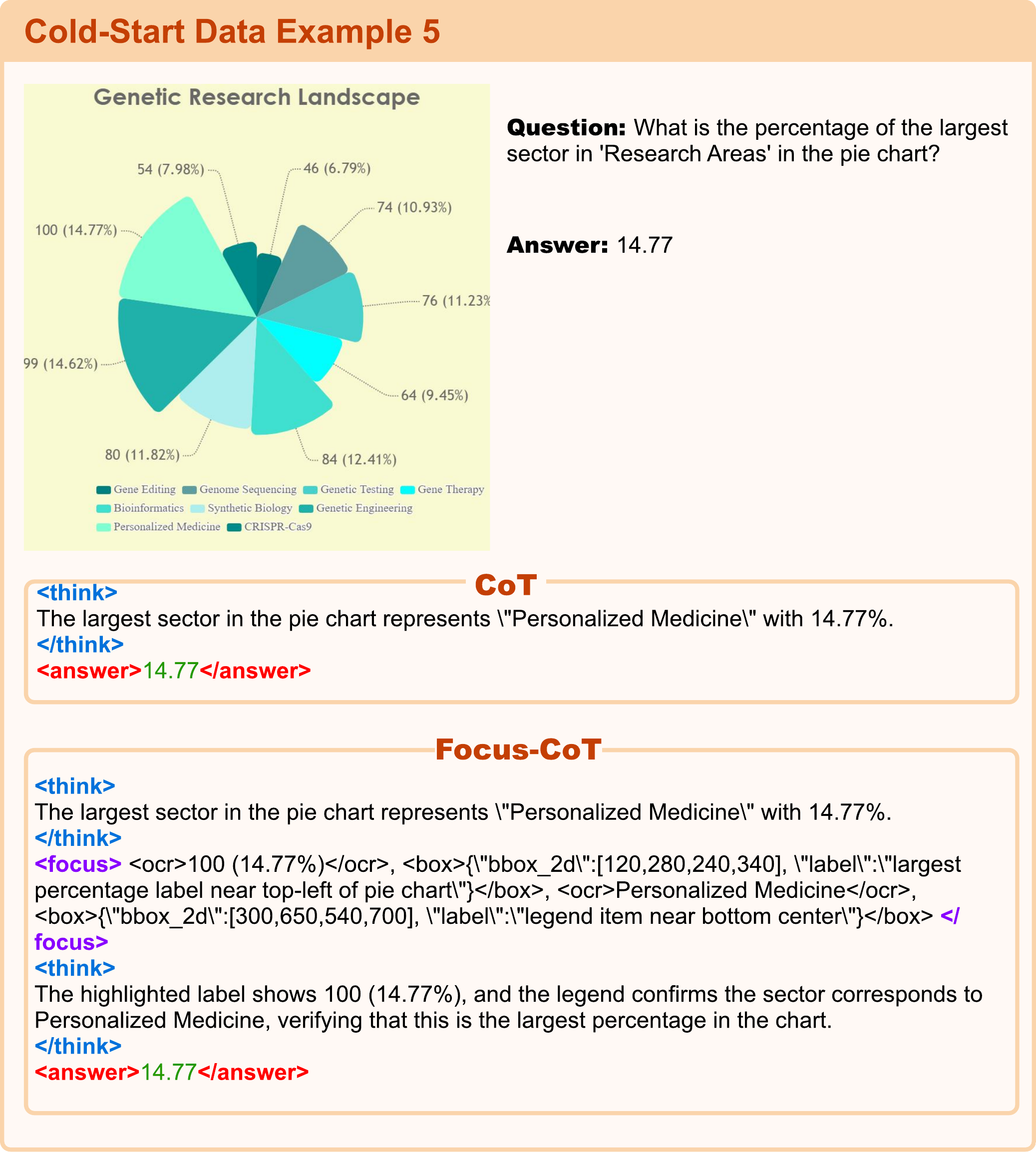}
\label{img:sft_case5}
\end{figure*}

\begin{figure*}[t]
\centering
\includegraphics[width=\textwidth]{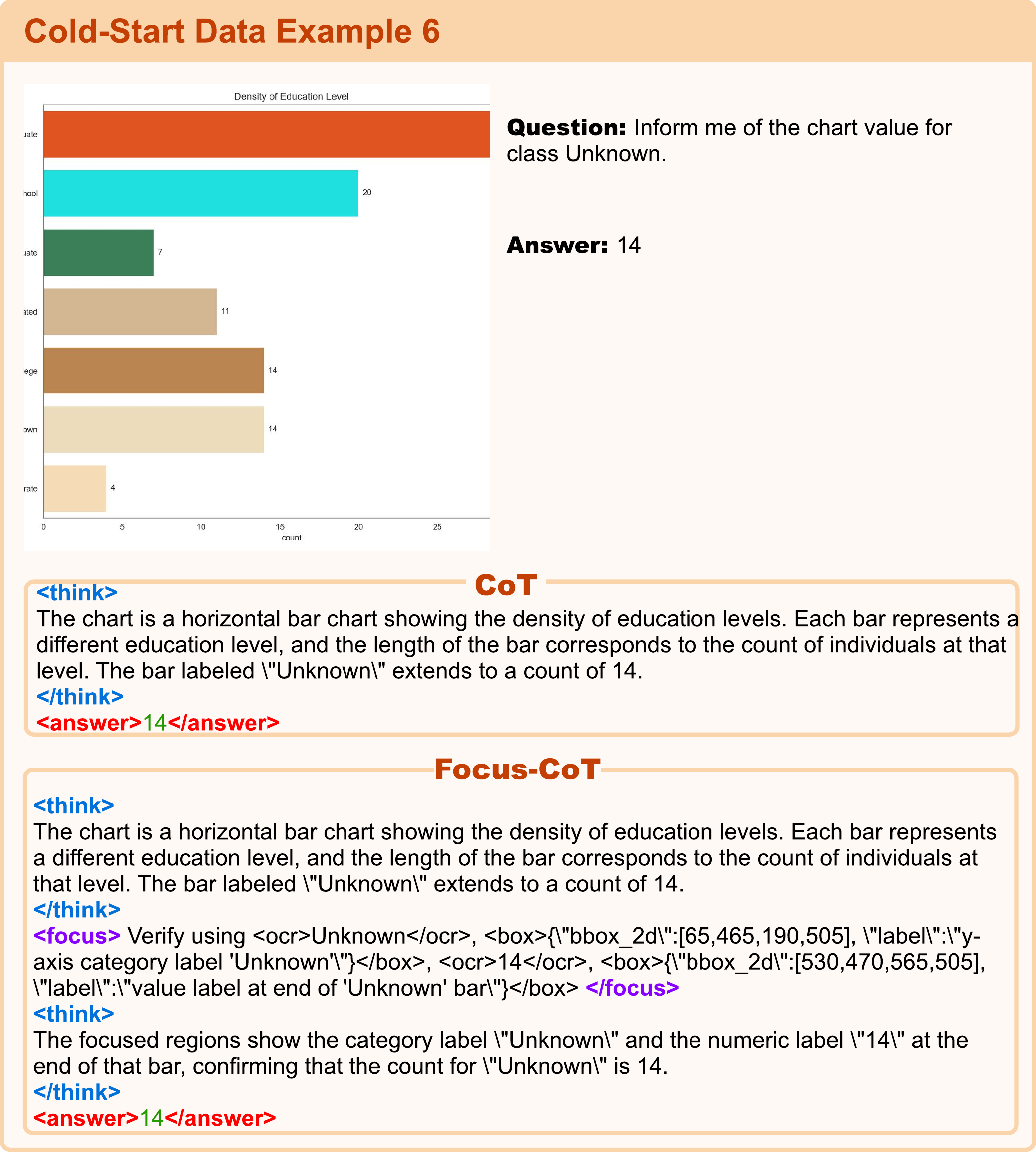}
\label{img:sft_case6}
\end{figure*}

\begin{figure*}[t]
\centering
\includegraphics[width=\textwidth]{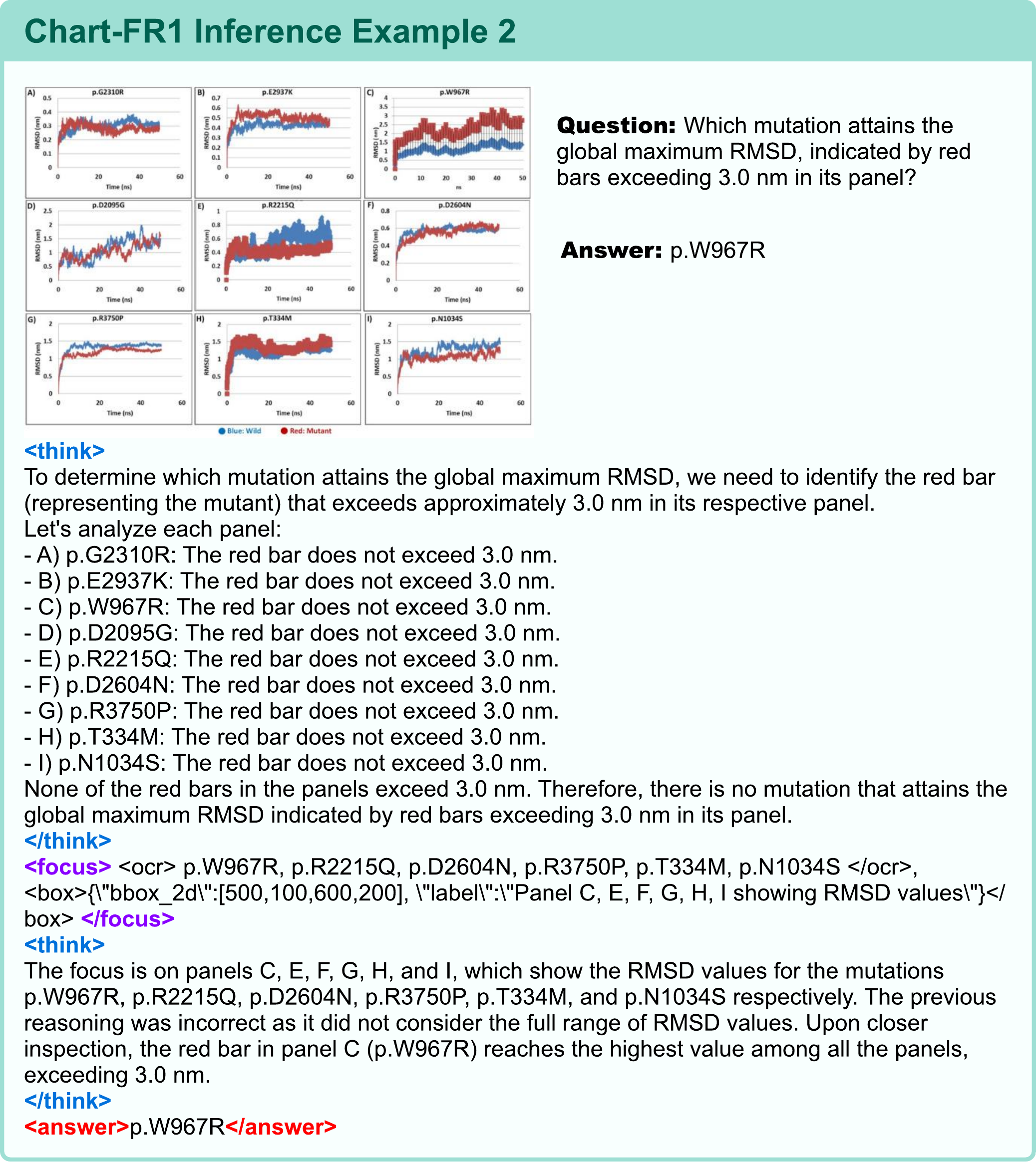}
\label{img:inference_case2}
\end{figure*}

\begin{figure*}[t]
\centering
\includegraphics[width=\textwidth]{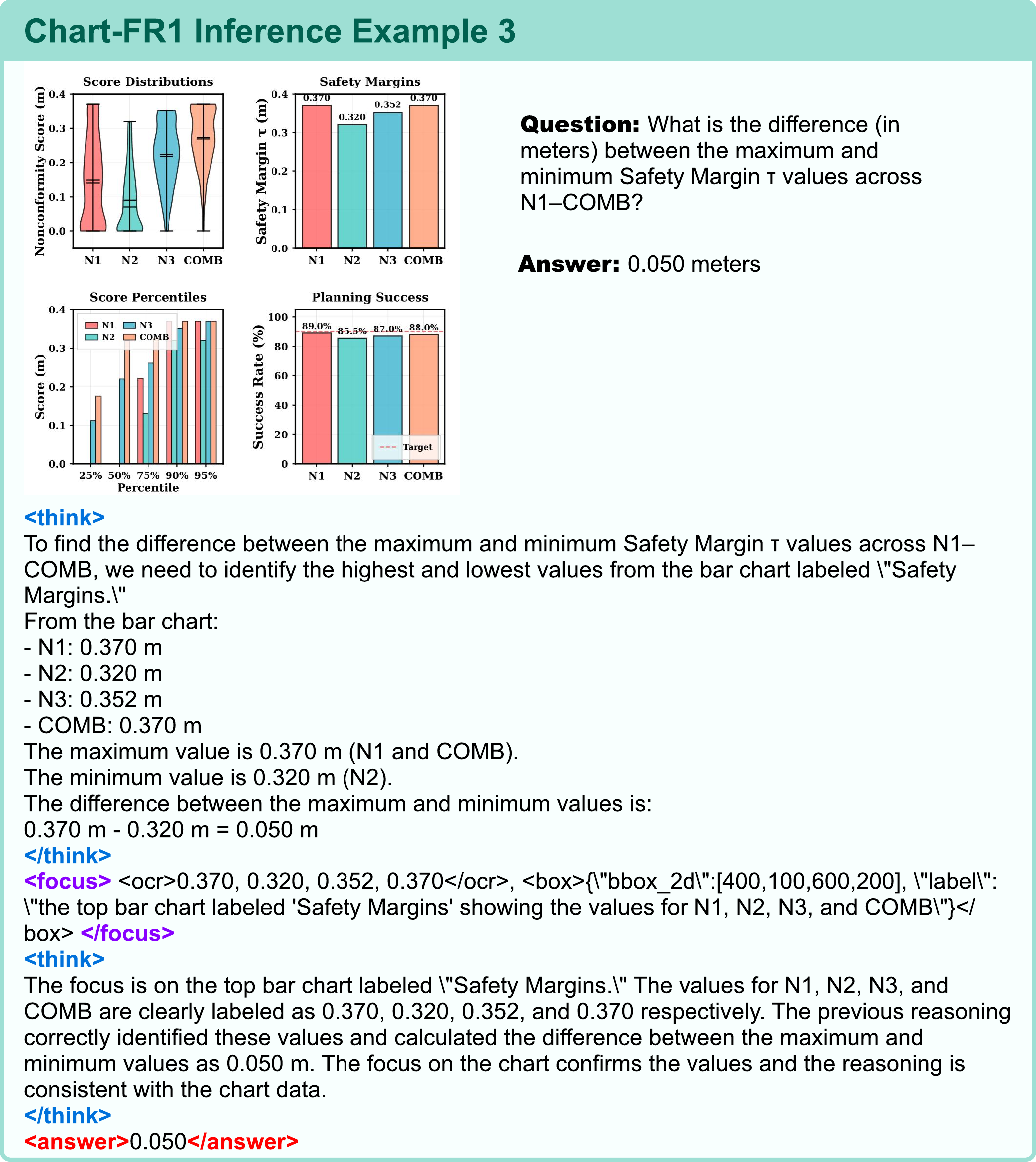}
\label{img:inference_case3}
\end{figure*}

\begin{figure*}[t]
\centering
\includegraphics[width=\textwidth]{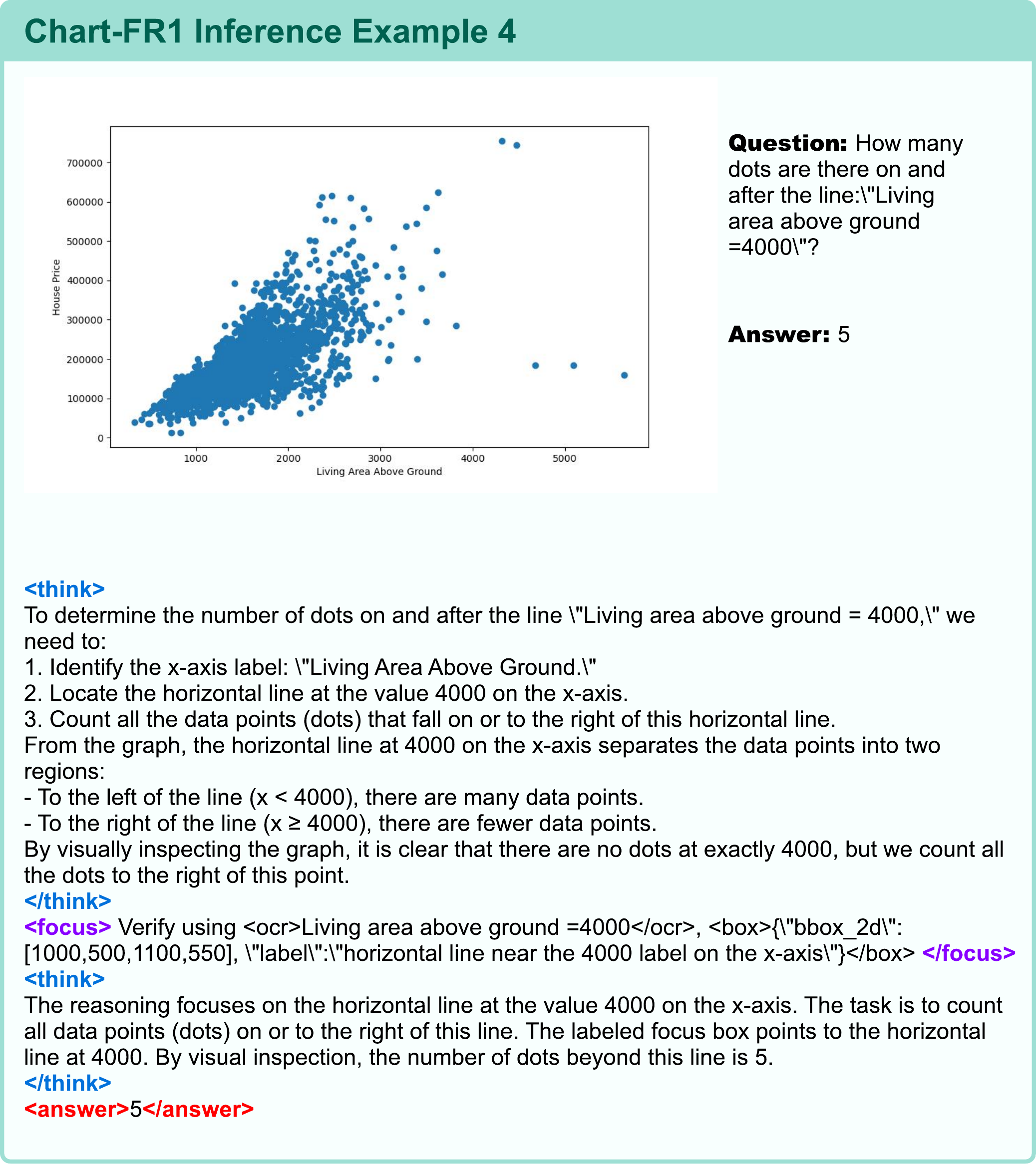}
\label{img:inference_case4}
\end{figure*}

\begin{figure*}[t]
\centering
\includegraphics[width=\textwidth]{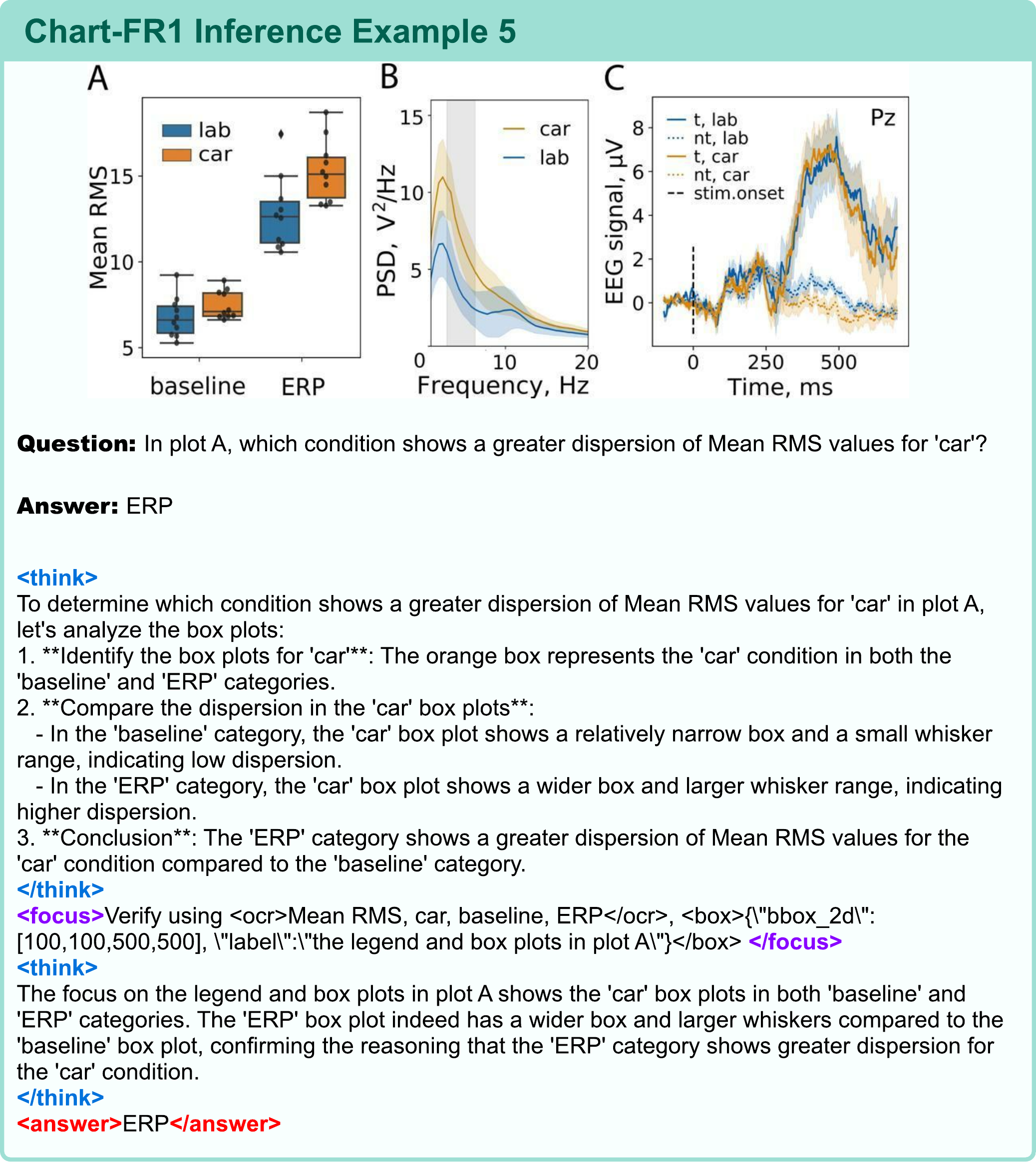}
\label{img:inference_case5}
\end{figure*}

\begin{figure*}[t]
\centering
\includegraphics[width=\textwidth]{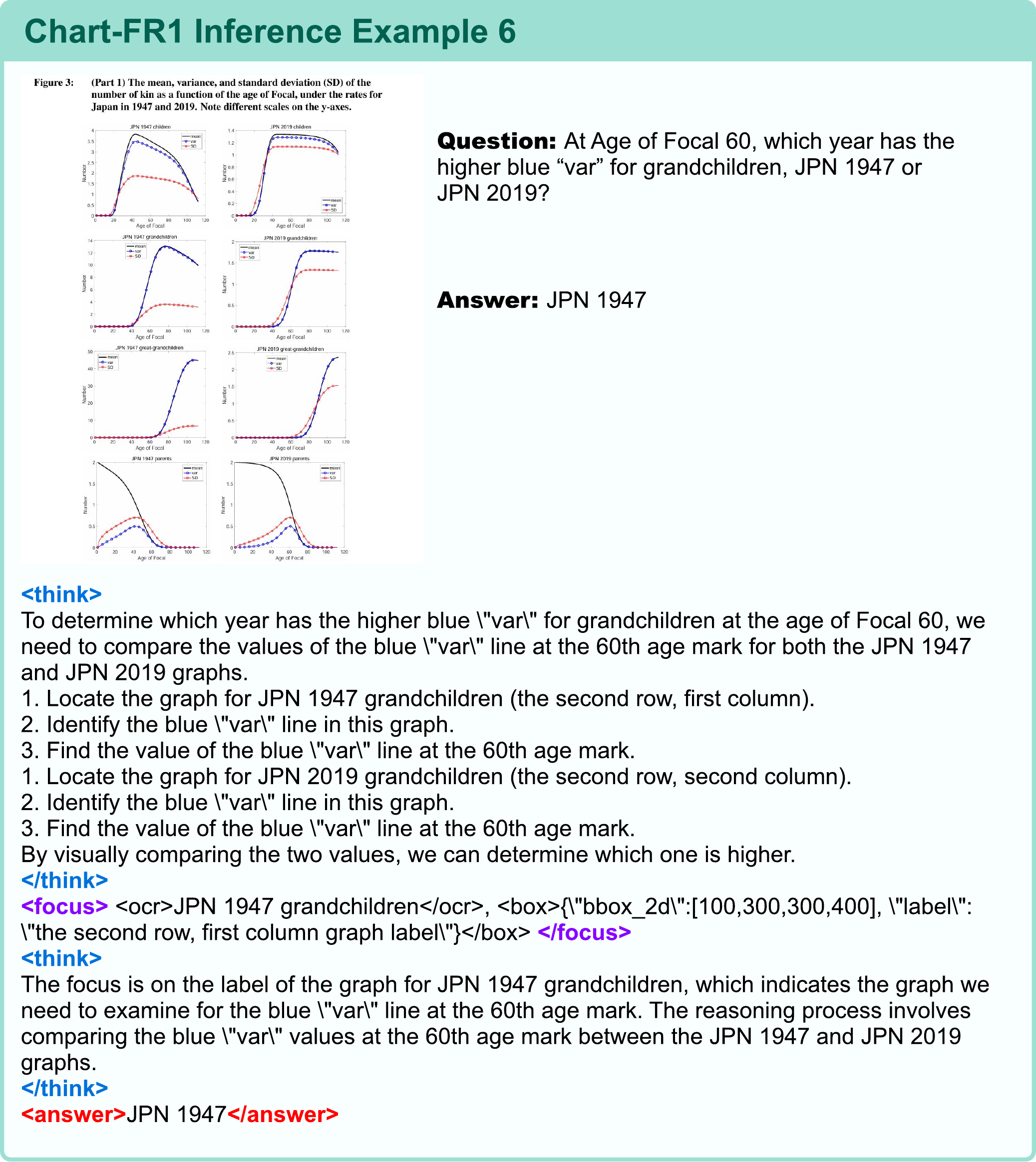}
\label{img:inference_case6}
\end{figure*}

\begin{figure*}[t]
\centering
\includegraphics[width=\textwidth]{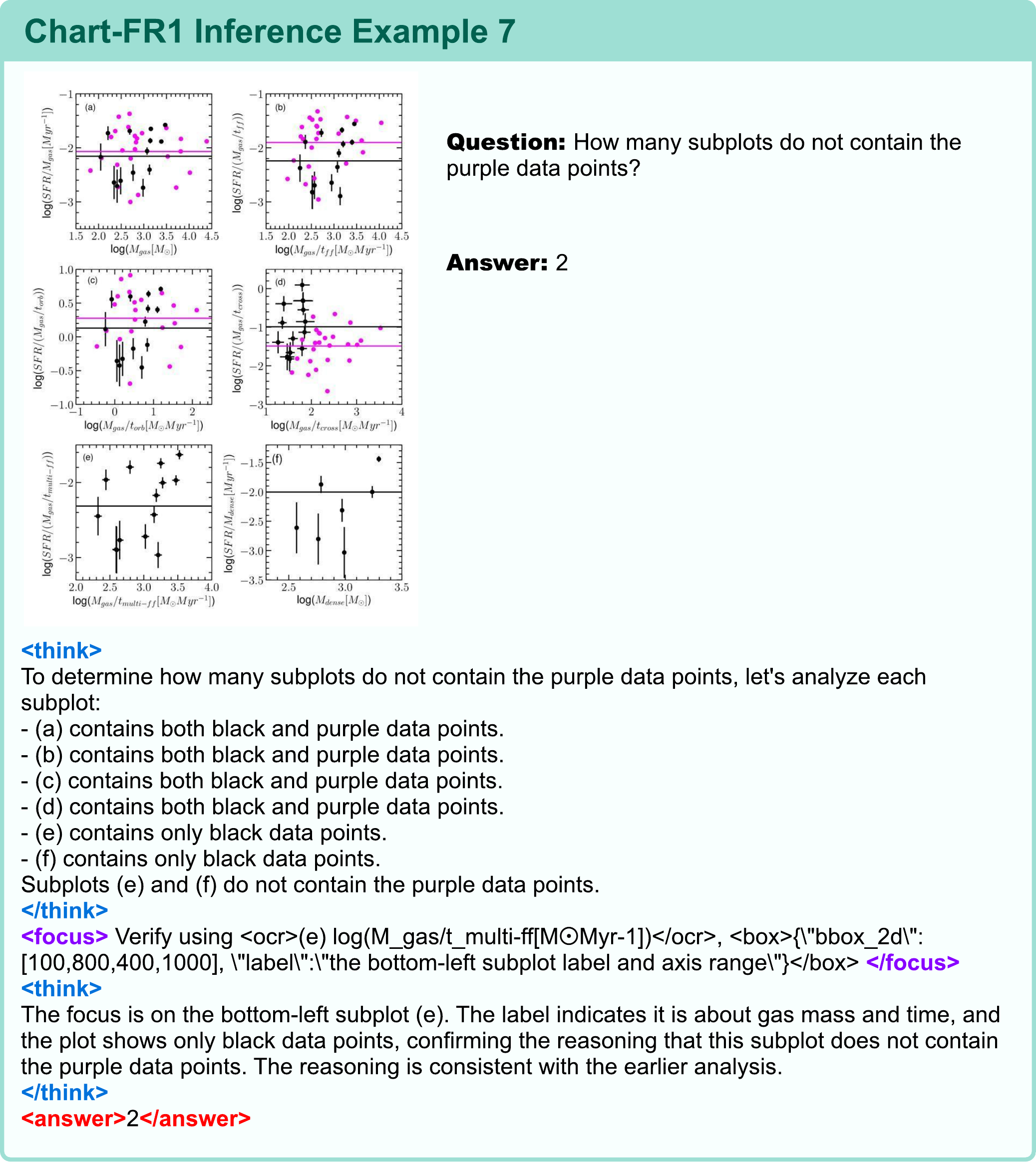}
\label{img:inference_case7}
\end{figure*}

\begin{figure*}[t]
\centering
\includegraphics[width=\textwidth]{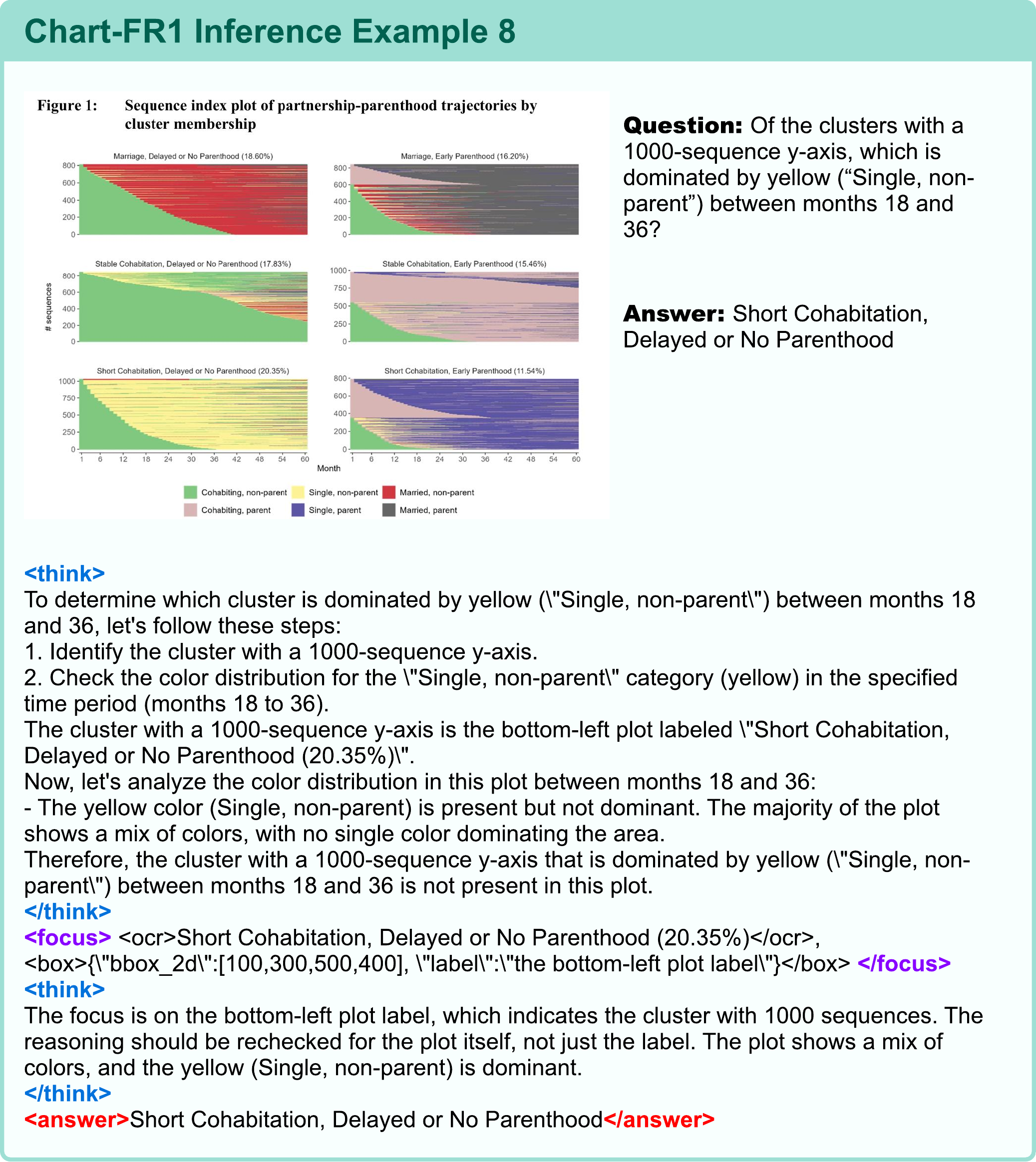}
\label{img:inference_case8}
\end{figure*}

\begin{figure*}[t]
\centering
\includegraphics[width=\textwidth]{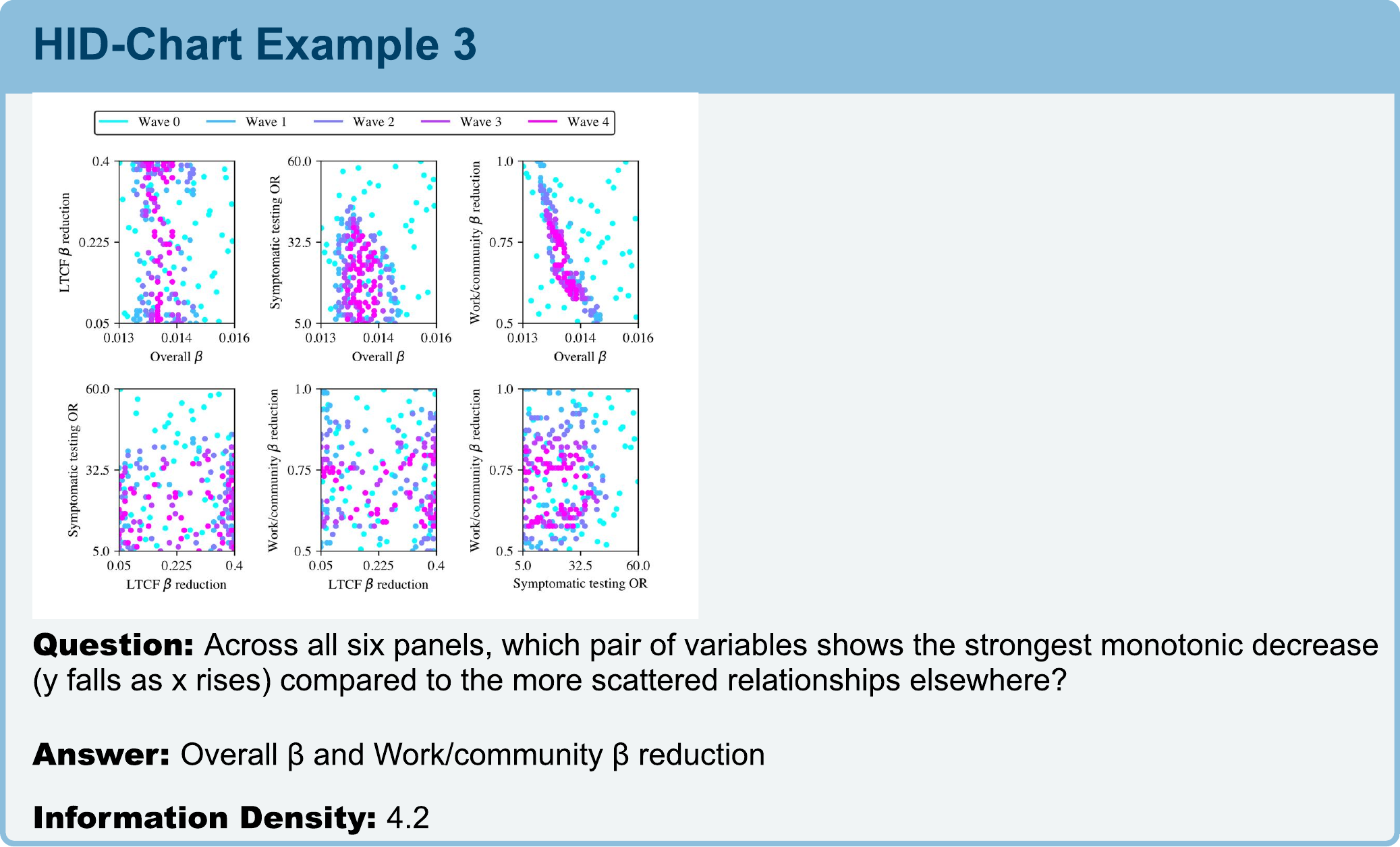}
\label{img:benchmark_case3}
\end{figure*}

\begin{figure*}[t]
\centering
\includegraphics[width=\textwidth]{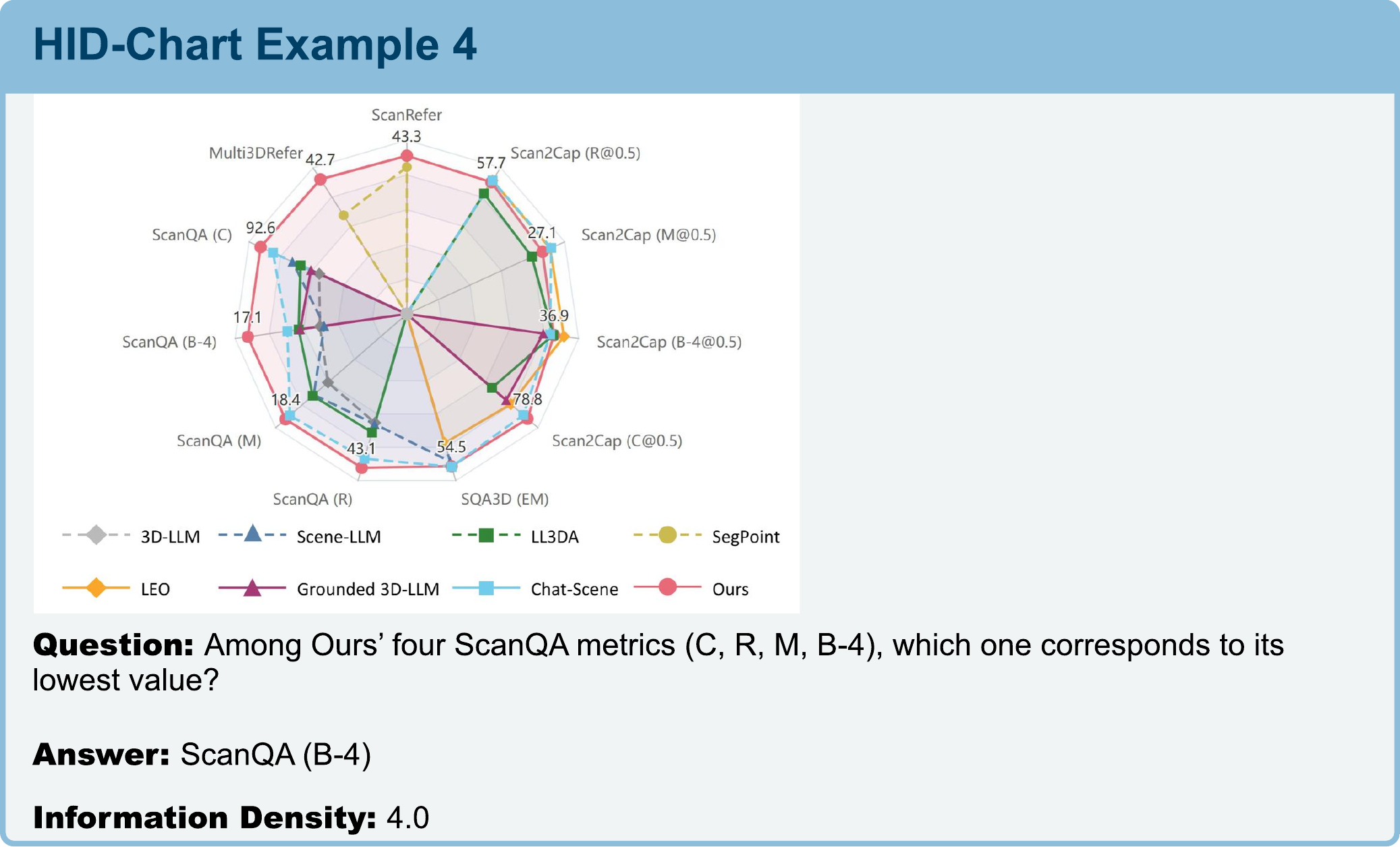}
\label{img:benchmark_case4}
\end{figure*}

\begin{figure*}[t]
\centering
\includegraphics[width=\textwidth]{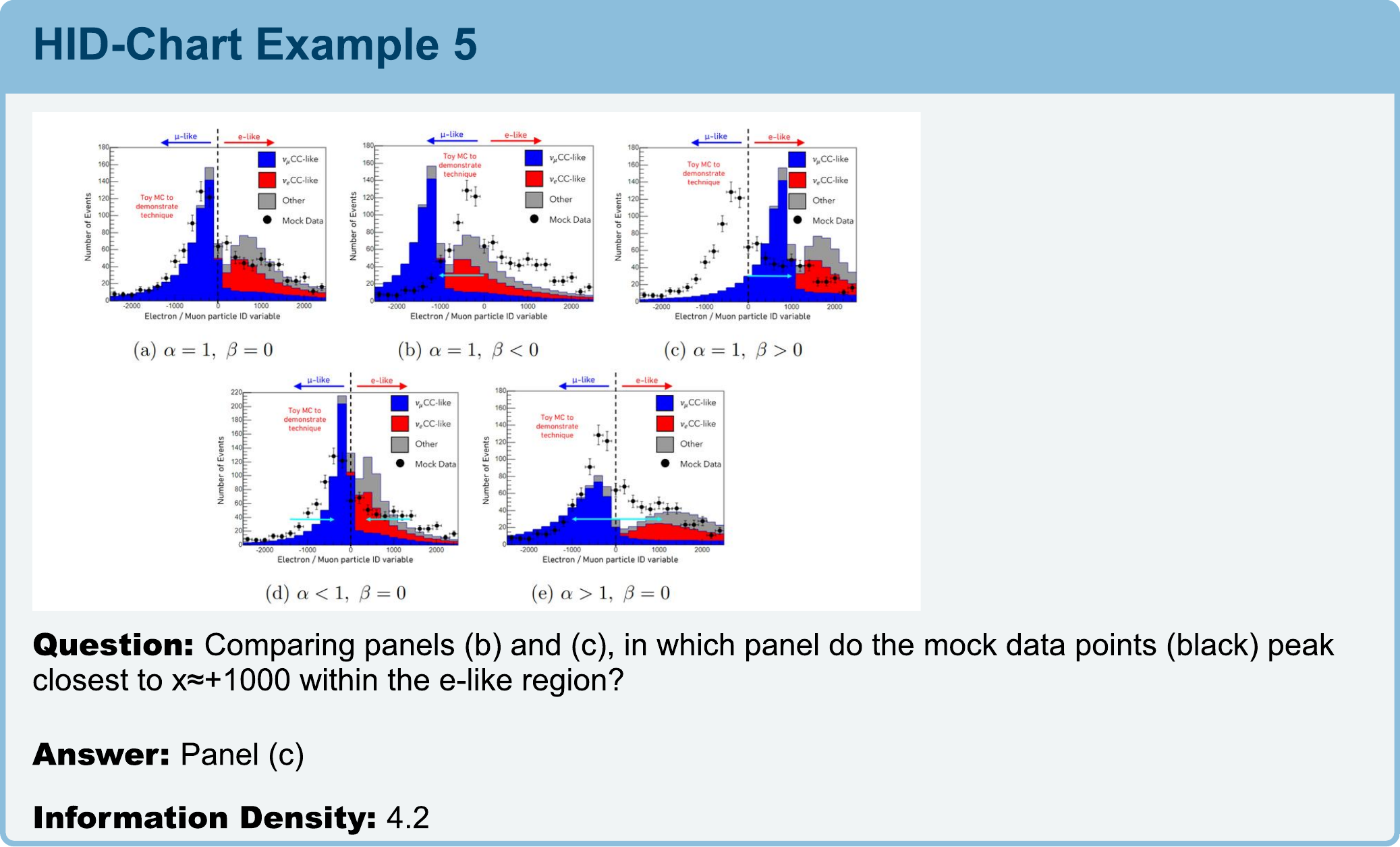}
\label{img:benchmark_case5}
\end{figure*}

\begin{figure*}[t]
\centering
\includegraphics[width=\textwidth]{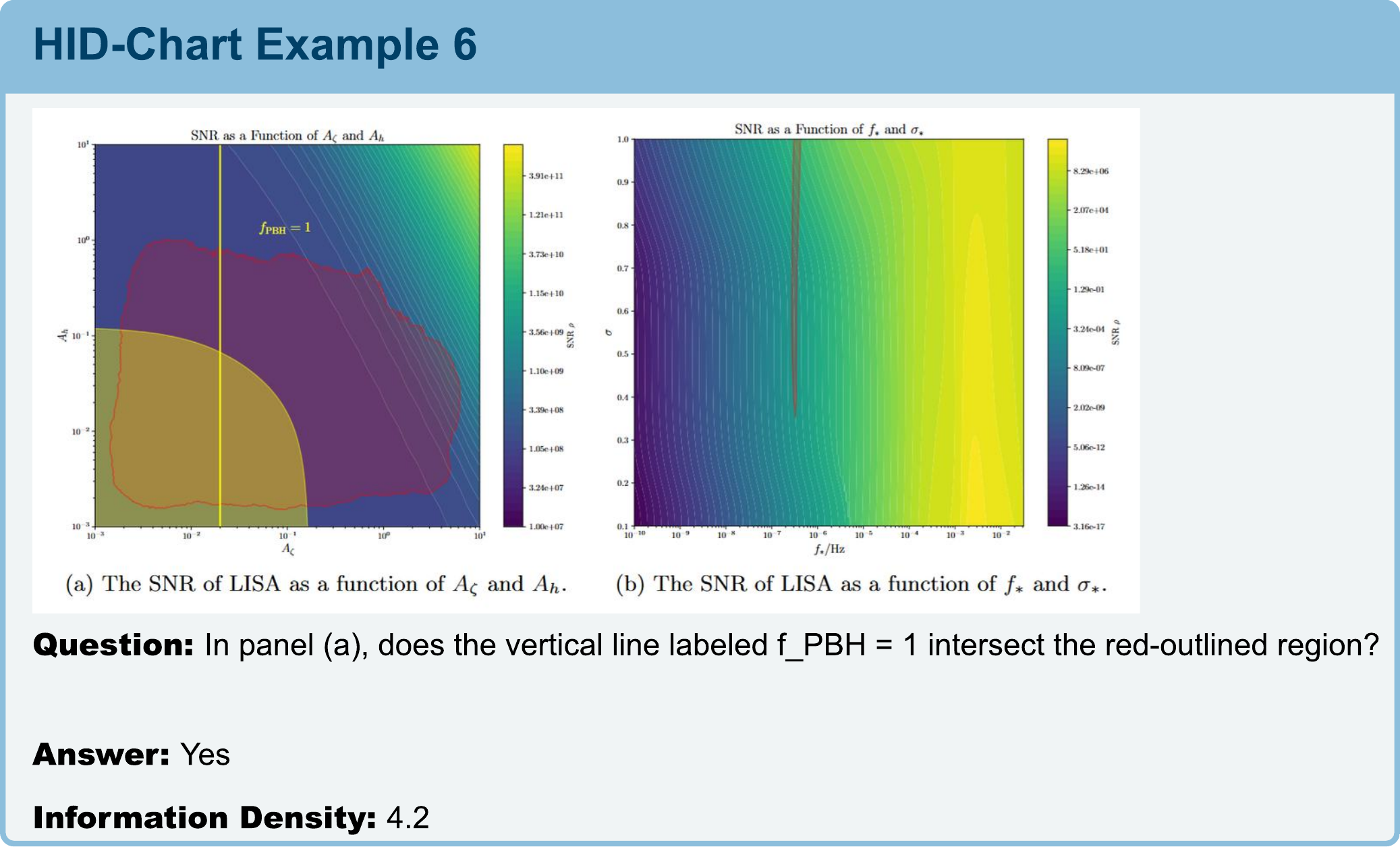}
\label{img:benchmark_case6}
\end{figure*}

\begin{figure*}[t]
\centering
\includegraphics[width=\textwidth]{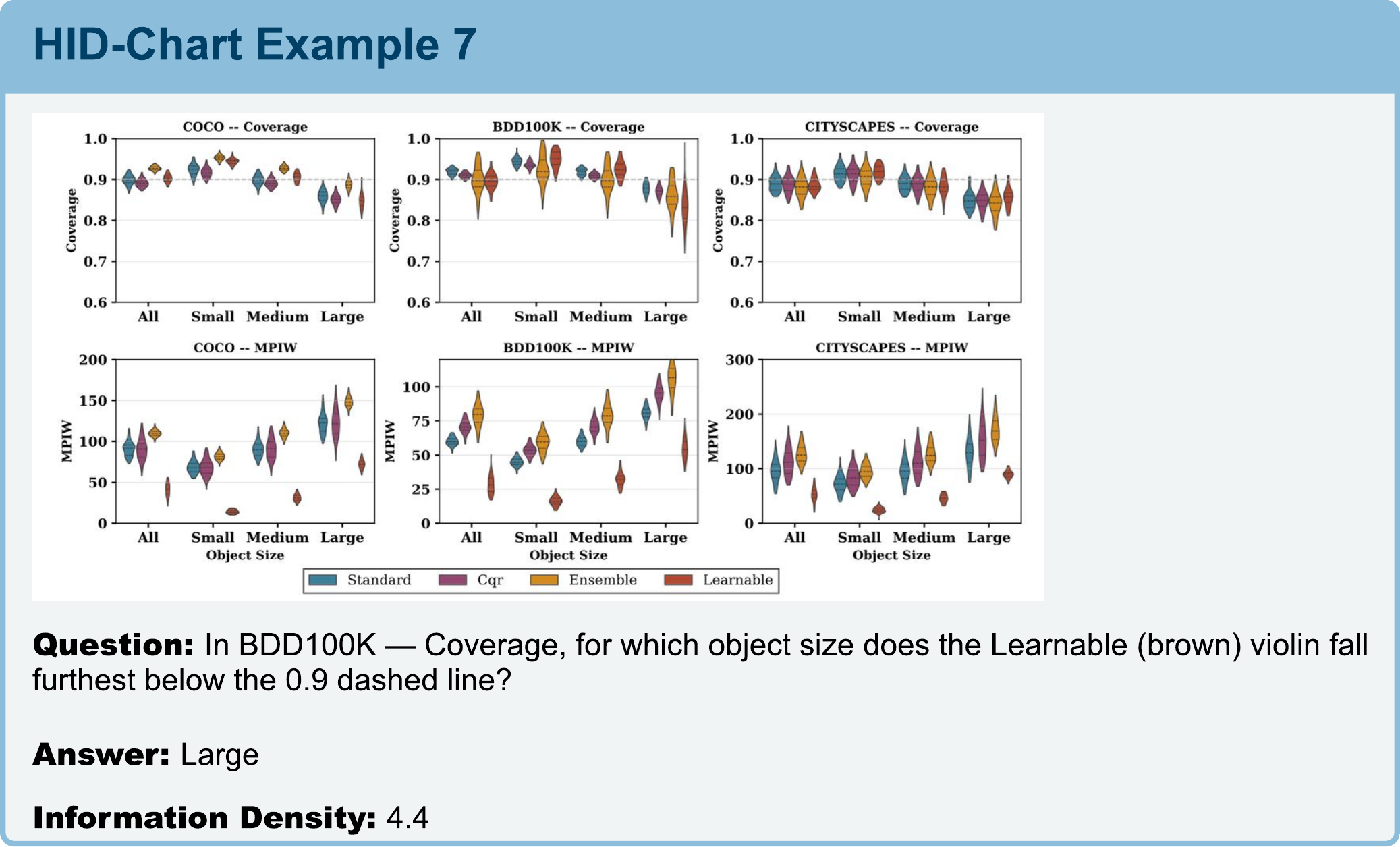}
\label{img:benchmark_case7}
\end{figure*}

\begin{figure*}[t]
\centering
\includegraphics[width=\textwidth]{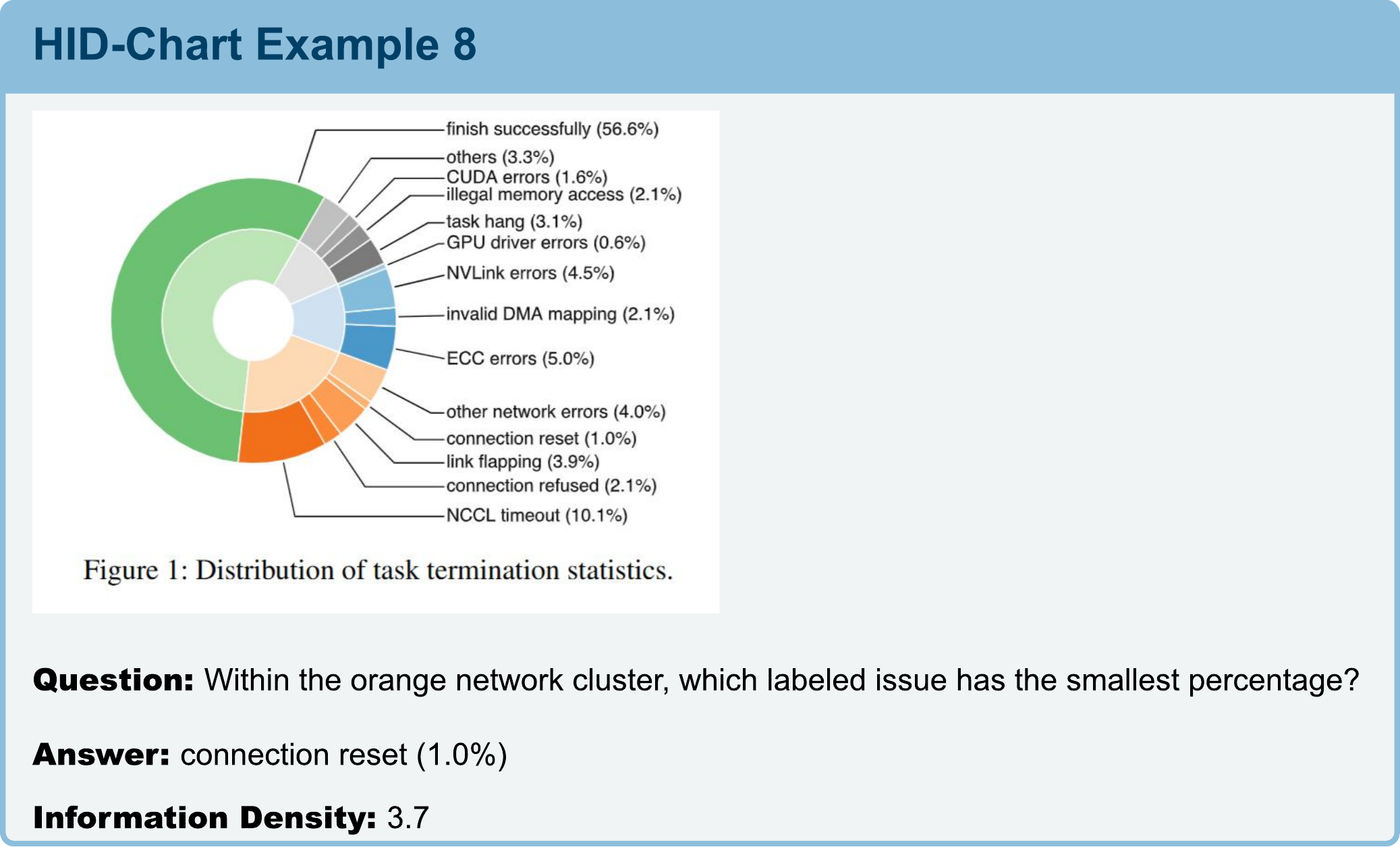}
\label{img:benchmark_case8}
\end{figure*}

\begin{figure*}[t]
\centering
\includegraphics[width=\textwidth]{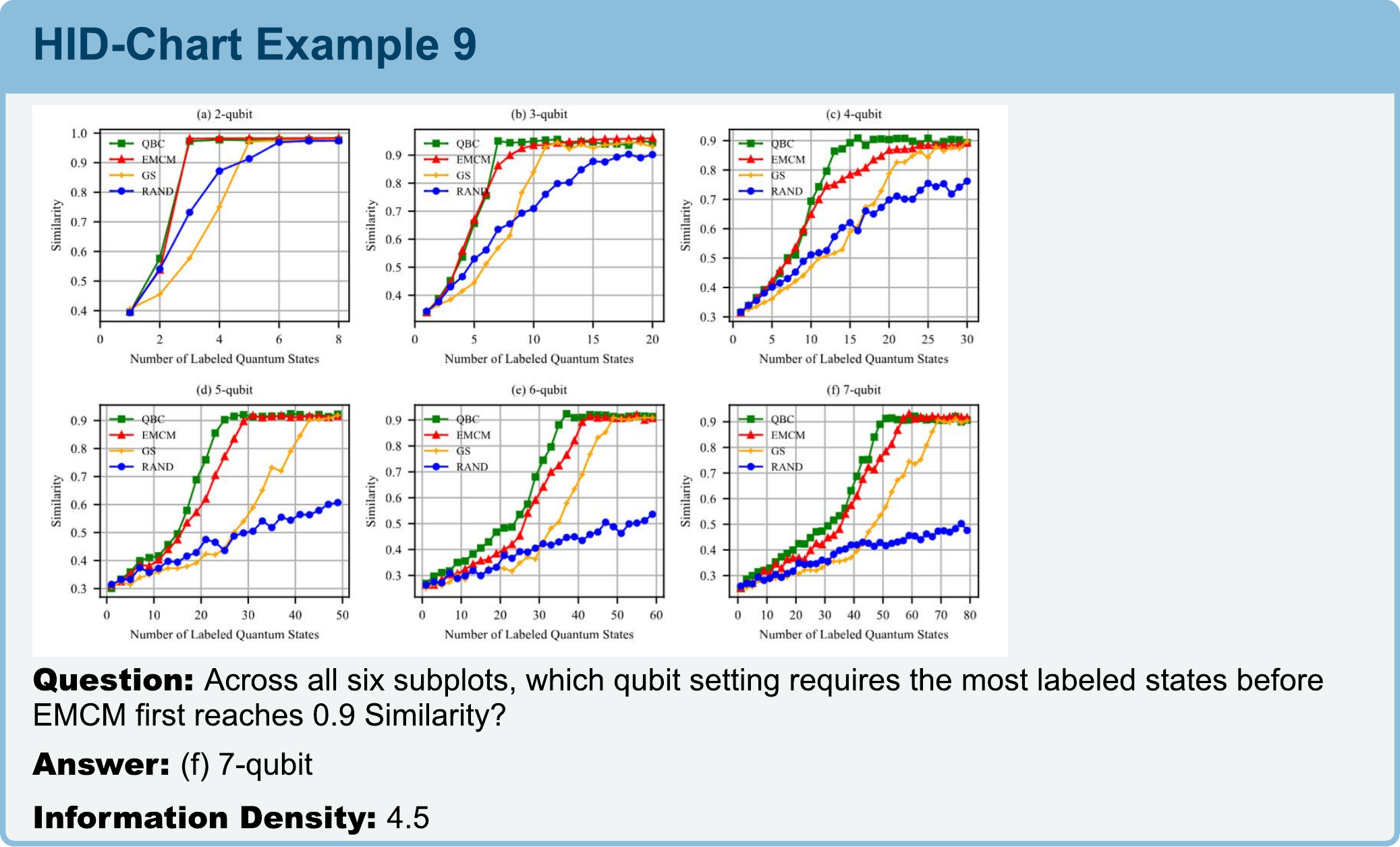}
\label{img:benchmark_case9}
\end{figure*}

\begin{figure*}[t]
\centering
\includegraphics[width=\textwidth]{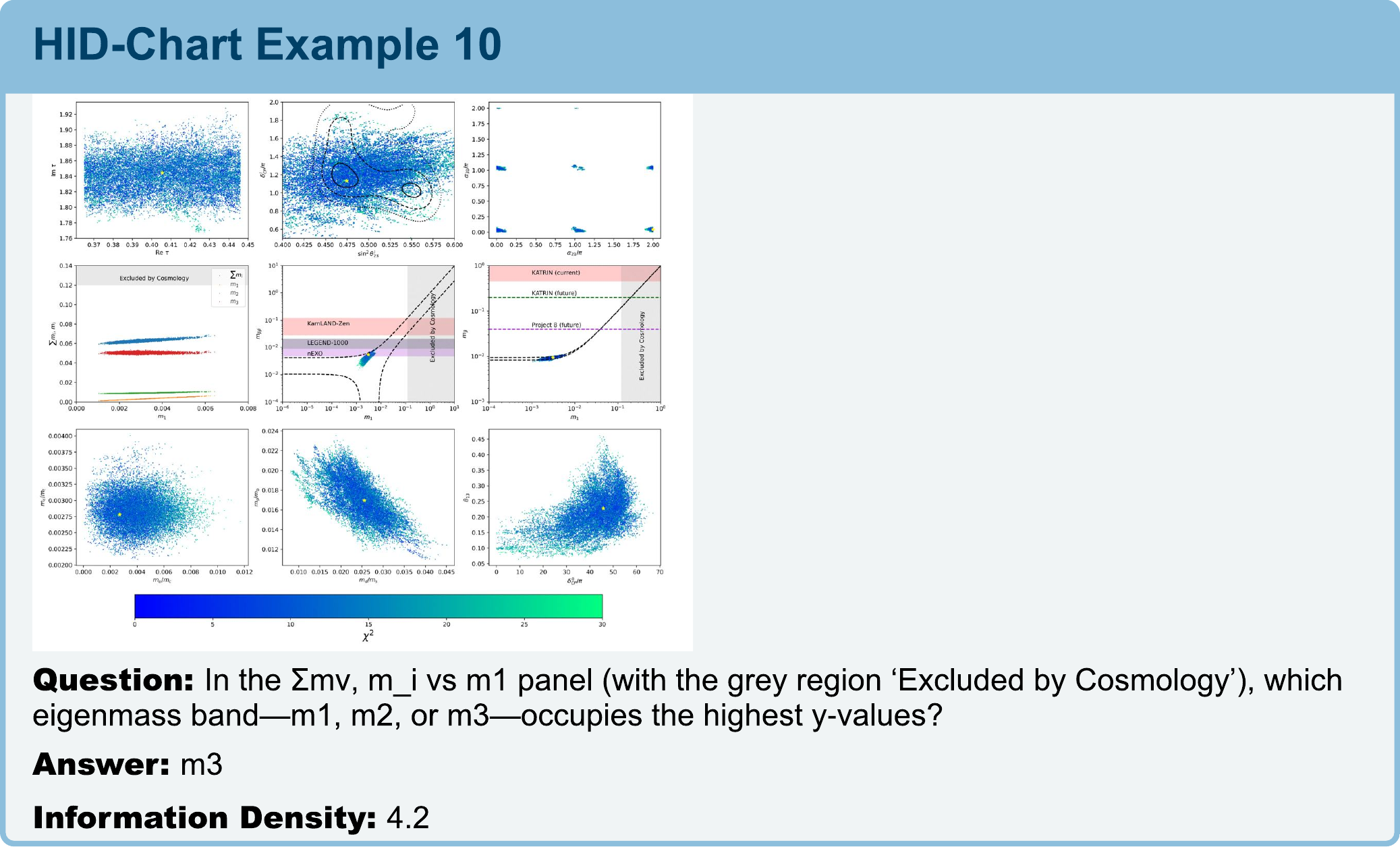}
\label{img:benchmark_case10}
\end{figure*}

\begin{figure*}[t]
\centering
\includegraphics[width=\textwidth]{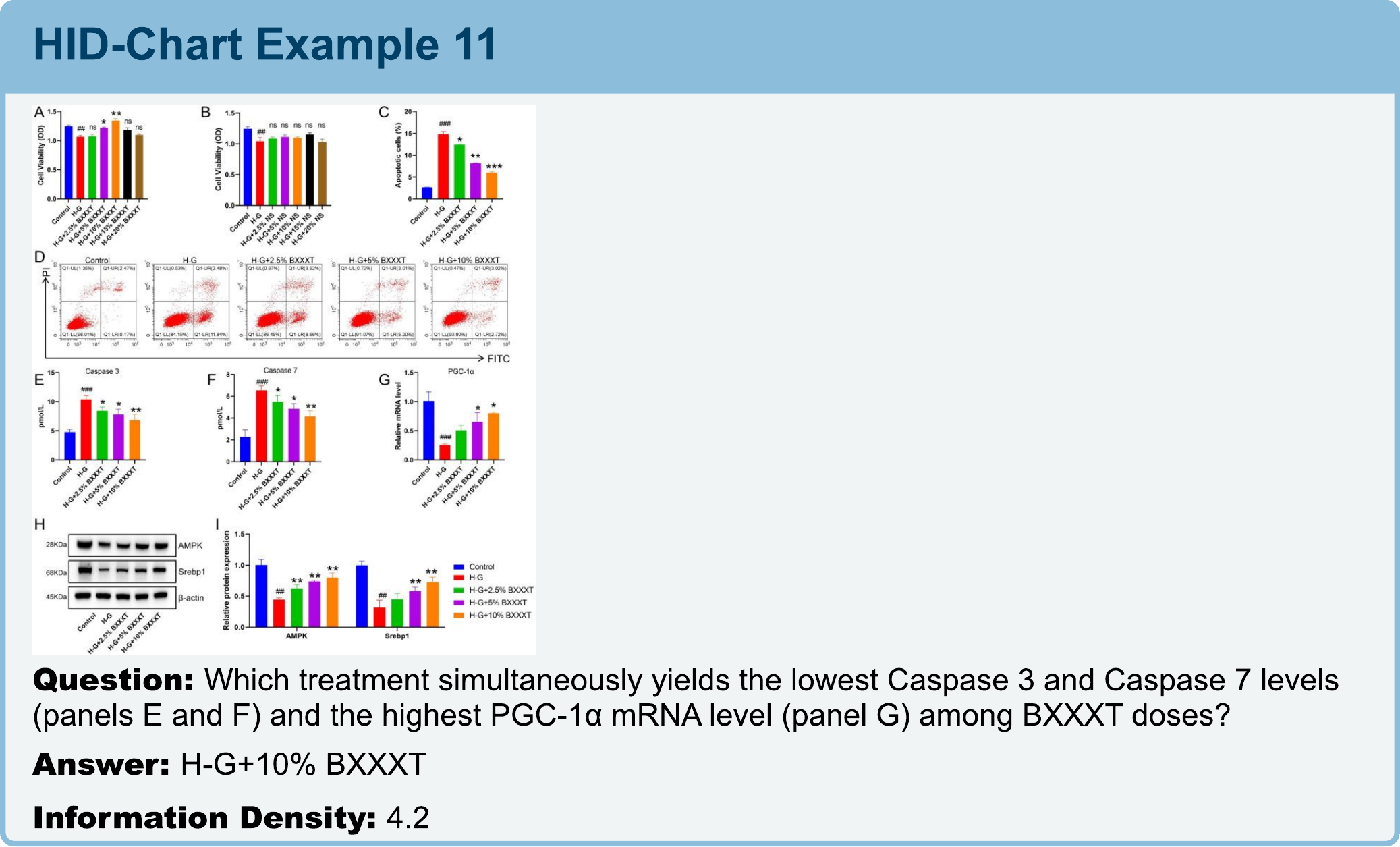}
\label{img:benchmark_case11}
\end{figure*}

\begin{figure*}[t]
\centering
\includegraphics[width=\textwidth]{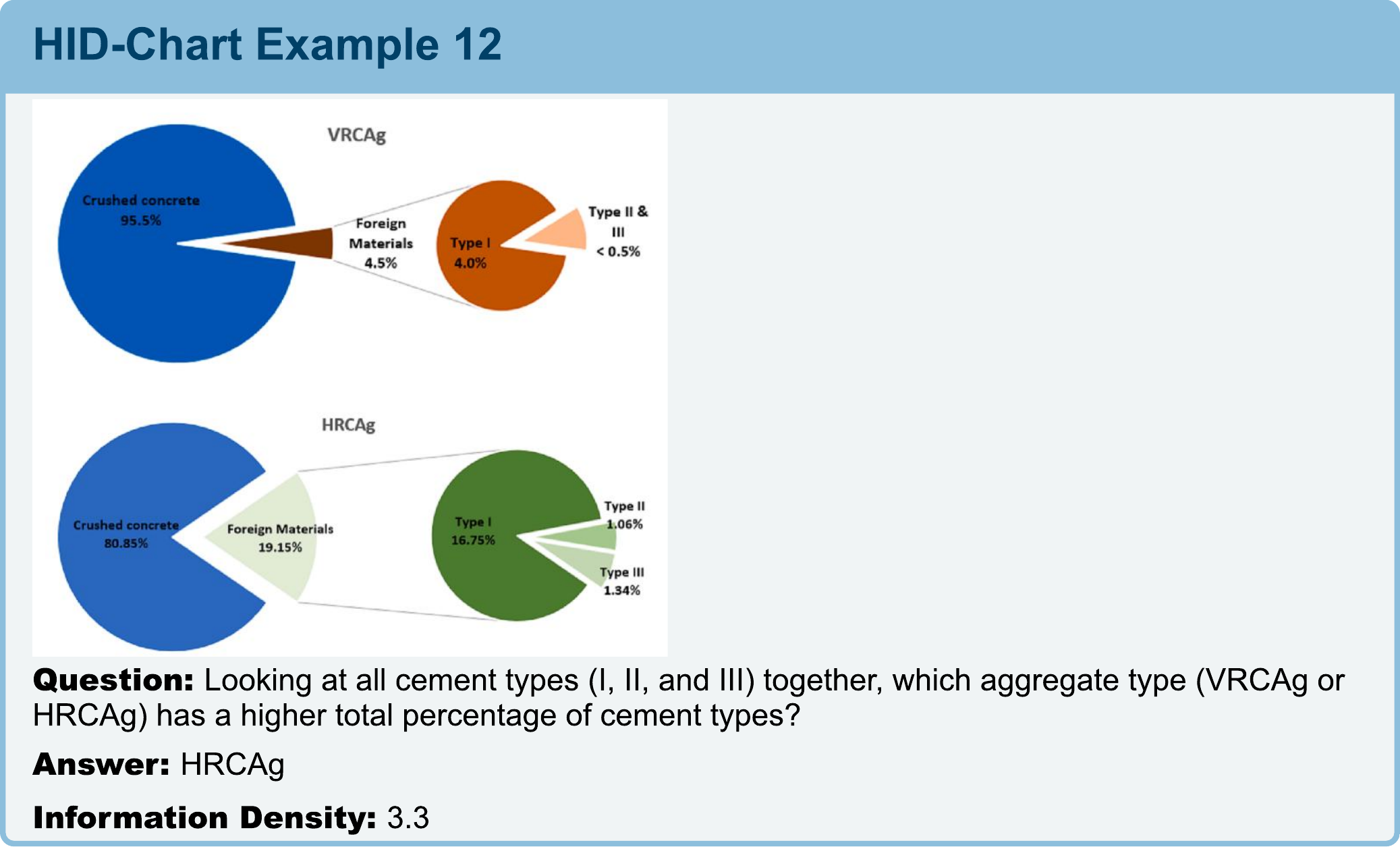}
\label{img:benchmark_case12}
\end{figure*}

\begin{figure*}[t]
\centering
\includegraphics[width=\textwidth]{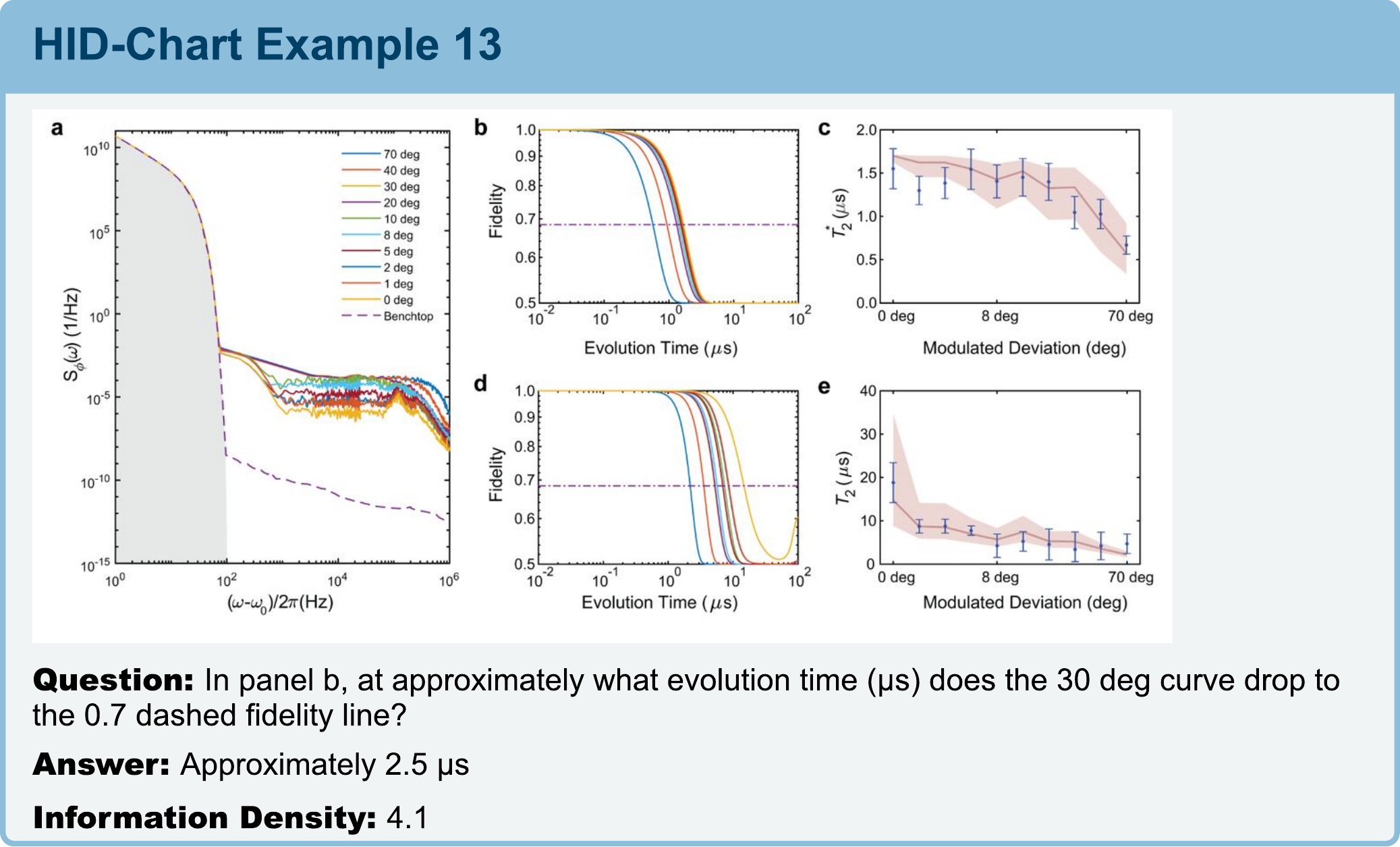}
\label{img:benchmark_case13}
\end{figure*}

\begin{figure*}[t]
\centering
\includegraphics[width=\textwidth]{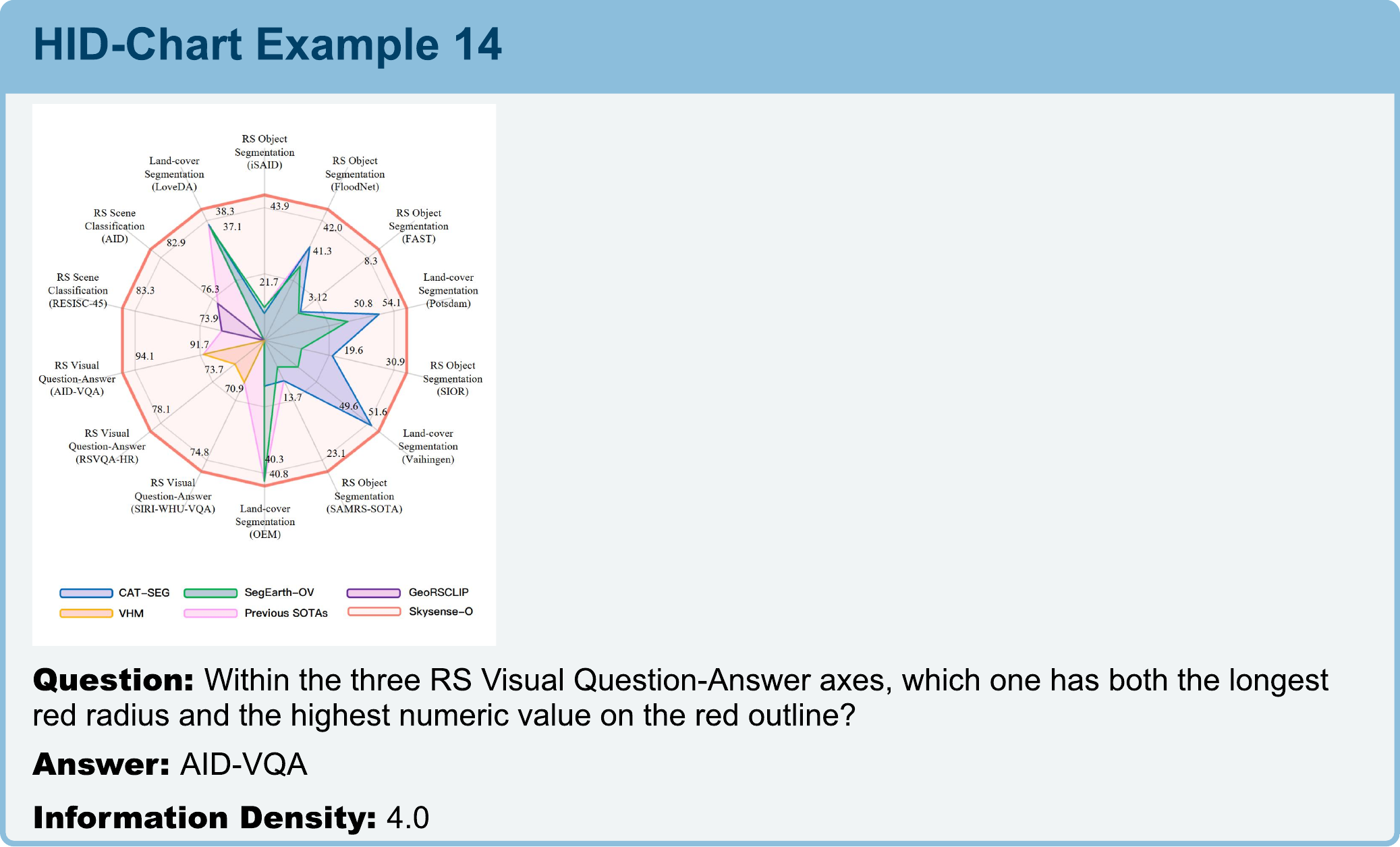}
\label{img:benchmark_case14}
\end{figure*}

\begin{figure*}[t]
\centering
\includegraphics[width=\textwidth]{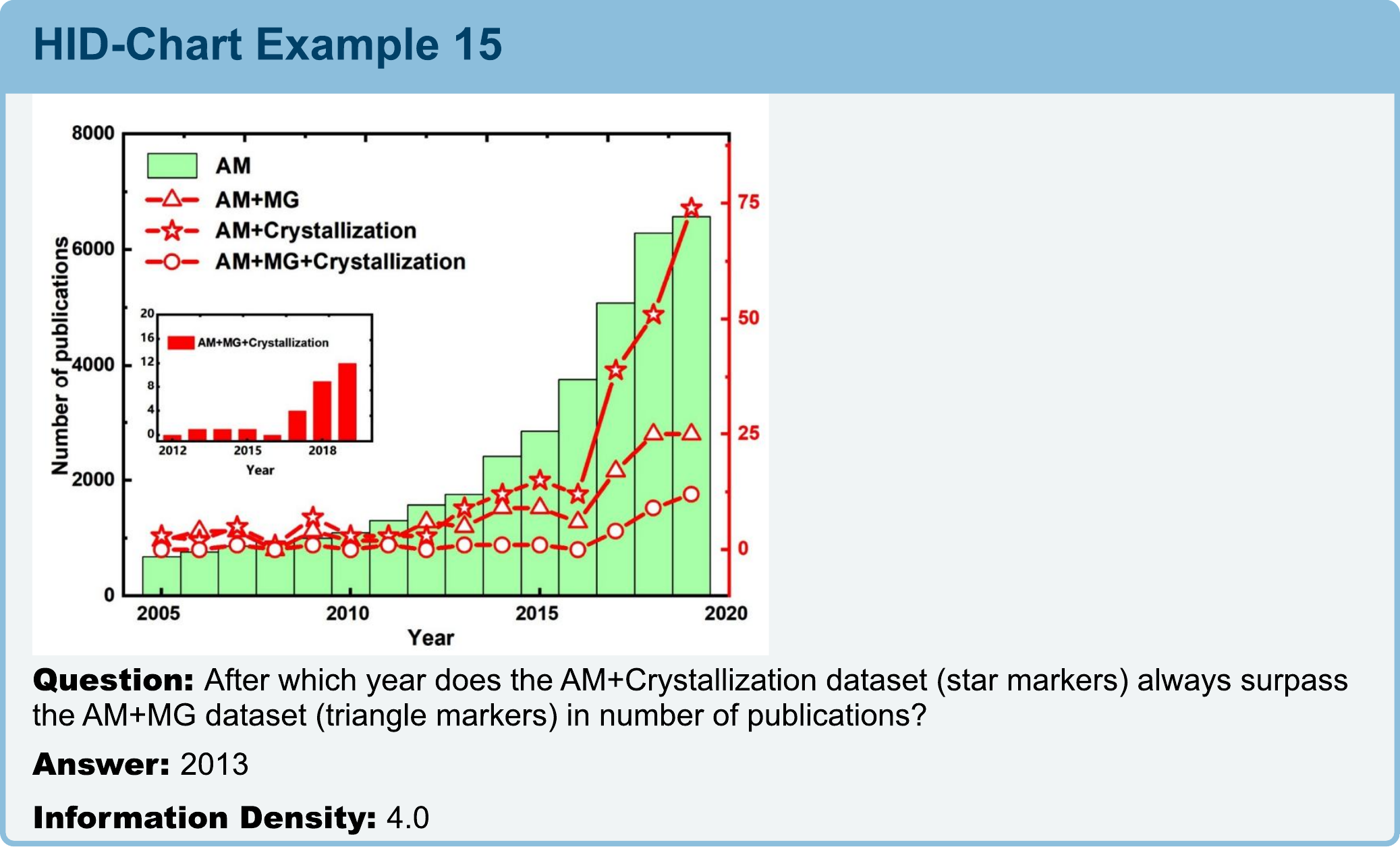}
\label{img:benchmark_case15}
\end{figure*}

\begin{figure*}[t]
\centering
\includegraphics[width=\textwidth]{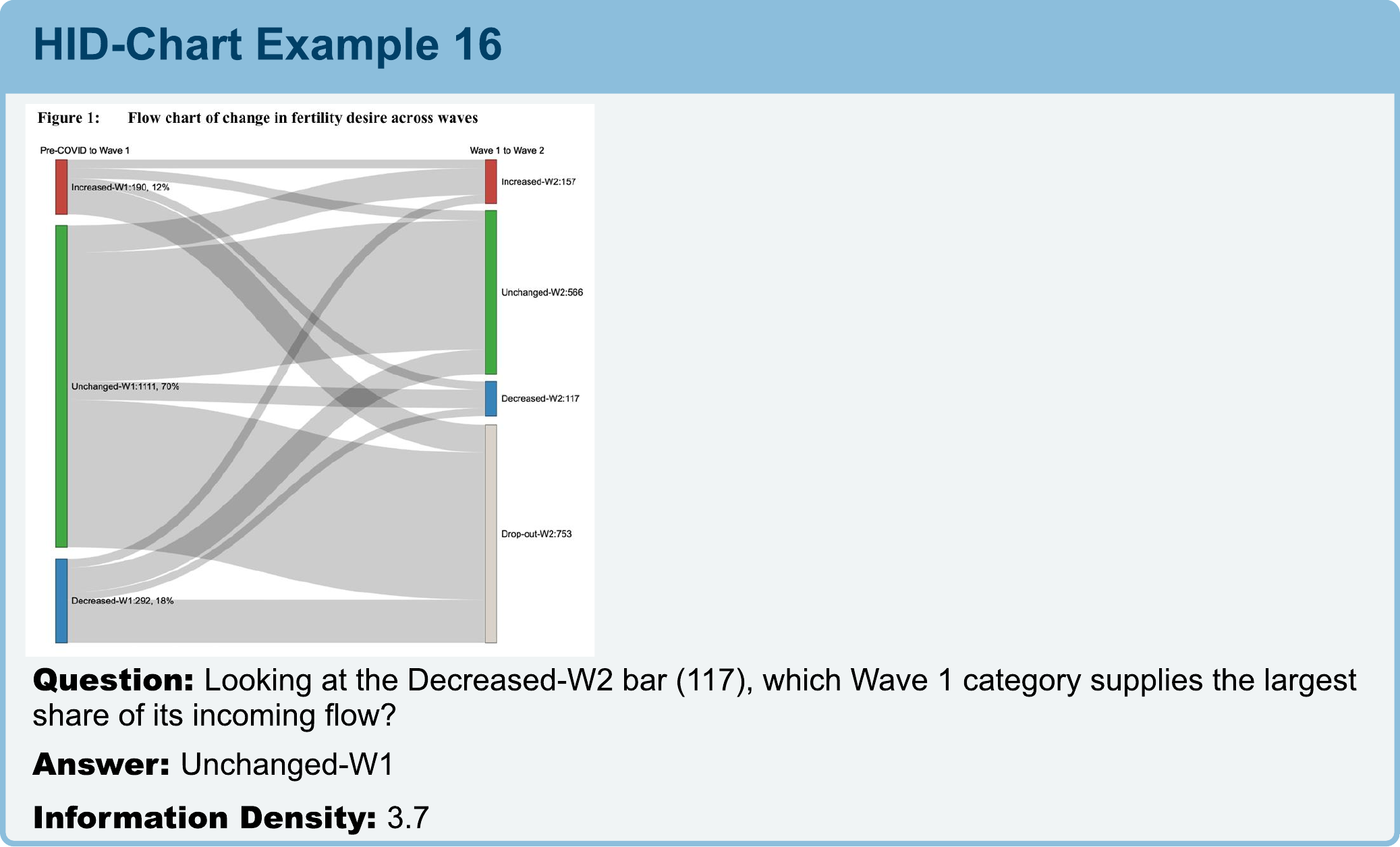}
\label{img:benchmark_case16}
\end{figure*}

\begin{figure*}[t]
\centering
\includegraphics[width=\textwidth]{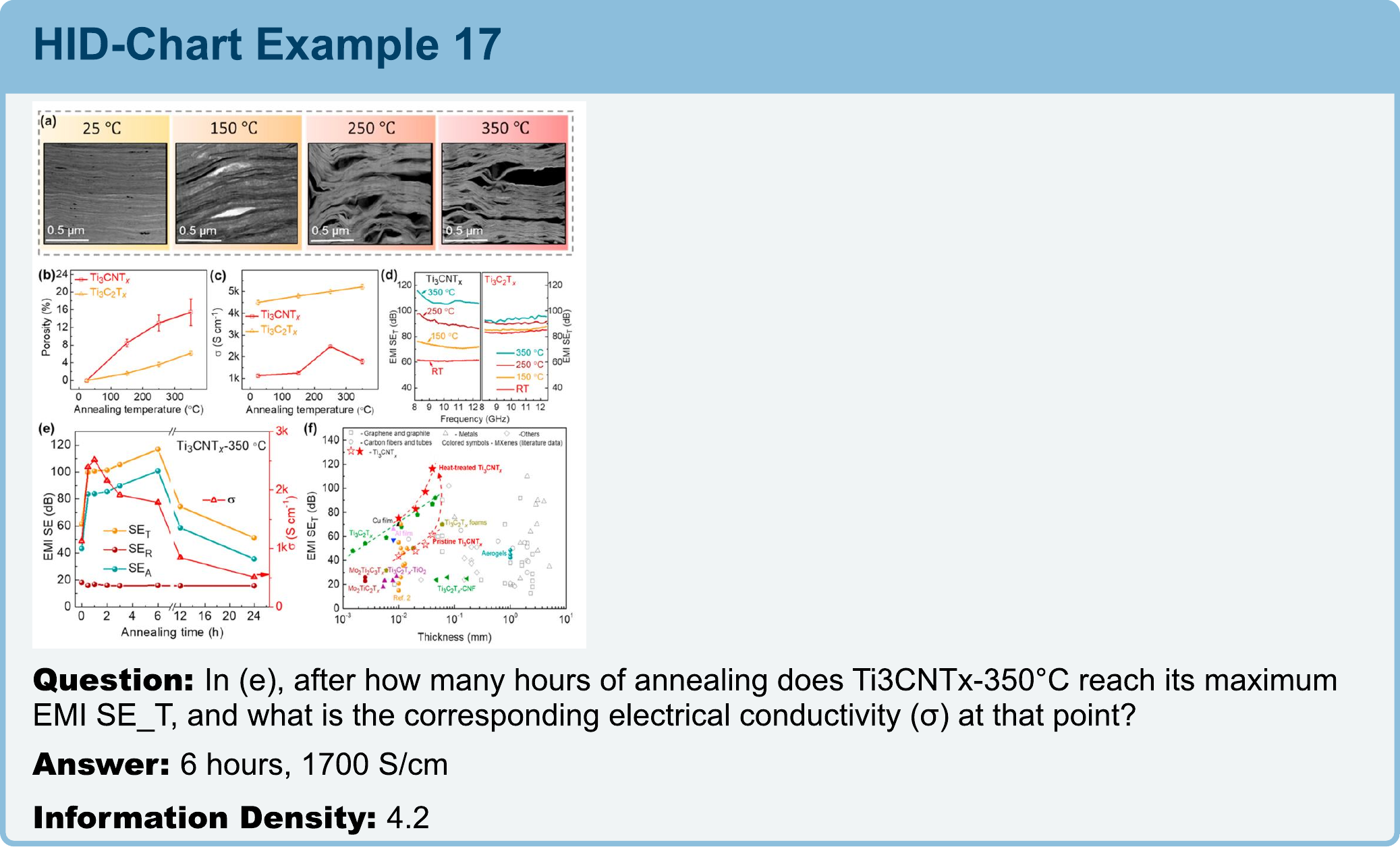}
\label{img:benchmark_case17}
\end{figure*}

\begin{figure*}[t]
\centering
\includegraphics[width=\textwidth]{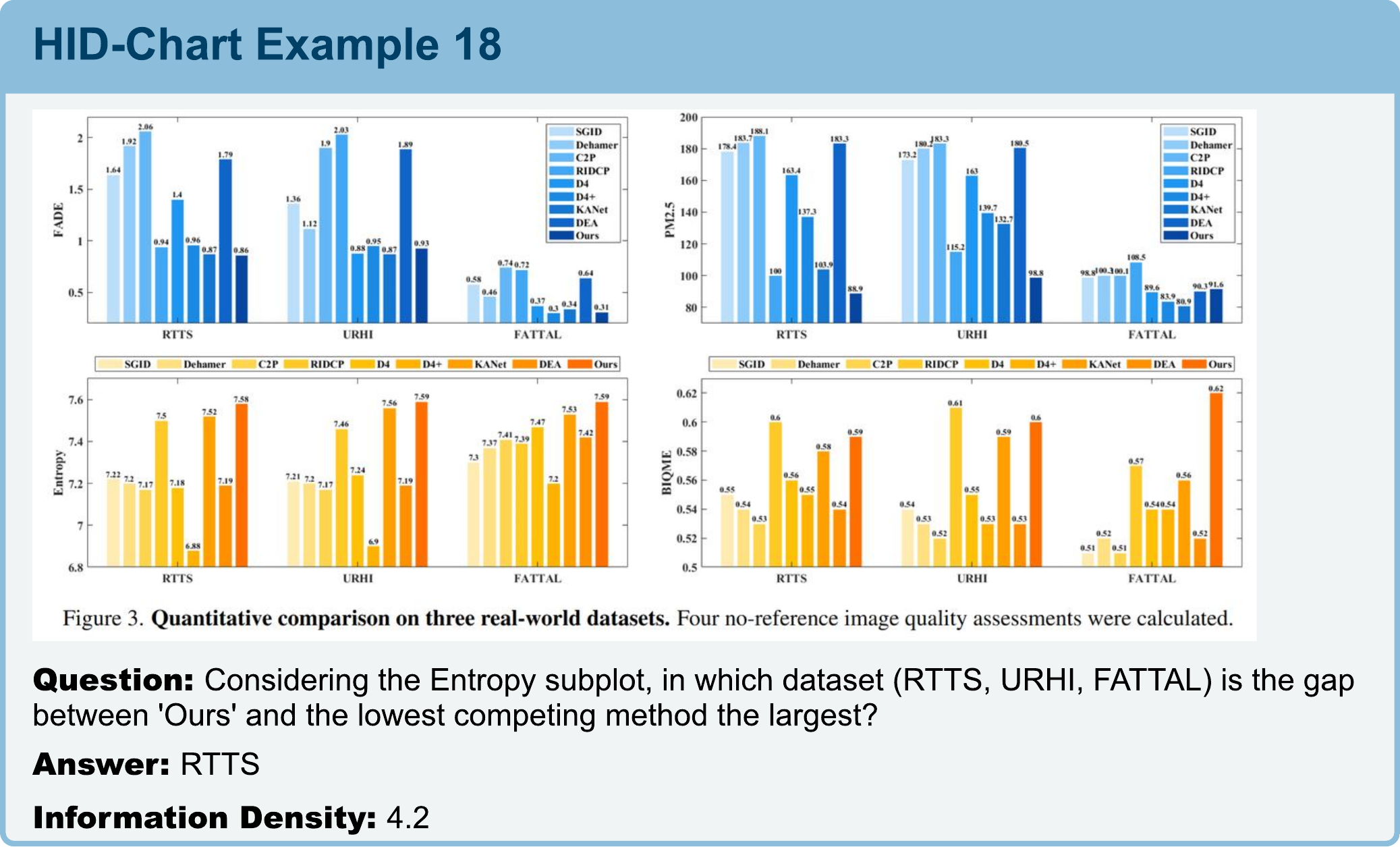}
\label{img:benchmark_case18}
\end{figure*}

\begin{figure*}[t]
\centering
\includegraphics[width=\textwidth]{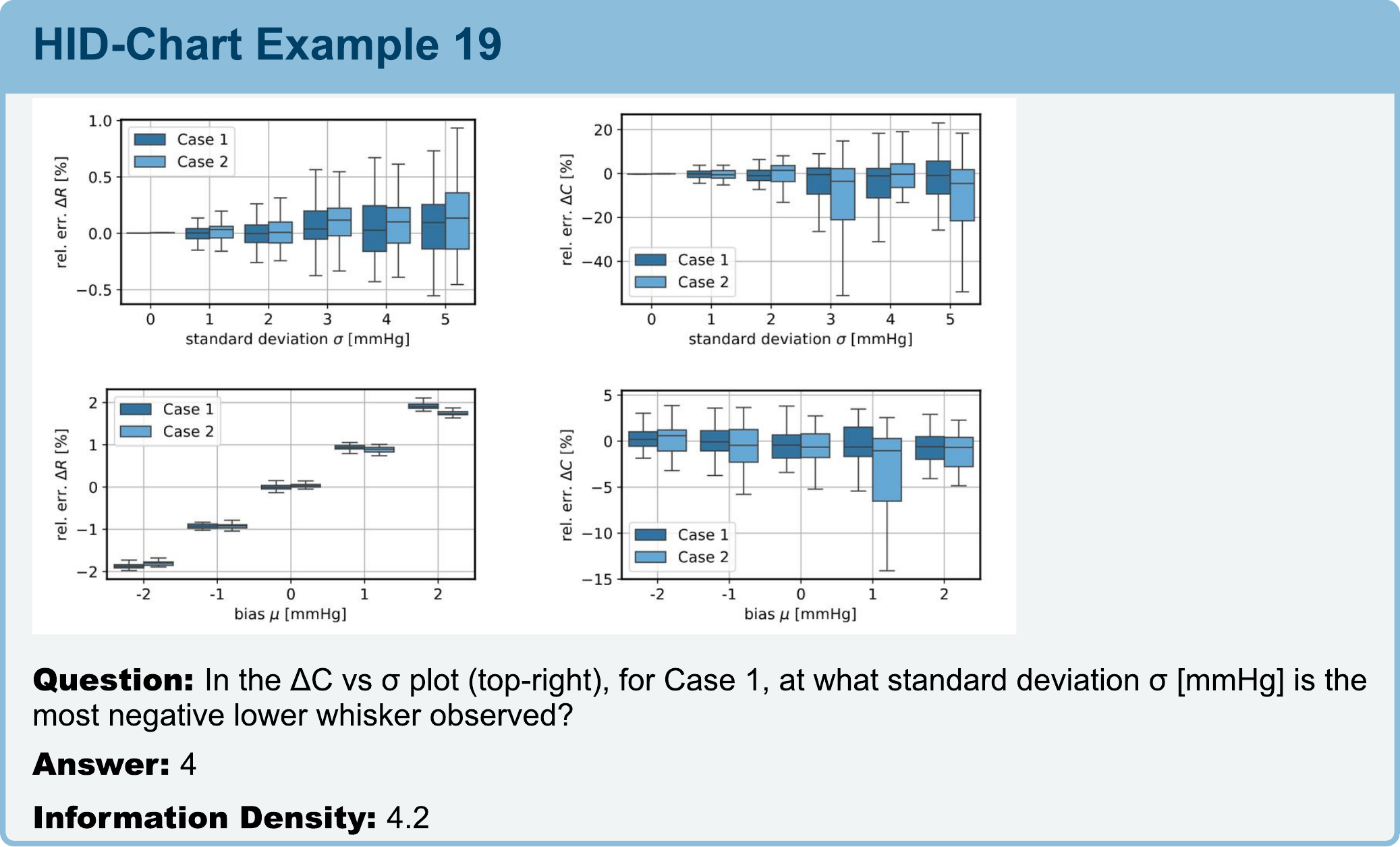}
\label{img:benchmark_case19}
\end{figure*}

\begin{figure*}[t]
\centering
\includegraphics[width=\textwidth]{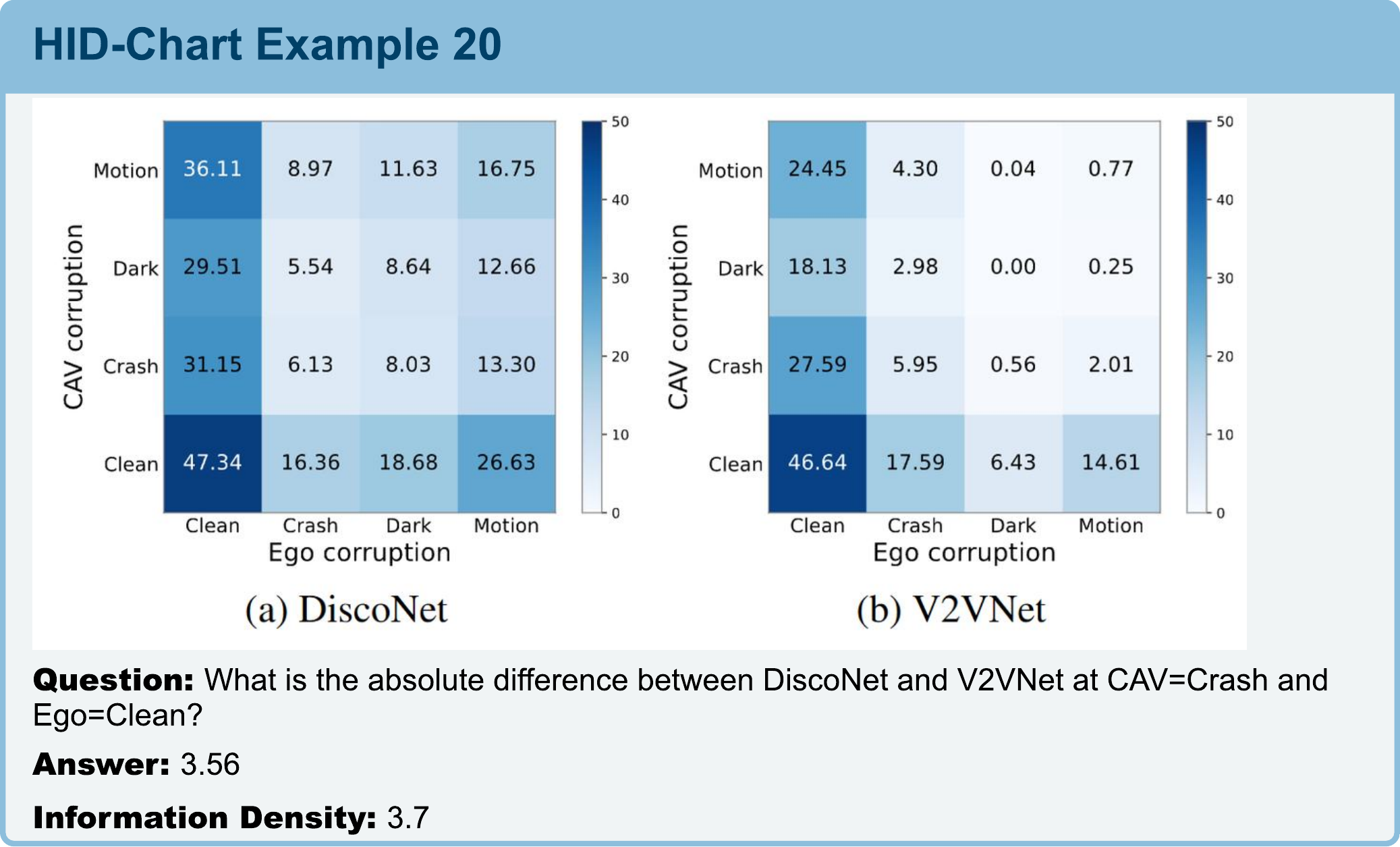}
\label{img:benchmark_case20}
\end{figure*}

\begin{figure*}[t]
\centering
\includegraphics[width=\textwidth]{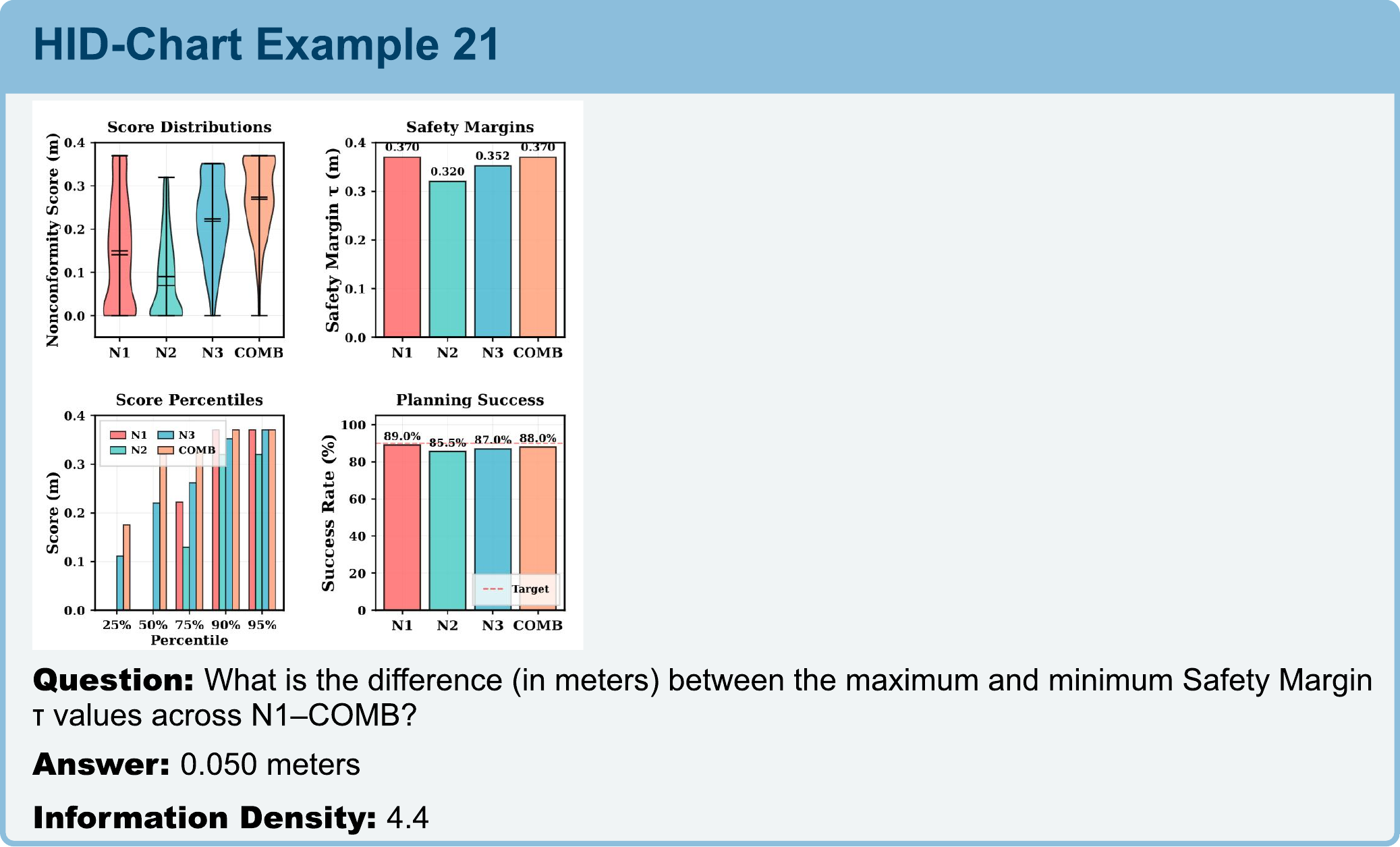}
\label{img:benchmark_case21}
\end{figure*}

\FloatBarrier


\end{document}